\documentclass{article} 
\usepackage{iclr2024_conference,times}

\usepackage{amsmath,amsfonts,bm}

\def\eqref#1{equation~\ref{#1}}

\def\1{\bm{1}}

\DeclareMathAlphabet{\mathsfit}{\encodingdefault}{\sfdefault}{m}{sl}
\SetMathAlphabet{\mathsfit}{bold}{\encodingdefault}{\sfdefault}{bx}{n}

\def\sD{{\mathbb{D}}}

\def\sS{{\mathbb{S}}}

\usepackage{subfigure}
\usepackage{hyperref}
\usepackage{url}
\usepackage{booktabs}
\usepackage{algorithm}
\usepackage{algorithmic}
\usepackage{amssymb}
\usepackage{multirow}
\usepackage{xcolor}
\definecolor{mygreen}{RGB}{11,200,11}
\definecolor{mygreen2}{RGB}{0,150,0}
\usepackage{array}
\usepackage{makecell}
\usepackage{graphicx}
\usepackage{wrapfig}

\title{\textbf{E}scape \textbf{S}ky-high \textbf{C}ost: \textbf{E}arly-stopping \textbf{S}elf-\textbf{C}onsistency for Multi-step Reasoning}

\author {
    \textbf{Yiwei Li}\textsuperscript{\rm 1}\footnotemark[2], \hspace{0.4cm}
    \textbf{Peiwen Yuan}\textsuperscript{\rm 1}\footnotemark[2], \hspace{0.4cm}
    \textbf{Shaoxiong Feng}\textsuperscript{\rm 2}, \hspace{0.4cm}
    \textbf{Boyuan Pan}\textsuperscript{\rm 2}, \hspace{0.4cm}
    \textbf{Xinglin Wang}\textsuperscript{\rm 1}, \hspace{1cm} \\ 
    \textbf{Bin Sun}\textsuperscript{\rm 1}, 
    \hspace{0.4cm}\textbf{Heda Wang}\textsuperscript{\rm 2}, 
    \hspace{0.4cm} \textbf{Kan Li}\textsuperscript{\rm 1}\footnotemark[3] \\
    \textsuperscript{\rm 1} School of Computer Science, Beijing Institute of Technology \\
    \textsuperscript{\rm 2} Xiaohongshu Inc \\
    \texttt{\{liyiwei,peiwenyuan,binsun,wangxinglin,likan\}@bit.edu.cn} \\
    \texttt{\{shaoxiongfeng2023,whd.thu\}@gmail.com} \\ \texttt{\{panboyuan\}@xiaohongshu.com}
}

\iclrfinalcopy 
\begin{document}

\maketitle
\renewcommand{\thefootnote}{\fnsymbol{footnote}} 
\footnotetext[2]{Equal contributions.} 
\footnotetext[3]{Corresponding author.} 

\begin{abstract}
Self-consistency (SC) has been a widely used decoding strategy for chain-of-thought reasoning. 
Despite bringing significant performance improvements across a variety of multi-step reasoning tasks, it is a high-cost method that requires multiple sampling with the preset size. 
In this paper, we propose a simple and scalable sampling process, \textbf{E}arly-Stopping \textbf{S}elf-\textbf{C}onsistency (ESC), to greatly reduce the cost of SC without sacrificing performance. 
On this basis, one control scheme for ESC is further derivated to dynamically choose the performance-cost balance for different tasks and models. 
To demonstrate ESC's effectiveness, we conducted extensive experiments on three popular categories of reasoning tasks: arithmetic, commonsense and symbolic reasoning over language models with varying scales. 
The empirical results show that ESC reduces the average number of sampling of chain-of-thought reasoning by a significant margin on six benchmarks, including MATH (-33.8\%), GSM8K (-80.1\%), StrategyQA (-76.8\%), CommonsenseQA (-78.5\%), Coin Flip (-84.2\%) and Last Letters (-67.4\%), while attaining comparable performances\footnotemark[1].
\end{abstract}

\footnotetext[1]{Our code and data have been released on \url{https://github.com/Yiwei98/ESC}.}
\section{Introduction}
Large language models (LLMs) have exhibited strong reasoning capabilities \citep{AGI}, especially with chain-of-thought (CoT) prompting \citep{COT}. 
Based on this, \citet{SC} introduce a simple decoding strategy called self-consistency (SC) to further improve reasoning performances, which takes advantage of the fact that complex reasoning tasks typically allow for more than one reasoning paths leading to the correct answer. 
In contrast to the standard chain-of-thought prompting which only generates the greedy one, this method samples multiple reasoning paths according to the predetermined sample size, and then derives the final answer through voting-based scheme.

However, despite generally leading to improvements, the SC strategy incurs a significant overhead proportional to the number of sampled outputs, under the assumption that the sampled outputs are of equal length.

Taking the most popular arithmetic reasoning benchmark for LLMs as an example, testing the entire MATH dataset with SC (sampling size is 64 as \citet{Minerva}) costs about 2000\$ through GPT-4 API, which is a significant burden for many researchers and organizations. Therefore, it is essential to minimize the cost of SC while maintaining performance.

\begin{figure}[ht]
\begin{center}
\includegraphics[width=1.0\textwidth]{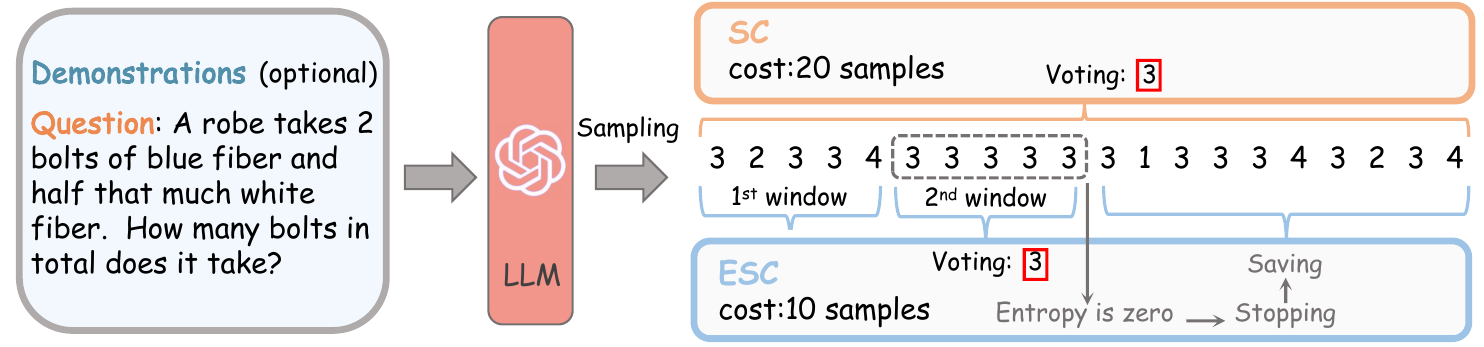}

\end{center}
\caption{Full process of ESC compared with original SC. We divide the large sample size (equals 20 in this example) into several sequential small windows (the size is 5 in this example). Stop sampling when answers within a window are all the same, i.e., the entropy score of predicted answer distribution is zero. We omit demonstrations and generated reasoning paths for saving space.}
\label{fig:intro}
\end{figure}

The process of generating multiple samples in SC can be viewed as approximating the true answer distribution predicted by the language model under a specific sampling temperature. 
Then the most frequent one is taken as the final answer to mitigate the stochasticity of the single-sampling strategy. 
However, given that only the most confident answer is needed for SC, it is not necessary whether the whole answer distribution fits perfectly. 
In some cases, the final result can be obtained by voting only within small part of the generated samples. 
Therefore, we argue that it is not necessary to directly generate all reasoning paths aligning with the preset sampling size for every input. 
Instead, the generation process can be serialized as smaller parts, each of which is named as sampling window. 

Considering both the small window and the large number of sampling outputs are generated from the same predicted answer distribution, it can be deemed as a probe to reveal some information of the true distribution with only a small sampling number.

For the answer distribution, one conjecture is that the correct ones are often concentrated, and the incorrect answers are scattered. 
We employ entropy as a representation of the answer distribution shape. 
Figure 2 shows the mean entropy value of distributions of correct and incorrect voting answer within a window respectively, showing that distributions with correct one as highest probability answer typically have much lower entropy values. 
Thus, it can be a indicator to determine whether sampling should continue. 
Base on this, we propose early-stopping self-consistency (ESC), truncating the sampling process with low entropy window. 
Figure~\ref{fig:intro} illustrates its process with an example. 
In order to maintain the performance as much as possible, we set the strictest threshold: the entropy equals zero, i.e., all generated samples within a window have the same answer. 
Stop sampling when this situation occurs to reduce sampling consumption while minimize performance impact.

\begin{wrapfigure}{r}{0.4\textwidth}
  \vspace*{0pt}
  \begin{centering}
    \includegraphics[scale=0.33]{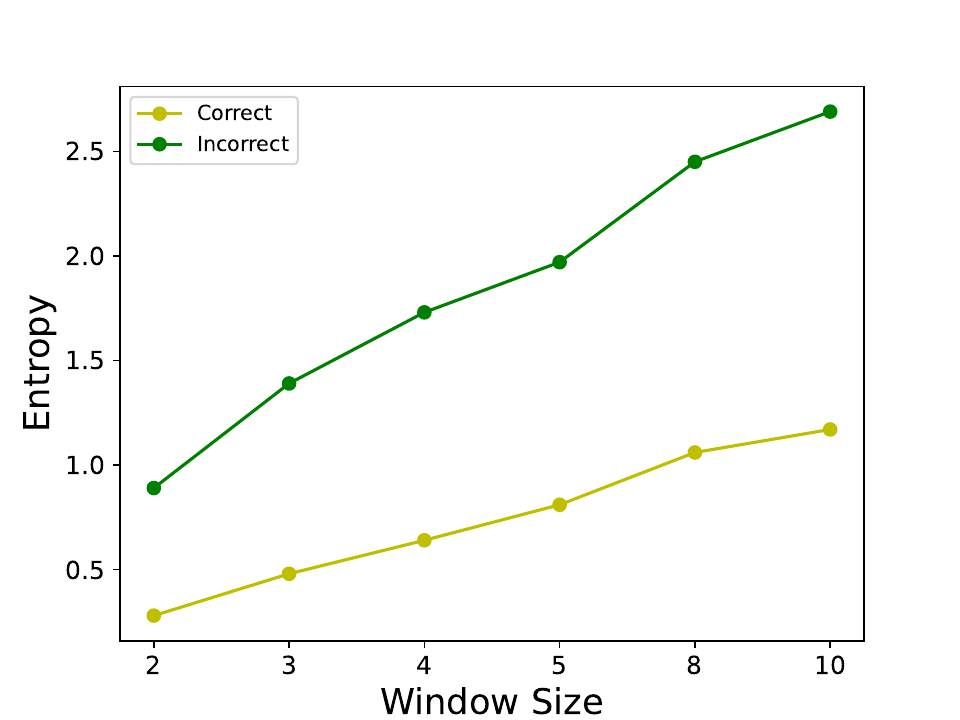}
    \setlength{\abovecaptionskip}{0pt}
    \caption{The mean entropy score within the window on MATH dataset from GPT-4.}
  \end{centering}
\label{fig:ent}
\end{wrapfigure}

Early stopping \citep{es} is a widely used technique to prevent poor generalization when training models. But in this paper, we introduce this strategy to early stop the generation process for saving the cost. 
As with original SC, ESC is completely unsupervised and model-agnostic, without any human annotation and additional training.
We derive the theoretical upper bound of inconsistent probability of the results with or without early stopping scheme in SC, indicating that ESC is highly likely to maintain performance.
In addition, one control scheme for ESC is further derivated to dynamically choose the performance-cost balance for different tasks and models by selecting the size of window and maximum sampling times, thereby meeting the practice requirements.

We evaluate ESC on a wide range of arithmetic, commonsense and symbolic reasoning tasks over three language models with varying scales: GPT-4, GPT-3.5-Turbo and Llama-2 7b. The empirical results show that ESC reduces the sampling times of chain-of-thought reasoning by a significant margin on six popular benchmarks, including MATH \citep{MATH} (-33.8\%), GSM8K \citep{GSM8K} (-80.1\%), StrategyQA \citep{SQA} (-76.8\%), CommonsenseQA \citep{CSQA} (-78.5\%), Coin Flip (-84.2\%) and Last Letters \citep{COT} (-67.4\%) , while attaining comparable performances. In additional experiments, we show our control scheme for ESC can predict the performance-cost balance accurately across various tasks and models, showcasing reliable application prospects. We also show ESC can robustly save cost considering different decoding settings and prompts.
 
\section{Method}

\subsection{Revisiting Self-Consistency}
Self-consistency \citep{SC} capitalizes on the notion that a intricate problem requiring logical thinking usually admits several distinct approaches that all lead to the same correct answer. Based on this,
multiple candidate predictions $\{\hat{y}^l\}^L$ to problem $x$ are suggested to generate through sampling, and the most consistent $\hat{y}$ is selected as the final prediction through a voting process:
\begin{equation}
\small
\hat{y} = \arg\max_{i} \sum_{l=1}^{L} \1_{\mathrm{\hat{y}}^l \mathrm{= i}}
\label{eq:sc1}
\end{equation}
where $\1_\mathrm{\hat{y}^l = i}$ is the indicator function (equal to 1 when $\hat{y}^l$ is equal to prediction $i$, and 0 otherwise). We conduct a more in-depth derivation as follows:
\begin{equation}
\small
\begin{aligned}
\hat{y} &= \arg\max_{i} \frac{\sum_{l=1}^{L} \1_{\mathrm{\hat{y}}^l \mathrm{= i}} }{L} 
    = \arg\max_{i} f^L(i)
\label{eq:sc2}
\end{aligned}
\end{equation}
where $f^L(i)$ denotes the frequency of the model's predicted outcome being $i$ in $L$ sampling instances.
According to the Law of Large Numbers \citep{LawofLargeNumber}, as the sample size $L$ approaches infinity, $f^L(i) = P(i)$, where $P(i)$ represents the true probability of the model predicting the outcome as $i$. On this basis, we can further deduce:
\begin{equation}
\small
\hat{y} = \arg\max_{i} P(i)\ \ (Given\ L \to \infty)
\label{eq:sc3}
\end{equation}

According to ~\eqref{eq:sc3}, we can re-conceptualize SC as the process of mitigating noise introduced by individual sampling through multiple samplings. Its objective is to ensure that the prediction with the highest probability, denoted as $\arg\max_{i} P(i)$ is chosen as the final answer.

From this perspective, we contend that when the entropy of $P$ is low (an extreme case being the one-hot-like distribution), a smaller value of $L$ ($L=1$ is enough for one-hot-like distribution) can mitigate the impact of sampling noise according to the Law of Large Numbers. 

\subsection{Early-Stopping Self-Consistency}
Building upon the aforementioned analysis, we propose \textbf{E}arly-Stopping \textbf{S}elf-\textbf{C}onsistency (ESC) to achieve comparable performance to SC at a much lower sampling cost by adaptively adjusting the sampling times. The execution flow of ESC is illustrated in Algorithm \ref{alg:hard}.

\begin{algorithm}
\small
\caption{Early-Stopping Self-Consistency.}\label{alg:hard}
\begin{algorithmic}
\REQUIRE model \textbf{$\mathcal{M}$}, dataset \textbf{$\sD=\{(x,y)\}^N$}, window size \textbf{$w$}, max sampling size \textbf{$L$}, past sampling set \textbf{$\sS_{past}$}
\ENSURE predictions set \textbf{$\sS_{predictions}$}
\STATE $\sS_{predictions}\leftarrow \varnothing$
\FOR{$i \in$ range $N$}
    \STATE $\sS_{candidates} \leftarrow \varnothing$,  $\sS_{window} \leftarrow \sS_{past}^i$ \textbf{if} $\sS_{past} \neq \varnothing$ \textbf{else} $\varnothing$
    \FOR{$j \in$ range $(L//w)$}
        \STATE $\sS_{window} \leftarrow$ Sampling predictions $w$ times from $\mathcal{M}$ given $x^i$ \textbf{if} $\sS_{window} = \varnothing$ \textbf{else} $\sS_{window}$
        \IF{predictions in $\sS_{window}$ are the same}
            \STATE $\sS_{candidates} \leftarrow \sS_{window}\, $ \textbf{break}
        \ENDIF
        \STATE $\sS_{candidates} \leftarrow \sS_{candidates}+\sS_{window}$, $\sS_{window} \leftarrow \varnothing$
    \ENDFOR
    \STATE $\sS_{predictions} \leftarrow \sS_{predictions}$.Append($\arg\max_{i} \sum_{p \in {\sS_{candidates}}} \1_{p\mathrm{=}i}$)
\ENDFOR 
\RETURN $\sS_{predictions}$
\end{algorithmic}
\end{algorithm}

We view a consecutive set of $w$ sampling predictions as an observation window for the probability distribution $P$ predicted by model $\mathcal{M}$ for input $x$. 
When these $w$ predictions are all the same, the entropy of $P$ is likely to be sufficiently low. At this point, the voting result of these $w$ samples is very likely to be equal to $\arg\max_{i} P(i)$, which is exactly the voting result that SC aims for. Therefore, we stop sampling when this situation occurs to save additional sampling consumption while barely affecting performance. We will iterate to obtain multiple observation windows until the preset sampling size $L$ is reached if no observation window meets the need during this procedure. The predictions of all the samples will constitute the final outcome according to ~\eqref{eq:sc1}.

\subsection{Theoretical Analysis}
\label{ch:theoretical}
Now, we analyze to what extent ESC will effect the performance compared with SC. For simplicity, we study the case where $L$ is infinite (we will examine the case of bounded $L$ in section \ref{ch:exp_per}). We conduct a \textit{one proportion z-test} with the following null hypothesis,
\[ H_0: \text{Prediction $p$ appears in an observation window $T$ times, where $p \neq \arg\max_{i} P(i)$.} \]
According to the definition of \textit{one proportion z-test}, the calculation formula for the z-statistic is:
\begin{equation}
\small
z = \frac{(\hat{T}-T_{\mu})*\sqrt{n}}{T_{\sigma}}
\label{eq:z-test}
\end{equation}
where $\hat{T}$ is the observed mean, $T_{\mu}$ represents the expected mean, $T_{\sigma}$ represents the expected standard deviation, and $n$ represents the observed times. 
Considering the null hypothesis $H_0$, where $p \neq \arg\max_{i} P(i)$, we can derive as follows:
\begin{equation}
\small
P(p) \leq (P(p) + max(P))/2 
    \leq \sum_{i} P(i)/2 
    = 1/2
\label{eq:p}
\end{equation}

As $w$ samples make up an observation window, according to Bernoulli distribution and ~\eqref{eq:p}:
\begin{equation}
\begin{gathered}
\small
T_{\mu} = w*P(p) \leq w/2
\small \\
T_{\sigma} = \sqrt{w*P(p)(1-P(p))} \leq \sqrt{w}/2
\end{gathered}
\label{eq:t_mu2sig}
\end{equation}

Taking ~\eqref{eq:t_mu2sig} into ~\eqref{eq:z-test}, z-statistic has the following lower bound when $\hat{T} \geq T_{\mu}$:
\begin{equation}
\small
z \geq \frac{(\hat{T}-w/2)*\sqrt{n}}{\sqrt{w}/2}
\label{eq:z-lower}
\end{equation}
The voting results of ESC are inconsistent with the voting results of SC if and only if $T$ equals to $w$ (when early-stop happens while $p \neq \arg\max_{i} P(i)$). When this situation occurs, we have the observed times $n = 1$ and $\hat{T} = w$, which we take into~\eqref{eq:z-lower}:
\begin{equation}
\small
z \geq \frac{(w-w/2)*\sqrt{1}}{\sqrt{w}/2} = \sqrt{w}
\label{eq:z-lower2}
\end{equation}

Suppose $w = 8$, and we choose to reject $H_0$ if $z \geq \sqrt{8}$ (the lower bound in ~\eqref{eq:z-lower2}). In this case, the probability of a false positive is $\leq 2 \times 10^{-3}$ (one-sided p-value corresponding to $z = \sqrt{8}$). This means that when we conduct ESC with $w = 8$, the probability of the voting outcome being inconsistent with the voting outcome of SC (false positive) is $\leq 2 \times 10^{-3}$, which is extremely low. According to ~\eqref{eq:z-lower2}, we can see that as $w$ increases, the lower bound of $z$ increases accordingly, resulting in a smaller upper bound of the probability that ESC performance being affected.

\subsection{Control Scheme for Early-stop Self-Consistency}
In practical applications, the desired scenario is that we can adjust the ESC strategy (window size $w$, max sampling size $L$) based on our sampling budgets and performance requirements.
Therefore, we propose a control scheme for ESC to achieve this goal. Specifically, we will deduce the expectation of voting performance and sampling cost under different ($w$, $L$) settings based on the first observation window (denote its window size as $w_0$).

\subsubsection{The Expectation of Sampling Cost}
\label{ch:exp_cost}

We first use the sampling frequency in the first observation window to approximate the true probability distribution $P$, and denote it as $\hat{P}$. Based on this, the probability of stopping sampling in each observation window (where the sampled values within the window are all the same) is:
\begin{equation}
\small
\hat{P}_{stop} = \sum_{i} \mathrm{pow}(\hat{P}(i),w)
\label{eq:p_stop}
\end{equation}
According to Algorithm \ref{alg:hard}, the expected average sampling times $\hat{L}$ of dataset $\sD$ is:
\begin{equation}
\small
\mathbb{E}(\hat{L}) = \mathbb{E}_{\hat{P}\in\mathcal{M}(\sD)} \sum_{j=0}^{L//w-1} [(\hat{P}_{stop}\times\mathrm{pow}(1-\hat{P}_{stop},j)\times j \times w) + \mathrm{pow}(1-\hat{P}_{stop}, L//w) \times L]  + w_0
\label{eq:l_num}
\end{equation}
see Appendix \ref{appendix:derivation} for detailed derivation. According to ~\eqref{eq:l_num}, we can determine the mapping relationship from the choice of ($w$, $L$) to the expected sampling cost $\hat{L}$ based on the sampling results of the first observation window.

\subsubsection{The Expectation of Voting Performance}
\label{ch:exp_per}
When $L$ is bounded (practical scenarios), we discuss the probability of ESC voting outcome being inconsistent with $\arg\max_{i} P(i)$ with (denoted as $Q_w(\hat{P})$) and without (denoted as $Q_o(\hat{P})$) the occurrence of early-stop.

When early-stop happens, similar to the derivations in section \ref{ch:theoretical}, we substitute $\hat{P}$ to $P$ to calculate the z-statistic under the $\hat{P}$ distribution. We then calculate the sum of the probabilities that the voting outcome of ESC being inconsistent with the voting outcome of SC for all candidate $p$ when early-stop happens, where $p \neq \arg\max_{i} \hat{P}(i)$:
\begin{equation}
\small
Q_w(\hat{P}) = \sum_{p\neq \arg\max_{i} \hat{P}(i)} \mathrm{querying}(\frac{w-w*\hat{P}(p)}{\sqrt{w*\hat{P}(p)(1-\hat{P}(p))}})
\label{eq:Q_w}
\end{equation}
where $\mathrm{querying}(\cdot)$ represents the process of querying p-value corresponding to the z-statistic.

When early-stop not happens, ESC degenerates into SC. We view the whole $L$ samples as an observation window and has the following derivation of z-statistic:
\begin{equation}
\small
z = \frac{\hat{T}-L*\hat{P}(p)}{\sqrt{L*\hat{P}(p)(1-\hat{P}(p))}}
\label{eq:new_z}
\end{equation}
If prediction $p$ is selected as the voting outcome, then the the following inequality should hold: $\hat{T} \geq L/2$. Combining with ~\eqref{eq:new_z}, we derive the upper bound of $Q_o(\hat{P})$:
\begin{equation}
\small
Q_o(\hat{P}) \leq \sum_{p\neq \arg\max_{i} \hat{P}(i)} \mathrm{querying}(\frac{L/2-L*\hat{P}(p)}{\sqrt{L*\hat{P}(p)(1-\hat{P}(p))}})
\label{eq:Q_o}
\end{equation}

According to ~\eqref{eq:p_stop}, the probability of that early-stop not happens is $\mathrm{pow}(1-\hat{P}_{stop}, L//w)$. Thus, the upper bound of the expected probability of ESC voting outcome being inconsistent with $\arg\max_{i} P(i)$ is:
\begin{equation}
\small
\mathbb{E}(Q) \leq \mathbb{E}_{\hat{P}\in\mathcal{M}(\sD)} (1-\mathrm{pow}(1-\hat{P}_{stop}, L//w)) \times Q_w(\hat{P}) + \mathrm{pow}(1-\hat{P}_{stop}, L//w) \times Q_o(\hat{P})
\label{eq:Q}
\end{equation}

\subsubsection{Controllable Early-Stop Self-Consistency}
Based on the above derivation, we propose our control scheme for ESC as shown in Algorithm \ref{alg:scheme}. First, we sample $w_0$ times on the whole dataset. Based on the results of the first observation window, we calculate the expected sampling cost and performance under different settings of ($w$, $L$). Finally, considering the sampling budget and performance requirements, we choose appropriate values of ($w$, $L$) based on the respective expected values to execute ESC.

\begin{algorithm}
\small
\caption{Control Scheme for Early-Stop Self-Consistency.}
\begin{algorithmic}\label{alg:scheme}
\REQUIRE model \textbf{$\mathcal{M}$}, dataset \textbf{$\sD=\{(x,y)\}^N$}, initial window size \textbf{$w_0$} (recommended as 5), sampling budget \textbf{$B$}, performance expectation \textbf{$P$}.
\ENSURE Predictions set \textbf{$\sS_{predictions}$}
\STATE $\sS_{firstwindow}\leftarrow \varnothing$
\FOR{$(x,y) \in \sD$}
    \STATE $\sS_{window} \leftarrow$ Sampling predictions $w_0$ times from $\mathcal{M}$ given $x$
    \STATE $\sS_{firstwindow} \leftarrow \sS_{firstwindow}+\sS_{window}$
\ENDFOR
\FOR{$\forall$ ($w,L$)}
    \STATE $\mathbb{E}(\hat{L}) \leftarrow $ section \ref{ch:exp_cost} ($\sS_{firstwindow},\ w,\ L$),  $\mathbb{E}(\hat{Q}) \leftarrow $ section \ref{ch:exp_per} ($\sS_{firstwindow},\ w,\ L$)
    \IF{ $(\mathbb{E}(\hat{L}),\mathbb{E}(\hat{Q}))$ meets the need of $(B,P)$}
        \STATE $\hat{w} \leftarrow w, \ \hat{L} \leftarrow L \, $ \textbf{break}

    \ENDIF
\ENDFOR
\RETURN Algorithm \ref{alg:hard}($\mathcal{M}$, $\sD=\{(x,y)\}^N$, $\hat{w}$, $\hat{L}$, $\sS_{firstwindow}$)
\end{algorithmic}
\end{algorithm}

\section{Experiments}

\subsection{Experiment Setup}

We evaluate the proposed ESC on six benchmark datasets from three categories of reasoning tasks:
For arithmetic reasoning, we consider MATH \citep{MATH} and GSM8K \citep{GSM8K}. MultiArith \citep{multi}, SVAMP \citep{svamp}, AddSub \citep{AddSub} and ASDiv \citep{asdiv} are not chosen in this paper because they are relatively simple.
For commonsense reasoning, CommonsenseQA \citep{CSQA} and StrategyQA \citep{SQA} are used.

For symbolic reasoning, we use Last Letter Concatenation and Coin Flip from \citet{COT}. The data version is from \citet{ZeroCOT}.

ESC is evaluated across three language models with varying scales: GPT-4 \citep{GPT4}, GPT-3.5-Turbo and LLaMA-2 7b \citep{llama2}. All experiments are conducted in the few-shot setting without training or fine-tuning the language models. To ensure a fair comparison, we use the same prompts as \citet{COT}. 
Details on the prompts used are given in Appendix.

The sampling temperature $T$ for MATH is 0.5 while for other datasets is 0.7. GPT-4 and GPT-3.5-Turbo samples predictions without truncating.
For Llama 2, the threshold for top p truncation \citep{topp} is 0.9. Similarly to \citet{SC}, we provide an ablation study in Section~\ref{exp:robust} to show that ESC is generally robust to sampling strategies and parameters.

\subsection{Main Results}
The baseline we compare to is chain-of-thought prompting with greedy decoding (CoT) and self-consistency (SC) with sampling. Following \citet{Minerva}, the sample size $L$ for MATH is 64 and for others is 40, and ESC uses the same value as maximum sample size. Accordingly, the window size $w$ for MATH is 8 and for others is 5. We report the results averaged over 10 runs and omit variance for limited space.

\begin{table}[th]
    \small
    \renewcommand\tabcolsep{5.0pt}
    \caption{Accuracy (\%) and $\hat{L}$ (average actual number of generated samples in ESC, in gray) across six reasoning benchmarks. 
    $\hat{L}$-SC denotes the accuracy of SC with sample size as $\hat{L}$.}
    \begin{center}
    \begin{tabular}{c c l l l l l l}
    \toprule
         &  & MATH& GSM8K & CSQA & SQA & Letter & Coinflip \\ \toprule
      \multirow{5}{*}{GPT-4} &  CoT  & 50.44  & 87.70  & 83.71 &78.63  & 93.12  & 100.00  \\
       & SC&  60.32  & 89.29  & 87.18  & 81.67  & 95.00   & \ \ \ \ \ \ /\\
      & ESC &  60.32 \tiny \textcolor{mygreen}{(0.00)}  & 89.29 \tiny \textcolor{mygreen}{(0.00)}  & 87.18 \tiny \textcolor{mygreen}{(0.00)}  & 81.70 \tiny \textcolor{mygreen}{(+0.03)}  & 94.98 \tiny \textcolor{red}{(-0.02)}  & \ \ \ \ \ \ /\\ 
    & $\hat{L}$ &  \textcolor{gray}{42.40} \tiny \textcolor{mygreen}{(-21.60)} & \ \ \textcolor{gray}{7.98} \tiny \textcolor{mygreen}{(-32.02)}  & \ \ \textcolor{gray}{9.29} \tiny \textcolor{mygreen}{(-30.71)}  & \ \ \textcolor{gray}{7.19} \tiny \textcolor{mygreen}{(-31.39)}  & \ \ \textcolor{gray}{6.32} \tiny \textcolor{mygreen}{(-33.68)} &  \ \ \ \ \ \ / \\
    & $\hat{L}$-SC & 59.98 \tiny \textcolor{mygreen}{(-0.34)} & 89.07 \tiny \textcolor{mygreen}{(-0.22)}& 86.49 \tiny \textcolor{mygreen}{(-0.69)} & 81.40 \tiny \textcolor{mygreen}{(-0.27)} & 94.59 \tiny \textcolor{mygreen}{(-0.39)} & \ \ \ \ \ \ /\\ \midrule
      \multirow{5}{*}{\makecell[c]{GPT-3.5\\Turbo}} &  CoT  & 35.53 & 75.83 & 74.17  & 67.66  & 80.50 &\ \ 83.74 \\
       & SC&  49.97   & 85.69  & 78.10  & 75.90  & 83.21  & \ \ 99.54  \\
      & ESC &  49.96 \tiny \textcolor{red}{(-0.01)} & 85.67 \tiny\textcolor{red}{(-0.02)}  & 78.10 \tiny\textcolor{mygreen}{(0.00)}  & 75.71 \tiny\textcolor{red}{(-0.19)}  & 83.15 \tiny \textcolor{red}{(-0.06)}  & \ \ 99.49  \tiny\textcolor{red}{(-0.05)} \\ 
    & $\hat{L}$ &  \textcolor{gray}{52.37} \tiny \textcolor{mygreen}{(-11.63)} & \textcolor{gray}{14.65} \tiny \textcolor{mygreen}{(-25.35)}  & \textcolor{gray}{11.70} \tiny \textcolor{mygreen}{(-28.30)}  & \textcolor{gray}{8.51} \tiny \textcolor{mygreen}{(-27.93)} & \ \ \textcolor{gray}{8.82} \tiny \textcolor{mygreen}{(-31.18)}  & \ \ \textcolor{gray}{13.03} \tiny \textcolor{mygreen}{(-26.97)}\\
    & $\hat{L}$-SC & 49.79 \tiny\textcolor{mygreen}{(-0.13)} & 84.82 \tiny \textcolor{mygreen}{(-0.85)}& 77.67 \tiny \textcolor{mygreen}{(-0.43)} & 75.07 \tiny \textcolor{mygreen}{(-0.83)}  & 82.74 \tiny \textcolor{mygreen}{(-0.41)} & \ \ 98.67 \tiny \textcolor{mygreen}{(-0.82)} \\ \midrule
      \multirow{5}{*}{\makecell[c]{Llama-2\\7b}} &  CoT  & \ \ 5.09  & 18.07 & 65.28  & 46.23  & 14.87 & \ \ 54.74  \\
       & SC&  \ \ 7.68  & 21.75  & 67.70 & 63.15  & 23.32  & \ \ 59.13  \\
      & ESC &  \ \ 7.68 \tiny \textcolor{mygreen}{(0.00)}  & 21.74 \tiny\textcolor{red}{(-0.01)}  & 67.68 \tiny\textcolor{red}{(-0.02)}  & 63.01  \tiny\textcolor{red}{(-0.14)} & 23.32 \tiny \textcolor{mygreen}{(0.00)}  & \ \ 58.99 \tiny\textcolor{red}{(-0.14)} \\ 
    & $\hat{L}$ &  \textcolor{gray}{62.48} \tiny \textcolor{mygreen}{(-1.52)} & \textcolor{gray}{31.21} \tiny \textcolor{mygreen}{(-8.79)}  & \textcolor{gray}{11.82} \tiny \textcolor{mygreen}{(-28.18)}  & \textcolor{gray}{11.00} \tiny \textcolor{mygreen}{(-23.96)} & \textcolor{gray}{34.73} \tiny \textcolor{mygreen}{(-5.27)}  & \ \ \textcolor{gray}{14.87} \tiny \textcolor{mygreen}{(-25.13)} \\
    & $\hat{L}$-SC & \ \ 7.68 \tiny \textcolor{mygreen}{(0.00)} & 21.52 \tiny \textcolor{mygreen}{(-0.22)}& 66.97 \tiny \textcolor{mygreen}{(-0.71)}& 61.19 \tiny \textcolor{mygreen}{(-1.96)}& 23.11 \tiny \textcolor{mygreen}{(-0.21)}& \ \ 58.11 \tiny \textcolor{mygreen}{(-0.88)}\\
        \bottomrule
    \end{tabular}
    \end{center}
    \vspace{-0.2cm}
    \label{tb:main}
\end{table}

\paragraph{ESC significantly reduces costs while barely affecting performance.}
Table~\ref{tb:main} summarizes accuracy of CoT, SC, proposed ESC and $\hat{L}$, the average actual number of generated samples in ESC, for each dataset among three language models. The first observation is that SC outperforms CoT substantially, which confirms the effectiveness of the voting process for reasoning. For ESC, the $\hat{L}$ is largely smaller than the corresponding maximum sampling size $L$, while the accuracy has remained almost unchanged. Given that the accuracy of Coin Flip on GPT-4 is 100\%, there is no need to conduct SC and ESC on it.  
We also test SC with $\hat{L}$ as the sampling size ($\hat{L}$-SC), whose accuracy drops in accordance with the performance curve relative to the number of samples from \citet{SC}.
Overall, ESC can reduce costs significantly while barely affecting performance. In other words, ESC can get higher accuracy under the same sampling costs.
\begin{table}[th]
    \centering
    \small
    \renewcommand\tabcolsep{4.5pt}
    \caption{Reasoning accuracy (\%) and $\hat{L}$ with various max sampling size $L$. The window size is 8.}
    \begin{tabular}{c c l l l l l l}
    \toprule
        Model & Method & \ \ 16 & \ \ 24 & \ \ 32 & \ \ 40 &\ \ 48 &\ \  64 \\ \toprule
      \multirow{3}{*}{GPT-4} &  SC  & 58.92  & 59.40 & 59.77 & 59.95 & 60.07 & 60.31\\
       &  ESC   & 58.92  \tiny \textcolor{mygreen}{(0.00)}  & 59.40 \tiny \textcolor{mygreen}{(0.00)}  & 59.77  \tiny \textcolor{mygreen}{(0.00)}& 59.95 \tiny \textcolor{mygreen}{(0.00)} &60.07 \tiny \textcolor{mygreen}{(0.00)} &60.31 \tiny \textcolor{mygreen}{(0.00)} \\
      &  $\hat{L}$  & \textcolor{gray}{13.56} \tiny\textcolor{mygreen}{(-2.44)}& \textcolor{gray}{18.72} \tiny\textcolor{mygreen}{(-5.28)}&\textcolor{gray}{23.67}\tiny\textcolor{mygreen}{(-8.33)}&\textcolor{gray}{28.49} \tiny\textcolor{mygreen}{(-11.51)}&\textcolor{gray}{33.21} \tiny\textcolor{mygreen}{(-14.79)}&\textcolor{gray}{42.41} \tiny\textcolor{mygreen}{(-21.59)}\\ \midrule
      \multirow{3}{*}{\makecell[c]{GPT-3.5\\Turbo}} &  SC   & 47.34 &48.48 & 49.02 &49.40 & 49.65 & 49.96  \\
       &  ESC  & 47.33 \tiny \textcolor{red}{(-0.01)} & 48.49 \tiny \textcolor{mygreen}{(+0.01)}  & 49.02 \tiny \textcolor{mygreen}{(0.00)} &49.41 \tiny \textcolor{mygreen}{(+0.01)} &49.64 \tiny \textcolor{red}{(-0.01)}& 49.96 \tiny \textcolor{mygreen}{(0.00)} \\
      &  $\hat{L}$  & \textcolor{gray}{14.84} \tiny\textcolor{mygreen}{(-1.16)}& \textcolor{gray}{21.38} \tiny\textcolor{mygreen}{(-2.62)}&\textcolor{gray}{27.76} \tiny\textcolor{mygreen}{(-4.24)}&\textcolor{gray}{34.02} \tiny\textcolor{mygreen}{(-5.98)}&\textcolor{gray}{40.20} \tiny\textcolor{mygreen}{(-7.80)}& \textcolor{gray}{52.37} \tiny\textcolor{mygreen}{(-11.63)}\\ \midrule
      \multirow{3}{*}{\makecell[c]{Llama-2\\7b}} &  SC  & \ \ 7.10 & \ \ 7.28  & \ \ 7.40  & \ \ 7.45 & \ \ 7.54 &\ \ 7.70  \\
       &  ESC  &\ \ 7.10 \tiny \textcolor{mygreen}{(0.00)} & \ \ 7.28 \tiny \textcolor{mygreen}{(0.00)} & \ \ 7.40\tiny \ \ \textcolor{mygreen}{(0.00)}  & \ \ 7.45\tiny \ \ \textcolor{mygreen}{(0.00)}  & \ \ 7.54 \tiny \textcolor{mygreen}{(0.00)}& \ \ 7.70\tiny \ \ \textcolor{mygreen}{(0.00)} \\
      & $\hat{L}$ & \textcolor{gray}{15.88} \tiny\textcolor{mygreen}{(-0.12)} & \textcolor{gray}{23.72} \tiny\textcolor{mygreen}{(-0.28)} & \textcolor{gray}{31.52} \tiny\textcolor{mygreen}{(-0.48)}& \textcolor{gray}{39.29} \tiny\textcolor{mygreen}{(-0.71)}& \textcolor{gray}{47.04} \tiny\textcolor{mygreen}{(-0.96)}&\textcolor{gray}{62.48} \tiny\textcolor{mygreen}{(-1.52)} \\
        \bottomrule
    \end{tabular}
    \label{tb:length}
\end{table}

\begin{figure*}[!htb]
    \centering
    \subfigure{\includegraphics[width=0.32\hsize, height=0.25\hsize]{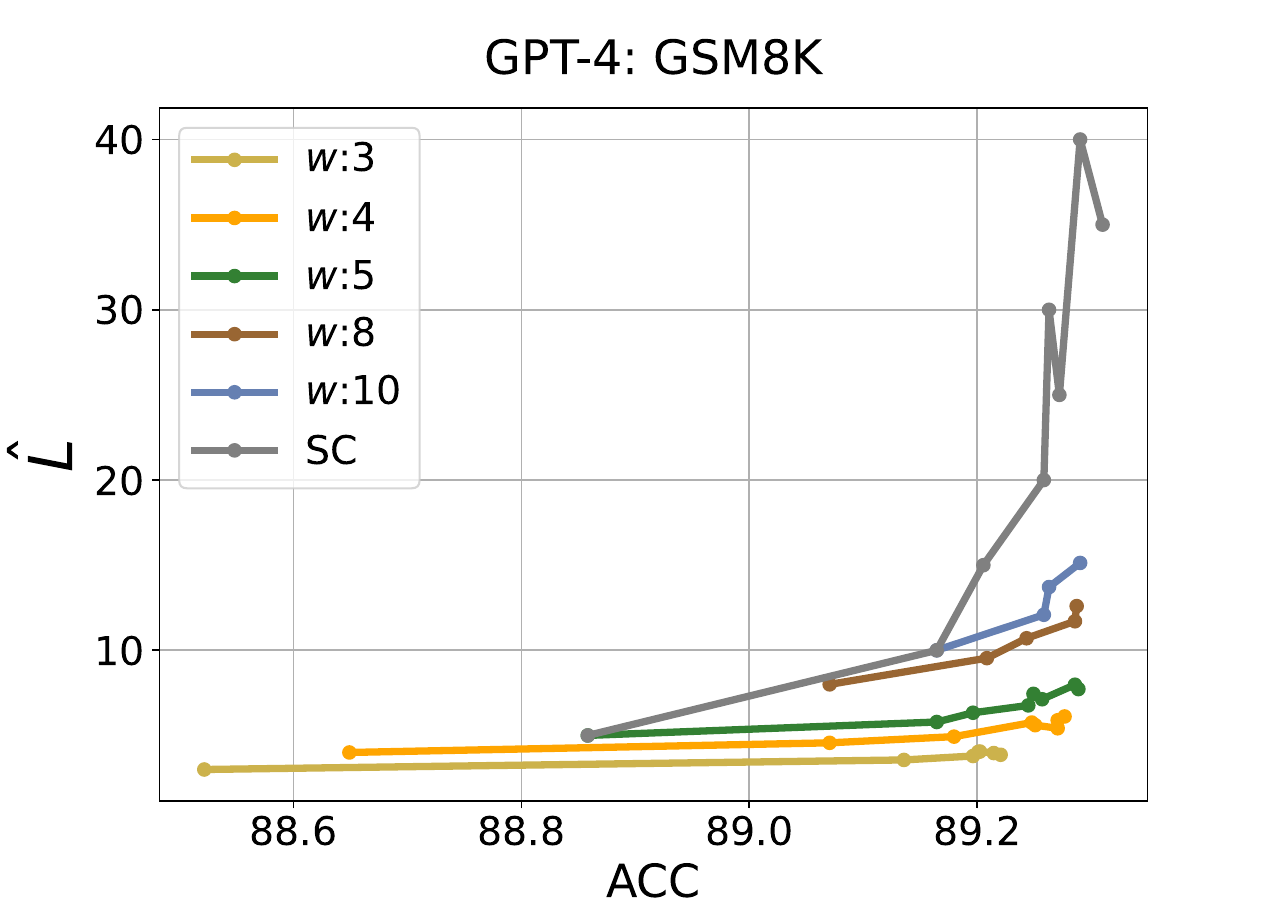}} \hspace{-0.50cm}
    \subfigure{\includegraphics[width=0.32\hsize, height=0.25\hsize]{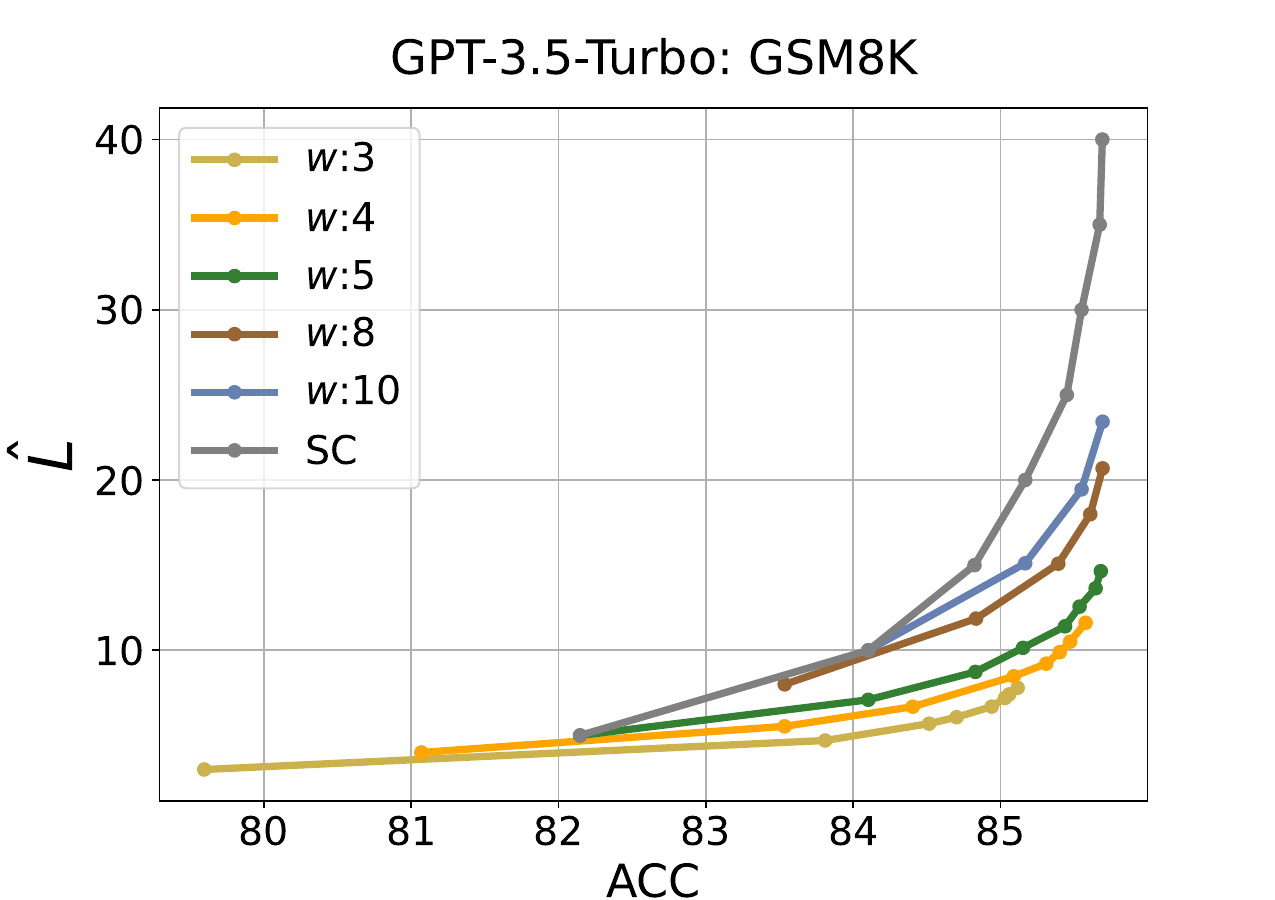} }\hspace{-0.47cm}
    \subfigure{\includegraphics[width=0.32\hsize, height=0.25\hsize]{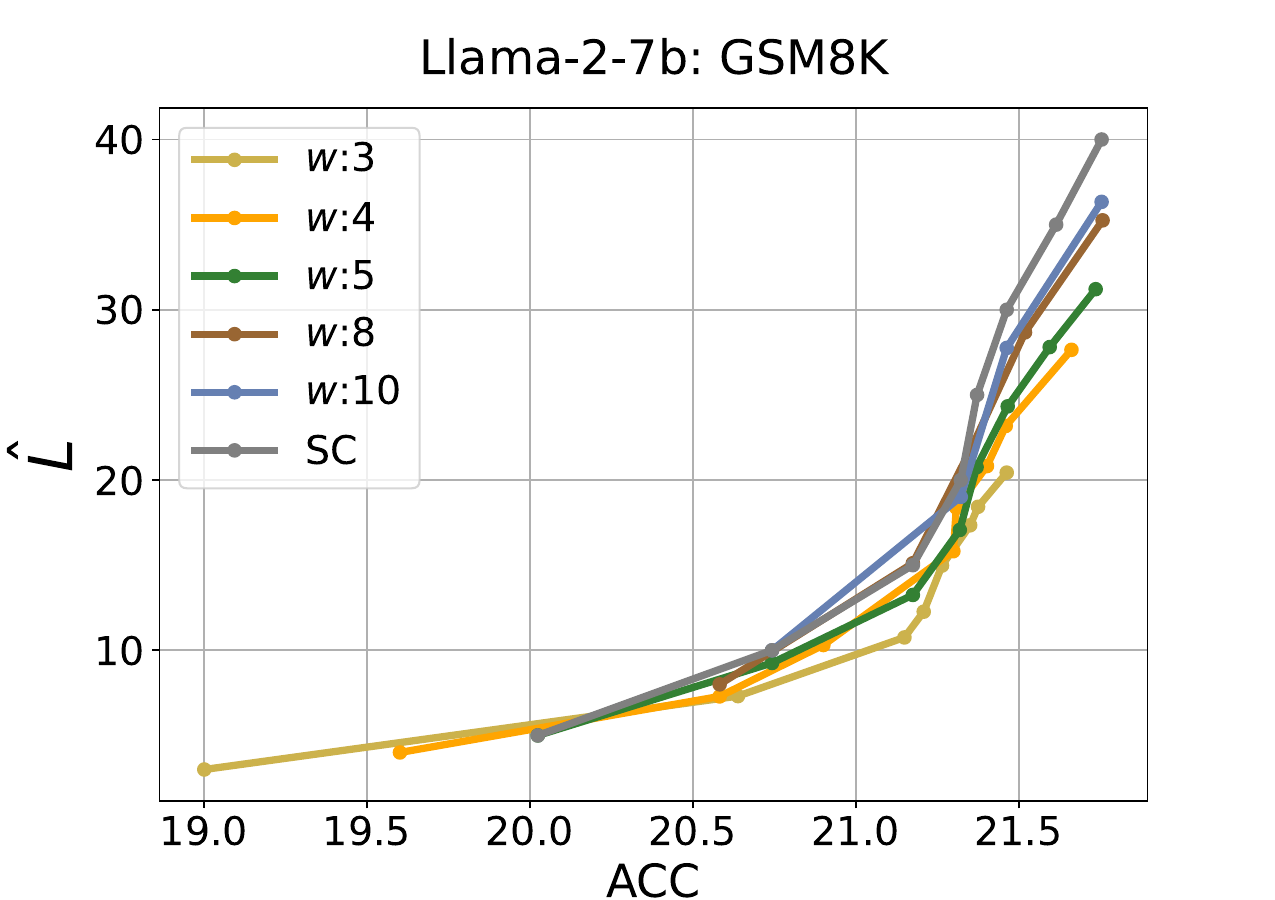}}\hspace{-0.4cm}
    \setlength{\belowcaptionskip}{0pt}
    \caption{Robustness analysis of ESC regarding the observation window size $w$ and max sampling size $L$ on GSM8K with different models.}
    \label{fig:scala}
\end{figure*}

\paragraph{ESC is a scalable decoding process across sampling and window size.}
We conduct experiments with various window size and sampling size to valid the scalability of ESC. 
Table~\ref{tb:length} shows the performance across different maximum sampling sizes. First we can see the performance of SC continuously improves as sampling size $L$ increases, which is consistent with the results in \citep{SC}. On this basis, ESC can significantly save costs while maintaining performance for different $L$. Figure~\ref{fig:scala} shows performance-cost balancing lines among three language models on GSM8K. ESC is robust to different window size and maximum sampling number. Please refer to Appendix-Figure~\ref{fig:appendix-figs} for results on other datasets.

\paragraph{Cost savings are positively correlated with performance.}
As shown in both Table~\ref{tb:main} and Table~\ref{tb:length}, an obvious phenomenon is that cost savings are positively correlated with performance. It is intuitive since better performance often eliminates the need for larger sample size as recommended in \citet{SC}. However, ESC does not require any prior knowledge of model capabilities and task difficulty. In addition, a control scheme for ESC is also proposed and will be evaluated in Section~\ref{exp:con}. 

\subsection{Effectiveness of control scheme for ESC}
\label{exp:con}
To assess the effectiveness of control scheme for ESC, we compared the consistency between our predicted and actual values of sampling cost $\hat{L}$ and performance change percentage between SC and ESC $P_{\delta}$ on the GSM8K dataset. We chose L1 norm and Pearson correlation to measure the consistency.\footnote{We vary the max sampling size to obtain multiple sets of results for calculating the Pearson correlation and mean L1 norm.} As shown in Table~\ref{tb:CSESC}, the Pearson correlations for both $\hat{L}$ and $P_{\delta}$ exceed 0.8, indicating a strong linear correlation between the predicted and actual values. Also, the L1 norm for both $\hat{L}$ and $P_{\delta}$ are very low. These results indicate that the predictions we obtain based on ~\eqref{eq:l_num} and ~\eqref{eq:Q} are highly reliable for balancing sampling cost and voting performance.
 
\begin{table}[th]
    \centering
    \small
    \caption{Consistency between the predicted (through control scheme for ESC) and actual values of $\hat{L}$: sampling times, and $P_{\delta}$ (\%): performance change percentage between SC and ESC. We choose L1 norm and Pearson correlation to measure the consistency. All the p-values $< 0.05$. }
    \begin{tabular}{c c c c c}
    \toprule
        Model & \ \ $\|\hat{L}^{act}-\hat{L}^{pre}\|_1$ & Pearson$(\hat{L}^{act},\hat{L}^{pre})$ & \ \ $\|P_{\delta}^{act}-P_{\delta}^{pre}\|_1$ & Pearson$(P_{\delta}^{act},P_{\delta}^{pre})$  \\ \toprule
        
      GPT-4  & 0.27  & 1.00 & 0.43 & 0.81 \\
      
       GPT-3.5 Turbo   & 0.62  & 1.00 & 0.06 & 0.86  \\
       
        Llama-2 7b & 1.43  & 1.00 & 0.52 & 0.86\\ 
        \bottomrule
    \end{tabular}
    \label{tb:CSESC}
\end{table}

\subsection{ESC for open-ended generations}
Original SC is only suitable for problems that have fixed answers, while \citet{open} extended it for open-ended generation tasks by replacing voting through text similarity matching. We conduct ESC on MBPP dataset \citep{MBPP} with various sampling size (window size is 5).
The results in Table~\ref{tb:open} shows that ESC is also suitable for open-ended task.
\begin{table}[th]
    \centering
    \small
    \caption{Reasoning accuracy (\%) and $\hat{L}$ for MBPP with various max length L on GPT-3.5-Turbo.}
    \begin{tabular}{c l l l l l}
    \toprule
        Method & \ \ 10 & \ \ 15 & \ \ 20 & \ \ 25 &\ \ 30  \\ \toprule
      SC  & 61.96  & 61.96 & 62.04 & 62.15 & 62.18\\
       ESC   & 61.96  \tiny \textcolor{mygreen}{(0.00)}  & 61.96 \tiny \textcolor{mygreen}{(0.00)}  & 62.02 \tiny \textcolor{red}{(-0.02)}& 62.11 \tiny \textcolor{red}{(-0.04)} &62.15 \tiny \textcolor{red}{(-0.03)}  \\
        $\hat{L}$  & \textcolor{gray}{5.62} \tiny\textcolor{mygreen}{(-4.38)}& \textcolor{gray}{6.02} \tiny\textcolor{mygreen}{(-8.98)}&\textcolor{gray}{6.32} \tiny\textcolor{mygreen}{(-13.68)}&\textcolor{gray}{6.57} \tiny\textcolor{mygreen}{(-18.43)}&\textcolor{gray}{6.79} \tiny\textcolor{mygreen}{(-23.21)}\\ 
        \bottomrule
    \end{tabular}
    \label{tb:open}
\end{table}

\subsection{Intersection Between ESC and SC}
\begin{wraptable}{r}{0.48\textwidth}
    \vspace{-30pt}
    \centering
    \small
    \caption{Intersection ratio between ESC and SC.}
    \begin{tabular}{l l l l}
    \toprule
       Model & GSM8K & SQA & Letter  \\ \toprule
      GPT-4  & 99.69  & 99.69 & 99.96 \\
      GPT-3.5-Turbo   & 99.92 & 99.25 & 99.76\\
      Llama-2-7b & 99.77 & 99.13  &99.75 \\ 
        \bottomrule
    \end{tabular}
    \label{tb:inter}
\end{wraptable}
According to Section~\ref{ch:theoretical}, the voting results of ESC should have a high probability of being consistent with SC. From Table~\ref{tb:inter} we can see that the intersection ratios of the voting results of SC and ESC are quite high, which indicates the upper bound of the performance being affected by ESC derived in Section~\ref{ch:exp_per} is reliable.

\subsection{Robustness of ESC}
\label{exp:robust}
\begin{figure*}[!htb]
    \centering

    \subfigure{\includegraphics[width=0.267\hsize, height=0.26\hsize]{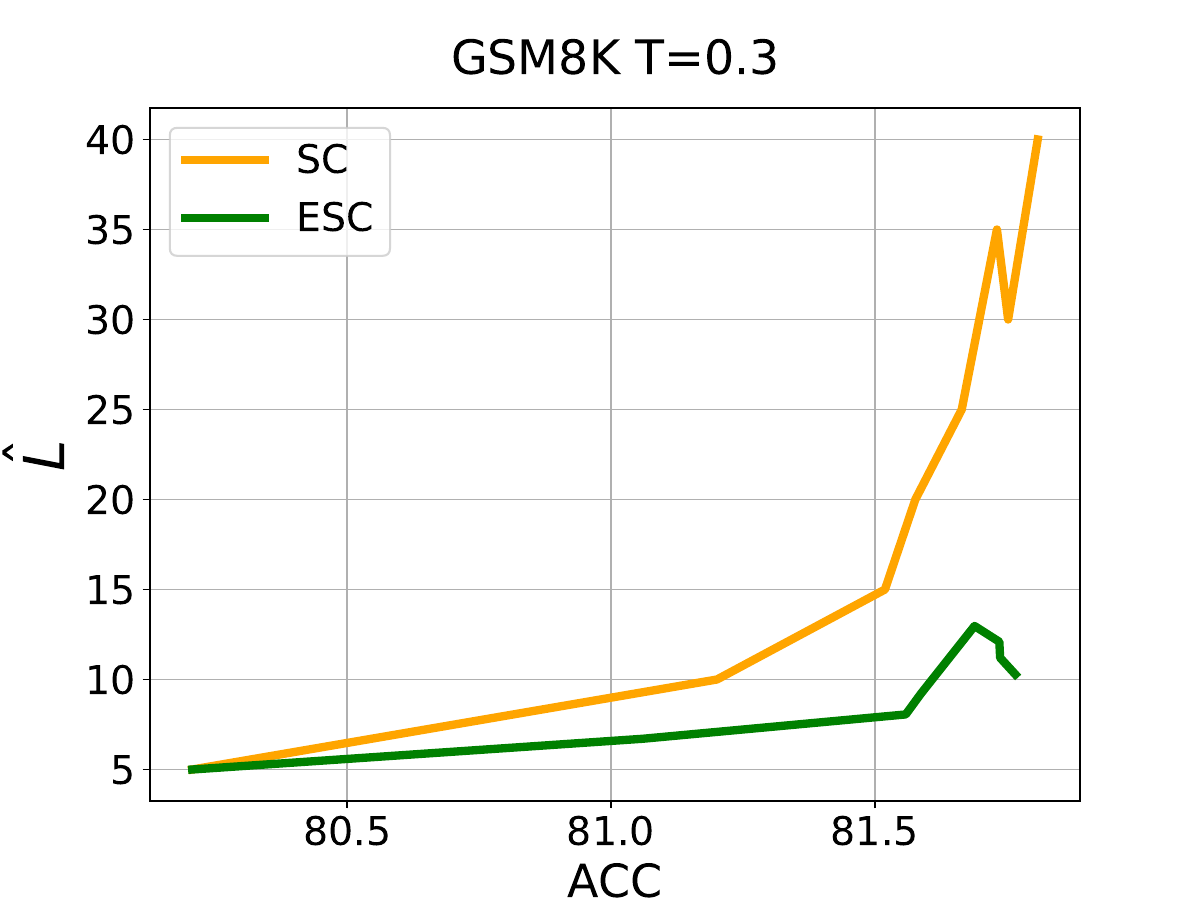}} \hspace{-0.50cm}
    \subfigure{\includegraphics[width=0.267\hsize, height=0.26\hsize]{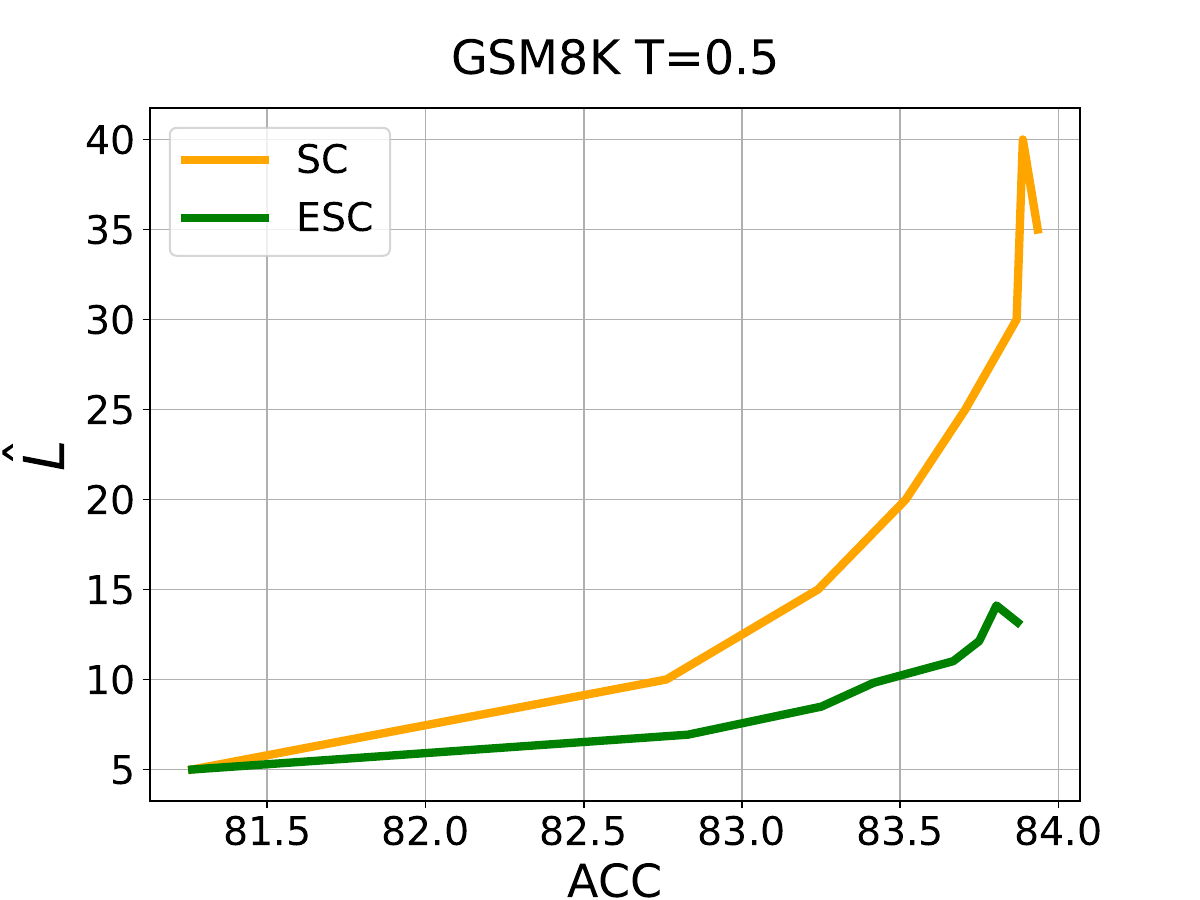}} \hspace{-0.47cm}
    \subfigure{\includegraphics[width=0.267\hsize, height=0.26\hsize]{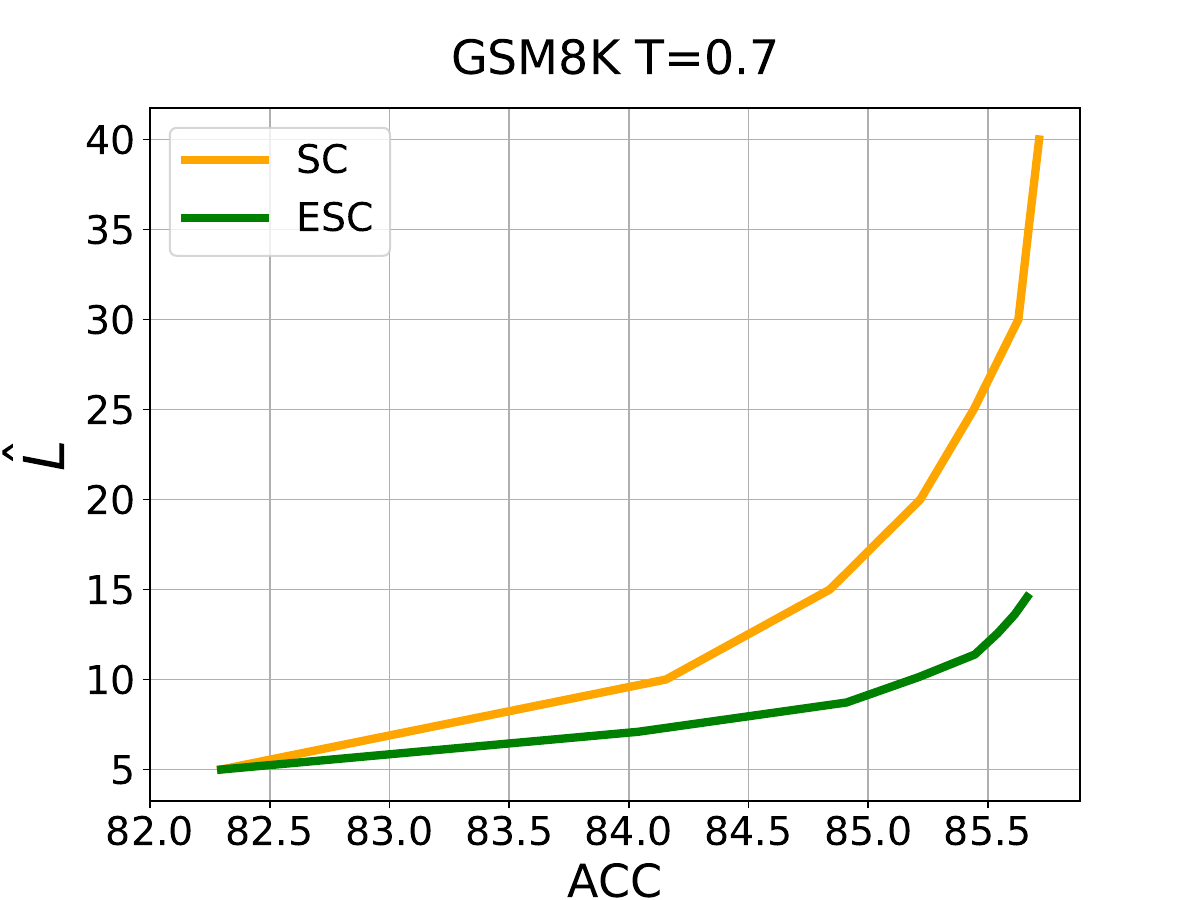}}\hspace{-0.4cm}
    \subfigure{\includegraphics[width=0.267\hsize, height=0.26\hsize]{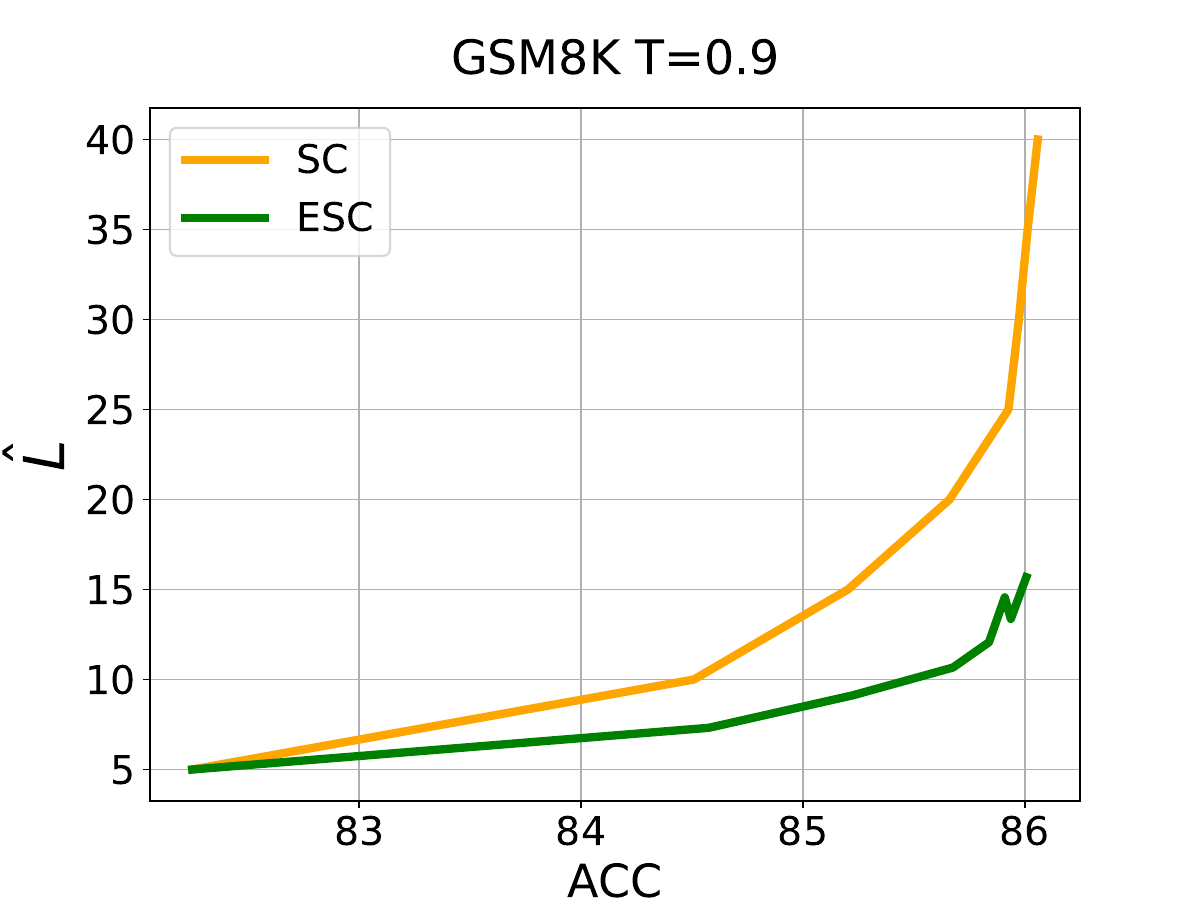}}
    \subfigure{\includegraphics[width=0.267\hsize, height=0.26\hsize]{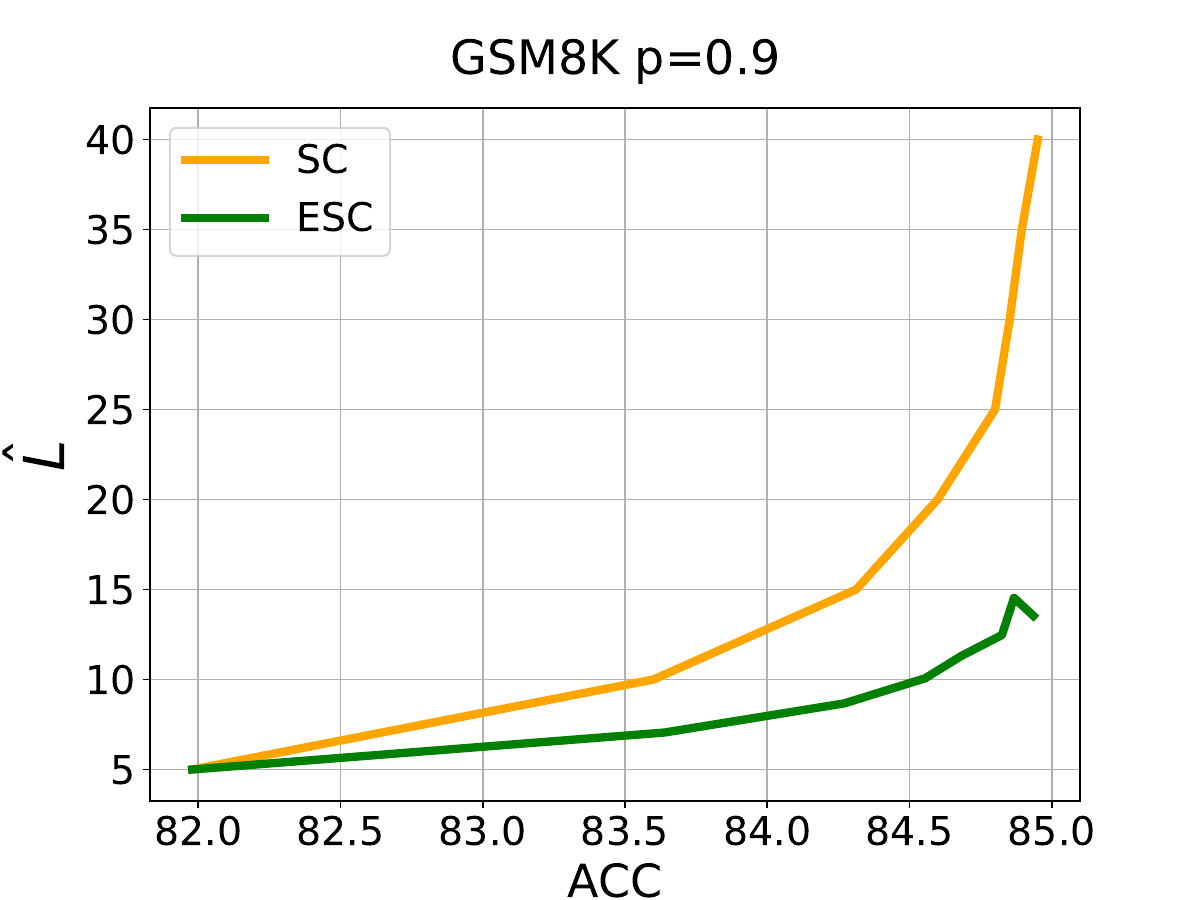}} \hspace{-0.50cm}
    \subfigure{\includegraphics[width=0.267\hsize, height=0.26\hsize]{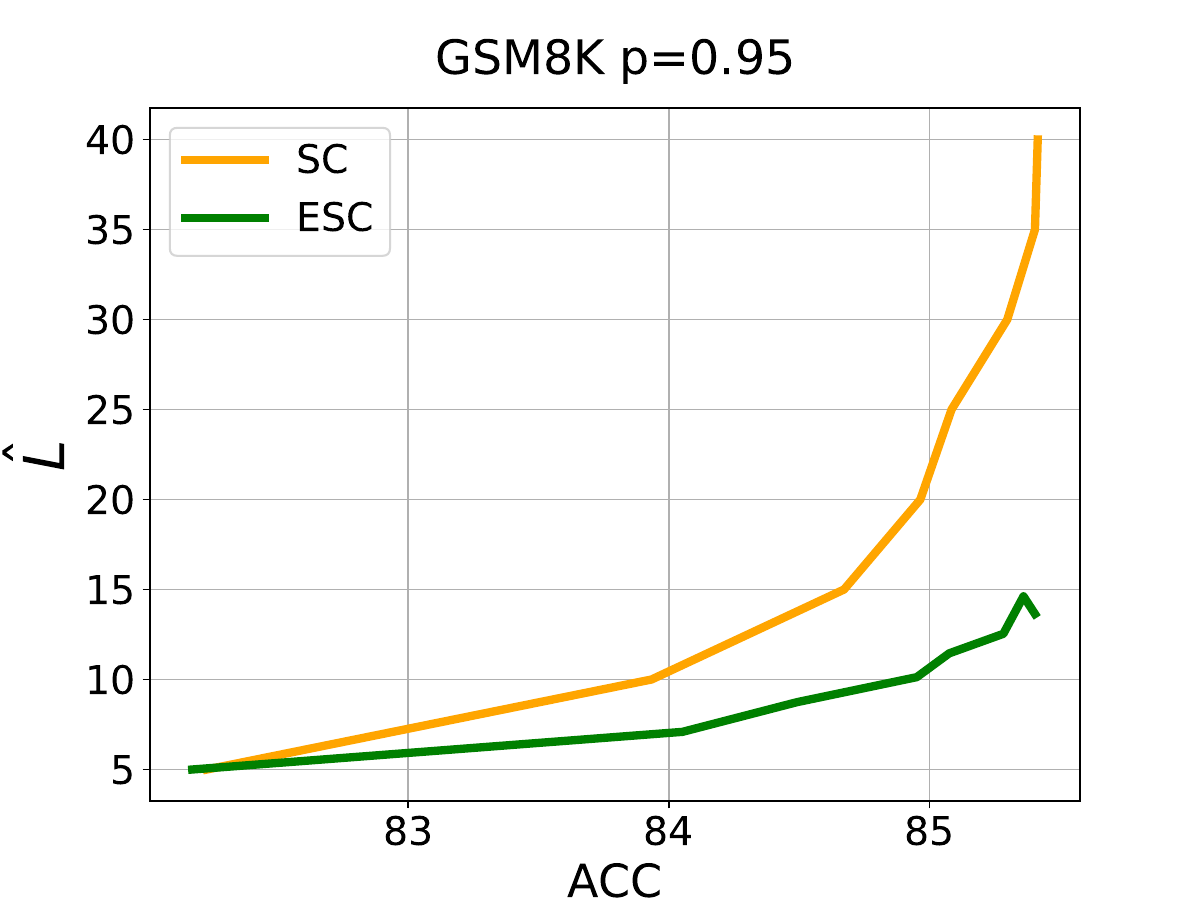}} \hspace{-0.47cm}
    \subfigure{\includegraphics[width=0.267\hsize, height=0.26\hsize]{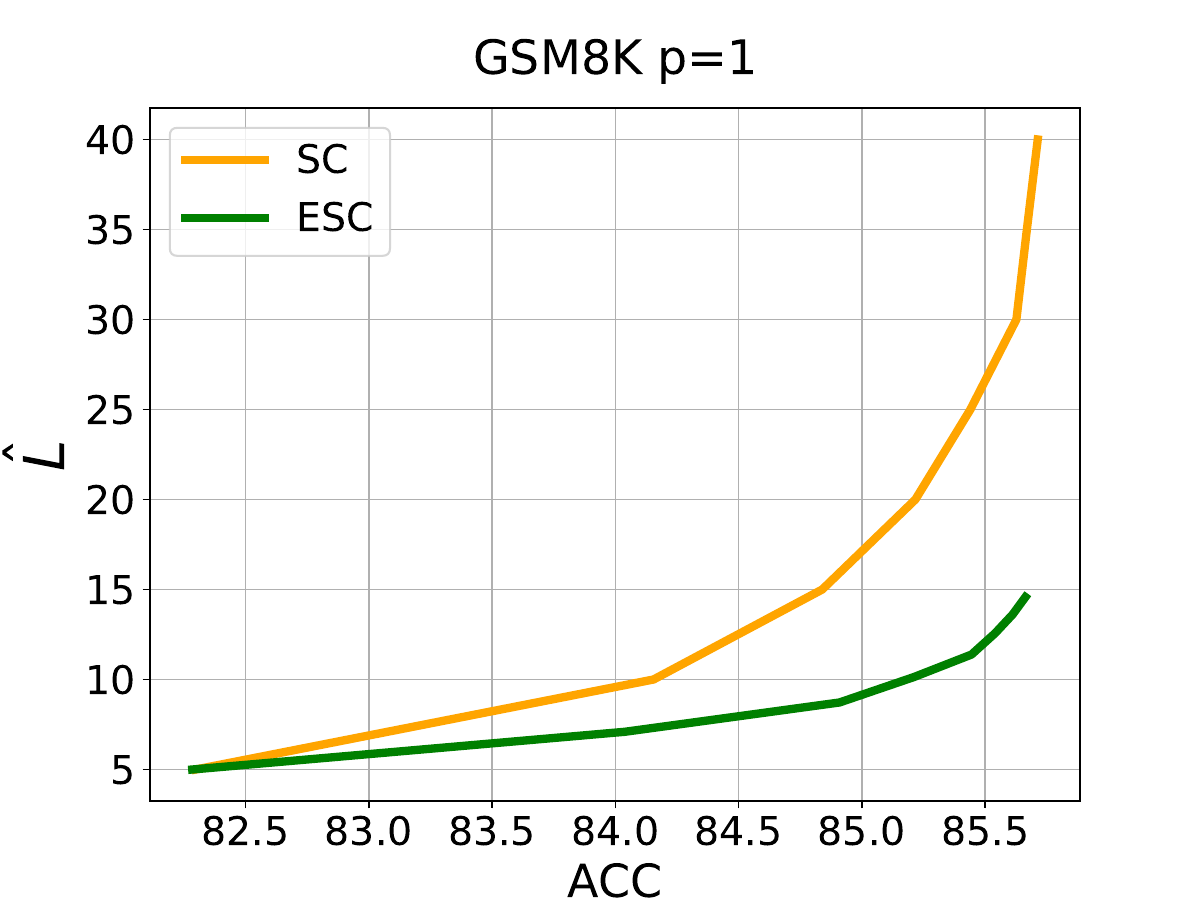}}\hspace{-0.4cm}
    \subfigure{\includegraphics[width=0.267\hsize, height=0.26\hsize]{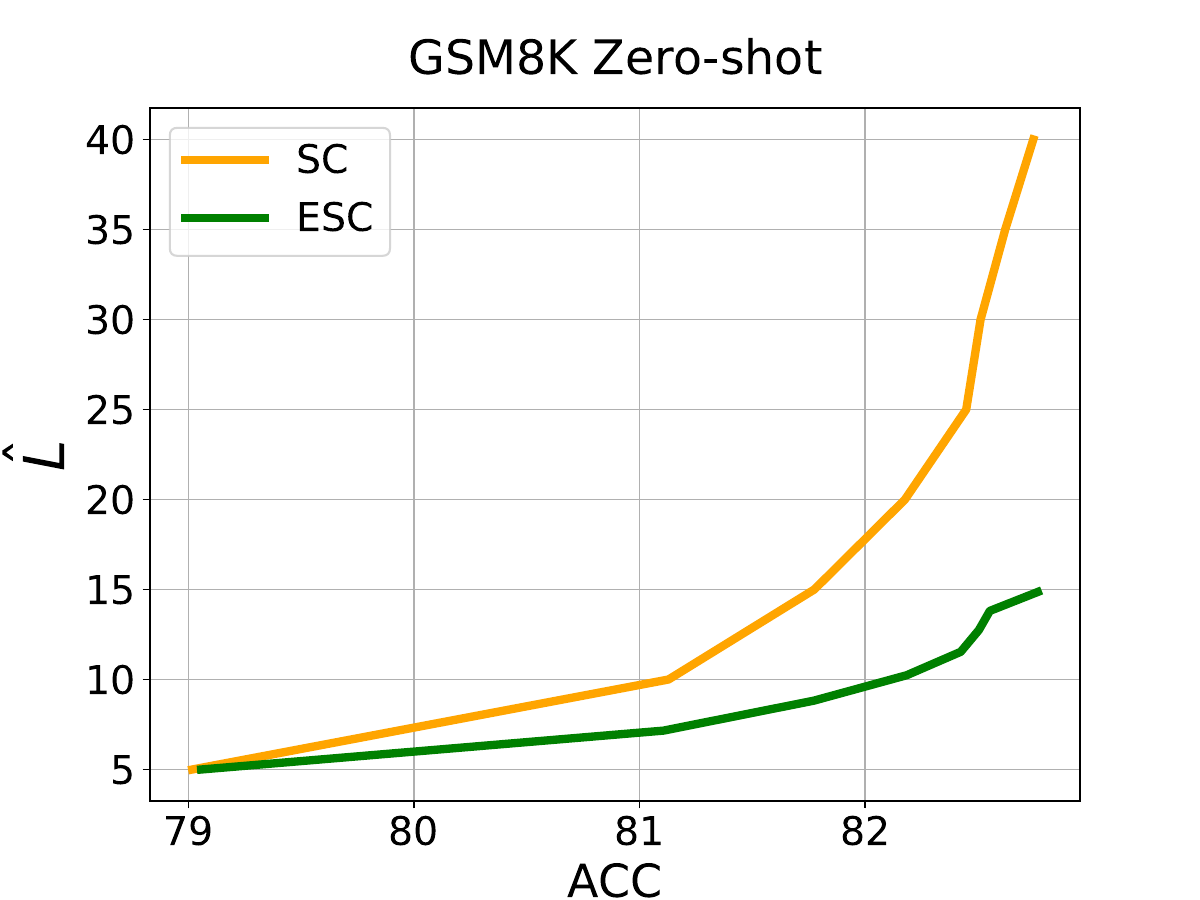}}
    \setlength{\belowcaptionskip}{0pt}
    \caption{Robustness analysis of ESC regarding the sampling temperature $T$, $p$ in nucleus sampling, and zero-shot demonstration on GSM8K with GPT-3.5-Turbo.}
    \label{fig:robust}
\end{figure*}

Additional experiments were conducted to further test the robustness of the proposed ESC, including its robustness to sampling parameters and prompts:
(1) In Figure~\ref{fig:robust} (up) we show how ESC behaves for GSM8K as the decoding sampling temperature increases. Savings are consistent across different generation temperatures.

(2) Figure~\ref{fig:robust} (bottom left corner) shows that ESC is robust to p values for top-p sampling.
(3) Figure~\ref{fig:robust} (bottom right corner) indicates ESC can generalize to zero-shot manner.
(4) Table~\ref{tb:prompts} shows the accuracy of ESC and SC with different groups of demonstrations. We can see that ESC is robust to various demonstrations (see Appendix-Figure~\ref{fig-appendix:robust} for results on StrategyQA).

\begin{table}[th]
    \centering
    \small
    \caption{Reasoning accuracy (\%) and $\hat{L}$ for GSM8K on GPT-3.5-Turbo with different demonstrations. The max sampling size is 40 and window size is 5.}
    \begin{tabular}{c l l l l l}
    \toprule
        Demonstration Groups & 1st & 2nd & 3rd & 4th & 5th  \\ \toprule
      SC  &85.69& 85.56  & 84.80 & 85.63 & 85.24\\
       ESC  &85.67 \tiny \textcolor{red}{(-0.02)}& 85.58  \tiny \textcolor{mygreen}{(0.02)}  & 84.80 \tiny \textcolor{mygreen}{(0.00)}  & 85.64 \tiny \textcolor{mygreen}{(0.01)}& 85.23 \tiny \textcolor{red}{(-0.01)}  \\
        $\hat{L}$  &\textcolor{gray}{14.65} \tiny\textcolor{mygreen}{(-25.35)}& \textcolor{gray}{14.53} \tiny\textcolor{mygreen}{(-25.47)}& \textcolor{gray}{15.64} \tiny\textcolor{mygreen}{(-24.36)}&\textcolor{gray}{14.76} \tiny\textcolor{mygreen}{(-25.24)}&\textcolor{gray}{14.59} \tiny\textcolor{mygreen}{(-25.41)}\\ 
        \bottomrule
    \end{tabular}
    \label{tb:prompts}
\end{table}
\section{Related Work}
\paragraph{Chain-of-thought Reasoning}
Chain-of-thought prompting has been proven to be an effective method of solving complex reasoning problems \citep{COT}. By following the pattern of gradually solving sub-problems, both few-shot CoT \citep{Specialize} and zero-shot CoT \citep{ZeroCOT} are capable of stimulating LLM reasoning abilities. On this basis, Least-to-most prompting \citep{LeastToMost} suggests explicitly splitting the problem and solving them step by step. \citet{PHP} iteratively generating answers and adding the historically generated answers as hints to the context to achieve the final convergence on the answer.

\paragraph{Self Consistency}
Self-consistency \citep{SC} refers to a simple decoding strategy for further improving reasoning performance, leveraging the fact that complex reasoning tasks typically allow for more than one correct reasoning path. \citet{open} extend it for open-ended generation tasks like code generation and text summarization by replacing voting through text similarity matching. \citet{ASC} assign appropriate weights for answer aggregation to achieve adaptive self-consistency. However, all of them require multiple sampling with the pre-set size, which will incur much more computation cost. \citet{AC} introduce an adaptive stopping criterion based on the amount of agreement between the samples so far, but it needs additional data to tune the hyperparameter and is sensitive to threshold. By contrast, ESC has no hyperparamter for stopping criterion and has a additional control scheme to meet the realistic requirements.
\section{Conclusion}
We have introduced a simple yet effective sampling process called early-stopping self-consistency (ESC). By stopping the decoding process with high confident window, ESC greatly reduce the cost of SC without sacrificing performance. A control scheme for ESC is further derivated to dynamically select the performance-cost balance for different tasks and models, which requires no extra prior knowledge of model capabilities and task difficulty. The empirical results show that ESC reduces the actual number of samples of chain-of-thought reasoning by a significant margin on six popular benchmarks, while attaining comparable performances. We also show control scheme for ESC can predict the performance-cost trade-off accurately across various tasks and models. The additional evaluations indicate that ESC can robustly save cost considering different decoding settings and prompts, and even on open-ended generation tasks.

\subsubsection*{Acknowledgments}
This work is supported by Beijing Natural Science Foundation (No.4222037, L181010).

\bibliography{iclr2024_conference}

\begin{thebibliography}{25}
\providecommand{\natexlab}[1]{#1}
\providecommand{\url}[1]{\texttt{#1}}
\expandafter\ifx\csname urlstyle\endcsname\relax
  \providecommand{\doi}[1]{doi: #1}\else
  \providecommand{\doi}{doi: \begingroup \urlstyle{rm}\Url}\fi

\bibitem[Aggarwal et~al.(2023)Aggarwal, Madaan, Yang, and Mausam]{AC}
Pranjal Aggarwal, Aman Madaan, Yiming Yang, and Mausam.
\newblock Let's sample step by step: Adaptive-consistency for efficient
  reasoning with llms.
\newblock \emph{CoRR}, abs/2305.11860, 2023.
\newblock \doi{10.48550/ARXIV.2305.11860}.
\newblock URL \url{https://doi.org/10.48550/arXiv.2305.11860}.

\bibitem[Austin et~al.(2021)Austin, Odena, Nye, Bosma, Michalewski, Dohan,
  Jiang, Cai, Terry, Le, and Sutton]{MBPP}
Jacob Austin, Augustus Odena, Maxwell~I. Nye, Maarten Bosma, Henryk
  Michalewski, David Dohan, Ellen Jiang, Carrie~J. Cai, Michael Terry, Quoc~V.
  Le, and Charles Sutton.
\newblock Program synthesis with large language models.
\newblock \emph{CoRR}, abs/2108.07732, 2021.
\newblock URL \url{https://arxiv.org/abs/2108.07732}.

\bibitem[Bubeck et~al.(2023)Bubeck, Chandrasekaran, Eldan, Gehrke, Horvitz,
  Kamar, Lee, Lee, Li, Lundberg, Nori, Palangi, Ribeiro, and Zhang]{AGI}
S{\'{e}}bastien Bubeck, Varun Chandrasekaran, Ronen Eldan, Johannes Gehrke,
  Eric Horvitz, Ece Kamar, Peter Lee, Yin~Tat Lee, Yuanzhi Li, Scott~M.
  Lundberg, Harsha Nori, Hamid Palangi, Marco~T{\'{u}}lio Ribeiro, and
  Yi~Zhang.
\newblock Sparks of artificial general intelligence: Early experiments with
  {GPT-4}.
\newblock \emph{CoRR}, abs/2303.12712, 2023.
\newblock \doi{10.48550/arXiv.2303.12712}.
\newblock URL \url{https://doi.org/10.48550/arXiv.2303.12712}.

\bibitem[Cobbe et~al.(2021)Cobbe, Kosaraju, Bavarian, Chen, Jun, Kaiser,
  Plappert, Tworek, Hilton, Nakano, Hesse, and Schulman]{GSM8K}
Karl Cobbe, Vineet Kosaraju, Mohammad Bavarian, Mark Chen, Heewoo Jun, Lukasz
  Kaiser, Matthias Plappert, Jerry Tworek, Jacob Hilton, Reiichiro Nakano,
  Christopher Hesse, and John Schulman.
\newblock Training verifiers to solve math word problems.
\newblock \emph{CoRR}, abs/2110.14168, 2021.
\newblock URL \url{https://arxiv.org/abs/2110.14168}.

\bibitem[Fu et~al.(2023)Fu, Peng, Ou, Sabharwal, and Khot]{Specialize}
Yao Fu, Hao Peng, Litu Ou, Ashish Sabharwal, and Tushar Khot.
\newblock Specializing smaller language models towards multi-step reasoning.
\newblock \emph{CoRR}, abs/2301.12726, 2023.
\newblock \doi{10.48550/arXiv.2301.12726}.
\newblock URL \url{https://doi.org/10.48550/arXiv.2301.12726}.

\bibitem[Geva et~al.(2021)Geva, Khashabi, Segal, Khot, Roth, and Berant]{SQA}
Mor Geva, Daniel Khashabi, Elad Segal, Tushar Khot, Dan Roth, and Jonathan
  Berant.
\newblock Did aristotle use a laptop? {A} question answering benchmark with
  implicit reasoning strategies.
\newblock \emph{Trans. Assoc. Comput. Linguistics}, 9:\penalty0 346--361, 2021.
\newblock \doi{10.1162/tacl\_a\_00370}.
\newblock URL \url{https://doi.org/10.1162/tacl\_a\_00370}.

\bibitem[Hendrycks et~al.(2021)Hendrycks, Burns, Kadavath, Arora, Basart, Tang,
  Song, and Steinhardt]{MATH}
Dan Hendrycks, Collin Burns, Saurav Kadavath, Akul Arora, Steven Basart, Eric
  Tang, Dawn Song, and Jacob Steinhardt.
\newblock Measuring mathematical problem solving with the {MATH} dataset.
\newblock In \emph{Proceedings of the Neural Information Processing Systems
  Track on Datasets and Benchmarks 1, NeurIPS Datasets and Benchmarks 2021,
  December 2021, virtual}, 2021.
\newblock URL
  \url{https://datasets-benchmarks-proceedings.neurips.cc/paper/2021/hash/be83ab3ecd0db773eb2dc1b0a17836a1-Abstract-round2.html}.

\bibitem[Holtzman et~al.(2020)Holtzman, Buys, Du, Forbes, and Choi]{topp}
Ari Holtzman, Jan Buys, Li~Du, Maxwell Forbes, and Yejin Choi.
\newblock The curious case of neural text degeneration.
\newblock In \emph{8th International Conference on Learning Representations,
  {ICLR} 2020, Addis Ababa, Ethiopia, April 26-30, 2020}. OpenReview.net, 2020.
\newblock URL \url{https://openreview.net/forum?id=rygGQyrFvH}.

\bibitem[Hosseini et~al.(2014)Hosseini, Hajishirzi, Etzioni, and
  Kushman]{AddSub}
Mohammad~Javad Hosseini, Hannaneh Hajishirzi, Oren Etzioni, and Nate Kushman.
\newblock Learning to solve arithmetic word problems with verb categorization.
\newblock In \emph{Proceedings of the 2014 Conference on Empirical Methods in
  Natural Language Processing, {EMNLP} 2014, October 25-29, 2014, Doha, Qatar,
  {A} meeting of SIGDAT, a Special Interest Group of the {ACL}}, pp.\
  523--533. {ACL}, 2014.
\newblock \doi{10.3115/v1/d14-1058}.
\newblock URL \url{https://doi.org/10.3115/v1/d14-1058}.

\bibitem[Jain et~al.(2023)Jain, Ma, Deoras, and Xiang]{open}
Siddhartha Jain, Xiaofei Ma, Anoop Deoras, and Bing Xiang.
\newblock Self-consistency for open-ended generations.
\newblock \emph{CoRR}, abs/2307.06857, 2023.
\newblock \doi{10.48550/arXiv.2307.06857}.
\newblock URL \url{https://doi.org/10.48550/arXiv.2307.06857}.

\bibitem[Kojima et~al.(2022)Kojima, Gu, Reid, Matsuo, and Iwasawa]{ZeroCOT}
Takeshi Kojima, Shixiang~Shane Gu, Machel Reid, Yutaka Matsuo, and Yusuke
  Iwasawa.
\newblock Large language models are zero-shot reasoners.
\newblock In \emph{NeurIPS}, 2022.
\newblock URL
  \url{http://papers.nips.cc/paper\_files/paper/2022/hash/8bb0d291acd4acf06ef112099c16f326-Abstract-Conference.html}.

\bibitem[Lewkowycz et~al.(2022)Lewkowycz, Andreassen, Dohan, Dyer, Michalewski,
  Ramasesh, Slone, Anil, Schlag, Gutman{-}Solo, Wu, Neyshabur, Gur{-}Ari, and
  Misra]{Minerva}
Aitor Lewkowycz, Anders Andreassen, David Dohan, Ethan Dyer, Henryk
  Michalewski, Vinay~V. Ramasesh, Ambrose Slone, Cem Anil, Imanol Schlag, Theo
  Gutman{-}Solo, Yuhuai Wu, Behnam Neyshabur, Guy Gur{-}Ari, and Vedant Misra.
\newblock Solving quantitative reasoning problems with language models.
\newblock In \emph{NeurIPS}, 2022.
\newblock URL
  \url{http://papers.nips.cc/paper\_files/paper/2022/hash/18abbeef8cfe9203fdf9053c9c4fe191-Abstract-Conference.html}.

\bibitem[Li et~al.(2023)Li, Yuan, Feng, Pan, Sun, Wang, Wang, and Li]{ASC}
Yiwei Li, Peiwen Yuan, Shaoxiong Feng, Boyuan Pan, Bin Sun, Xinglin Wang, Heda
  Wang, and Kan Li.
\newblock Turning dust into gold: Distilling complex reasoning capabilities
  from llms by leveraging negative data.
\newblock \emph{CoRR}, abs/2312.12832, 2023.
\newblock \doi{10.48550/ARXIV.2312.12832}.
\newblock URL \url{https://doi.org/10.48550/arXiv.2312.12832}.

\bibitem[Miao et~al.(2020)Miao, Liang, and Su]{asdiv}
Shen{-}Yun Miao, Chao{-}Chun Liang, and Keh{-}Yih Su.
\newblock A diverse corpus for evaluating and developing english math word
  problem solvers.
\newblock In \emph{Proceedings of the 58th Annual Meeting of the Association
  for Computational Linguistics, {ACL} 2020, Online, July 5-10, 2020}, pp.\
  975--984. Association for Computational Linguistics, 2020.
\newblock \doi{10.18653/v1/2020.acl-main.92}.
\newblock URL \url{https://doi.org/10.18653/v1/2020.acl-main.92}.

\bibitem[OpenAI(2023)]{GPT4}
OpenAI.
\newblock {GPT-4} technical report.
\newblock \emph{CoRR}, abs/2303.08774, 2023.
\newblock \doi{10.48550/arXiv.2303.08774}.
\newblock URL \url{https://doi.org/10.48550/arXiv.2303.08774}.

\bibitem[Papoulis(1990)]{LawofLargeNumber}
Athanasios Papoulis.
\newblock \emph{Probability and statistics}.
\newblock Prentice-Hall, Inc., 1990.

\bibitem[Patel et~al.(2021)Patel, Bhattamishra, and Goyal]{svamp}
Arkil Patel, Satwik Bhattamishra, and Navin Goyal.
\newblock Are {NLP} models really able to solve simple math word problems?
\newblock In \emph{Proceedings of the 2021 Conference of the North American
  Chapter of the Association for Computational Linguistics: Human Language
  Technologies, {NAACL-HLT} 2021, Online, June 6-11, 2021}, pp.\  2080--2094.
  Association for Computational Linguistics, 2021.
\newblock \doi{10.18653/v1/2021.naacl-main.168}.
\newblock URL \url{https://doi.org/10.18653/v1/2021.naacl-main.168}.

\bibitem[Roy \& Roth(2015)Roy and Roth]{multi}
Subhro Roy and Dan Roth.
\newblock Solving general arithmetic word problems.
\newblock In \emph{Proceedings of the 2015 Conference on Empirical Methods in
  Natural Language Processing, {EMNLP} 2015, Lisbon, Portugal, September 17-21,
  2015}, pp.\  1743--1752. The Association for Computational Linguistics, 2015.
\newblock \doi{10.18653/v1/d15-1202}.
\newblock URL \url{https://doi.org/10.18653/v1/d15-1202}.

\bibitem[Talmor et~al.(2019)Talmor, Herzig, Lourie, and Berant]{CSQA}
Alon Talmor, Jonathan Herzig, Nicholas Lourie, and Jonathan Berant.
\newblock Commonsenseqa: {A} question answering challenge targeting commonsense
  knowledge.
\newblock In \emph{Proceedings of the 2019 Conference of the North American
  Chapter of the Association for Computational Linguistics: Human Language
  Technologies, {NAACL-HLT} 2019, Minneapolis, MN, USA, June 2-7, 2019, Volume
  1 (Long and Short Papers)}, pp.\  4149--4158. Association for Computational
  Linguistics, 2019.
\newblock \doi{10.18653/v1/n19-1421}.
\newblock URL \url{https://doi.org/10.18653/v1/n19-1421}.

\bibitem[Touvron et~al.(2023)Touvron, Martin, Stone, Albert, Almahairi, Babaei,
  Bashlykov, Batra, Bhargava, Bhosale, Bikel, Blecher, Canton{-}Ferrer, Chen,
  Cucurull, Esiobu, Fernandes, Fu, Fu, Fuller, Gao, Goswami, Goyal, Hartshorn,
  Hosseini, Hou, Inan, Kardas, Kerkez, Khabsa, Kloumann, Korenev, Koura,
  Lachaux, Lavril, Lee, Liskovich, Lu, Mao, Martinet, Mihaylov, Mishra,
  Molybog, Nie, Poulton, Reizenstein, Rungta, Saladi, Schelten, Silva, Smith,
  Subramanian, Tan, Tang, Taylor, Williams, Kuan, Xu, Yan, Zarov, Zhang, Fan,
  Kambadur, Narang, Rodriguez, Stojnic, Edunov, and Scialom]{llama2}
Hugo Touvron, Louis Martin, Kevin Stone, Peter Albert, Amjad Almahairi, Yasmine
  Babaei, Nikolay Bashlykov, Soumya Batra, Prajjwal Bhargava, Shruti Bhosale,
  Dan Bikel, Lukas Blecher, Cristian Canton{-}Ferrer, Moya Chen, Guillem
  Cucurull, David Esiobu, Jude Fernandes, Jeremy Fu, Wenyin Fu, Brian Fuller,
  Cynthia Gao, Vedanuj Goswami, Naman Goyal, Anthony Hartshorn, Saghar
  Hosseini, Rui Hou, Hakan Inan, Marcin Kardas, Viktor Kerkez, Madian Khabsa,
  Isabel Kloumann, Artem Korenev, Punit~Singh Koura, Marie{-}Anne Lachaux,
  Thibaut Lavril, Jenya Lee, Diana Liskovich, Yinghai Lu, Yuning Mao, Xavier
  Martinet, Todor Mihaylov, Pushkar Mishra, Igor Molybog, Yixin Nie, Andrew
  Poulton, Jeremy Reizenstein, Rashi Rungta, Kalyan Saladi, Alan Schelten, Ruan
  Silva, Eric~Michael Smith, Ranjan Subramanian, Xiaoqing~Ellen Tan, Binh Tang,
  Ross Taylor, Adina Williams, Jian~Xiang Kuan, Puxin Xu, Zheng Yan, Iliyan
  Zarov, Yuchen Zhang, Angela Fan, Melanie Kambadur, Sharan Narang,
  Aur{\'{e}}lien Rodriguez, Robert Stojnic, Sergey Edunov, and Thomas Scialom.
\newblock Llama 2: Open foundation and fine-tuned chat models.
\newblock \emph{CoRR}, abs/2307.09288, 2023.
\newblock \doi{10.48550/arXiv.2307.09288}.
\newblock URL \url{https://doi.org/10.48550/arXiv.2307.09288}.

\bibitem[Wang et~al.(2023)Wang, Wei, Schuurmans, Le, Chi, Narang, Chowdhery,
  and Zhou]{SC}
Xuezhi Wang, Jason Wei, Dale Schuurmans, Quoc~V. Le, Ed~H. Chi, Sharan Narang,
  Aakanksha Chowdhery, and Denny Zhou.
\newblock Self-consistency improves chain of thought reasoning in language
  models.
\newblock In \emph{The Eleventh International Conference on Learning
  Representations, {ICLR} 2023, Kigali, Rwanda, May 1-5, 2023}. OpenReview.net,
  2023.
\newblock URL \url{https://openreview.net/pdf?id=1PL1NIMMrw}.

\bibitem[Wei et~al.(2022)Wei, Wang, Schuurmans, Bosma, Ichter, Xia, Chi, Le,
  and Zhou]{COT}
Jason Wei, Xuezhi Wang, Dale Schuurmans, Maarten Bosma, Brian Ichter, Fei Xia,
  Ed~H. Chi, Quoc~V. Le, and Denny Zhou.
\newblock Chain-of-thought prompting elicits reasoning in large language
  models.
\newblock In \emph{NeurIPS}, 2022.
\newblock URL
  \url{http://papers.nips.cc/paper\_files/paper/2022/hash/9d5609613524ecf4f15af0f7b31abca4-Abstract-Conference.html}.

\bibitem[Yao et~al.(2007)Yao, Rosasco, and Caponnetto]{es}
Yuan Yao, Lorenzo Rosasco, and Andrea Caponnetto.
\newblock On early stopping in gradient descent learning.
\newblock \emph{Constructive Approximation}, 26:\penalty0 289--315, 2007.

\bibitem[Zheng et~al.(2023)Zheng, Liu, Xie, Li, and Li]{PHP}
Chuanyang Zheng, Zhengying Liu, Enze Xie, Zhenguo Li, and Yu~Li.
\newblock Progressive-hint prompting improves reasoning in large language
  models.
\newblock \emph{CoRR}, abs/2304.09797, 2023.
\newblock \doi{10.48550/arXiv.2304.09797}.
\newblock URL \url{https://doi.org/10.48550/arXiv.2304.09797}.

\bibitem[Zhou et~al.(2023)Zhou, Sch{\"{a}}rli, Hou, Wei, Scales, Wang,
  Schuurmans, Cui, Bousquet, Le, and Chi]{LeastToMost}
Denny Zhou, Nathanael Sch{\"{a}}rli, Le~Hou, Jason Wei, Nathan Scales, Xuezhi
  Wang, Dale Schuurmans, Claire Cui, Olivier Bousquet, Quoc~V. Le, and Ed~H.
  Chi.
\newblock Least-to-most prompting enables complex reasoning in large language
  models.
\newblock In \emph{The Eleventh International Conference on Learning
  Representations, {ICLR} 2023, Kigali, Rwanda, May 1-5, 2023}. OpenReview.net,
  2023.
\newblock URL \url{https://openreview.net/pdf?id=WZH7099tgfM}.

\end{thebibliography}
\bibliographystyle{iclr2024_conference}

\clearpage
\appendix
\section{Appendix}

\subsection{Derivation of the expectation of sampling cost}
\label{appendix:derivation}
Given the probability of stopping sampling in each observation window as $\hat{P}_{stop}$, the probability of continue sampling is $1-\hat{P}_{stop}$. Therefore, the probability of early-stop occurring after $j$ windows is $\hat{P}_{stop}\times\mathrm{pow}(1-\hat{P}_{stop},j)$. On this basis, the expected sampling cost in the first $L//w-1$ windows can be denoted as:
\begin{equation}
\small
\mathbb{E}(\hat{L_1}) = \mathbb{E}_{\hat{P}\in\mathcal{M}(\sD)} \sum_{j=0}^{L//w-1} (\hat{P}_{stop}\times\mathrm{pow}(1-\hat{P}_{stop},j)\times j \times w)
\label{eq:appendix-1}
\end{equation}
If no early-stop occurs after observing $L//w-1$ windows, then the maximum sampling size $L$ is reached upon the next observation window. The expected sampling cost of this part can be denoted as:
\begin{equation}
\small
\mathbb{E}(\hat{L_2}) = \mathbb{E}_{\hat{P}\in\mathcal{M}(\sD)} \sum_{j=0}^{L//w-1} \mathrm{pow}(1-\hat{P}_{stop}, L//w) \times L
\label{eq:appendix-2}
\end{equation}
Including $w_0$, the overall expected sampling cost is:
\begin{equation}
\small
\mathbb{E}(\hat{L}) = \mathbb{E}_{\hat{P}\in\mathcal{M}(\sD)} \sum_{j=0}^{L//w-1} [(\hat{P}_{stop}\times\mathrm{pow}(1-\hat{P}_{stop},j)\times j \times w) + \mathrm{pow}(1-\hat{P}_{stop}, L//w) \times L] + w_0
\label{eq:appendix-3}
\end{equation}

\subsection{Comparison with Adaptive-Consistency (AC)}
 \citet{AC} introduce an adaptive stopping criterion based on the amount of agreement between the samples so far, but it needs additional data to tune the hyperparameter and is sensitive to threshold. But for proposed ESC, the stopping criterion needs no hyperparameter due to the most conservative strategy to maintain the performance, i.e., all the answers within a window are same. Thus ESC can be conducted directly for different tasks and models, without any validation set. 
 
 Another drawback of AC is generating samples step by step, which means each sample requires one input. Considering the demonstrations for in-context learning (usually 8 examples) have a lot of tokens, it will cost quite a portion of the budget. By contrast, ESC generates samples in multiple sampling windows, thus samples within one window can share the same input. Table~\ref{tb:AC} shows that ESC can get higher accuracy with less sampling cost comparing with AC.
 
\begin{table}[th]
\renewcommand{\arraystretch}{1.2}
\centering
\small
\setlength{\tabcolsep}{2.5pt} 
\caption{Prompt token count (tokens/item, denoted as \#prompt), completion token count (tokens/item, denoted as \# completion), average sampling cost (\$/item) and accuracy (\%) resulting from different methods. The best and second-best results are emphasized with bold and underline. } 
\begin{tabular}{ccccccp{0.001pt}cccc}
\toprule
\multicolumn{1}{c}{\multirow{2}{*}{Datasets}}&\multicolumn{1}{c}{\multirow{2}{*}{Method}} & \multicolumn{4}{c}{GPT-3.5 Turbo} && \multicolumn{4}{c}{GPT-4}  \\ \cline{3-6}\cline{8-11}
\multicolumn{1}{c}{} &\multicolumn{1}{c}{}&\# prompt&\# completion&Cost& Acc&&\# prompt&\# completion&Cost& Acc   \\ \midrule
\multirow{3}{*}{GSM8K}&  SC    & \ \ 496.9 & 2909.0 & \underline{0.0084} & \textbf{85.69}&&  \ \ 495.9&2646.5&0.2316&\textbf{89.29}\\
      &  AC    & 4930.3 & \ \ 813.2 & 0.0087&85.66 && 2927.6&\ \ 412.5&\underline{0.1501}&89.28\\
      &  ESC    & 1469.2 & 1220.8 & \textbf{0.0052}&\underline{85.67} & & \ \ 793.0&\ \ 564.8&\textbf{0.0769}&\textbf{89.29}\\\midrule
\multirow{3}{*}{CSQA}&  SC    & \ \ 428.7 & 1382.9 & \underline{0.0043}&\textbf{78.10} & & \ \ 427.8&1474.6&\underline{0.1351}&\textbf{87.18}\\
      &  AC    & 3522.3 & \ \ 294.2 & 0.0055&78.05 & & 2844.5&\ \ 271.5&0.1355& 87.15\\
      &  ESC    & 1003.2 & \ \ 420.9 & \textbf{0.0024}&\textbf{78.10} & & \ \ 794.6&\ \ 384.0&\textbf{0.0625}&\textbf{87.18}\\\midrule
\multirow{3}{*}{Letter}&  SC    & \ \ 166.0 & 1463.1 & 0.0041&\textbf{83.21} & & \ \ 165.0&1396.9&0.1184&\textbf{95.00}\\
      &  AC    & 1057.5 & \ \ 233.6 & \underline{0.0020}&\underline{83.15} & & \ \ 791.7&\ \ 167.5& \underline{0.0451}&94.97\\
          &  ESC    & \ \ 293.0 & \ \ 323.7 & \textbf{0.0012}&\underline{83.15} & & \ \ 208.7&\ \ 220.7&\textbf{0.0260}&\underline{94.98}\\
\bottomrule
\end{tabular}
\label{tb:AC}
\end{table}

\subsection{Comparisons between SC and ESC when L is relatively small (10 and 20)}
we conduct experiments with smaller sampling sizes $L$. From the results in Table~\ref{tb:length_small}, we found that ESC can consistently save sampling costs in such scenarios, while achieving improvement compared to SC under the same costs.

\begin{table}[th]
    \centering
    \small
    \setlength{\tabcolsep}{2.5pt}
    \caption{Reasoning accuracy (\%) and $\hat{L}$ with various max sampling size $L$. The window size is 5.}
    \begin{tabular}{c c l l l l l l l l}
    \toprule
        \multicolumn{1}{c}{\multirow{2}{*}{Model}}&\multicolumn{1}{c}{\multirow{2}{*}{Method}}  & \multicolumn{2}{c}{CSQA}&&\multicolumn{2}{c}{GSM8K}&&\multicolumn{2}{c}{Letter}\\ \cline{3-4}\cline{6-7} \cline{9-10} 
        \multicolumn{1}{c}{} &\multicolumn{1}{c}{}&$L=10$&$L=20$&&$L=10$&$L=20$&&$L=10$&$L=20$\\
        \toprule
      \multirow{4}{*}{GPT-4} &  SC  & 86.49  & 86.83 && 89.16 & 89.24&&94.66 & 94.79 \\
       &  ESC   & 86.49  \tiny \textcolor{mygreen}{(0.00)}  & 86.84 \tiny \textcolor{mygreen}{(+0.01)}  && 89.16  \tiny \textcolor{mygreen}{(0.00)}& 89.24 \tiny \textcolor{mygreen}{(0.00)}&& 94.66  \tiny \textcolor{mygreen}{(0.00)}& 94.80 \tiny \textcolor{mygreen}{(+0.01)}   \\
      &  $\hat{L}$  & \ \ \textcolor{gray}{5.92} \tiny\textcolor{mygreen}{(-4.08)}& \ \ \textcolor{gray}{7.27} \tiny\textcolor{mygreen}{(-12.73)}&&\ \ \textcolor{gray}{5.79} \tiny\textcolor{mygreen}{(-4.21)}&\ \ \textcolor{gray}{6.76} \tiny\textcolor{mygreen}{(-13.24)}&&\ \ \textcolor{gray}{5.57}\tiny\textcolor{mygreen}{(-4.43)}&\ \ \textcolor{gray}{5.92}\tiny\textcolor{mygreen}{(-14.08)} \\
      &  $\hat{L}$-SC  & 86.07 \tiny\textcolor{mygreen}{(-0.42)}& 86.31 \tiny\textcolor{mygreen}{(-0.52)}&&88.95 \tiny\textcolor{mygreen}{(-0.21)}&89.02 \tiny\textcolor{mygreen}{(-0.22)}&&94.58\tiny\textcolor{mygreen}{(-0.08)}&94.58\tiny\textcolor{mygreen}{(-0.21)}
      \\ \midrule
      \multirow{4}{*}{\makecell[c]{GPT-3.5\\Turbo}} &  SC  & 77.63  & 77.93 && 84.10 & 85.15&&82.80&83.00 \\
       &  ESC   & 77.63  \tiny \textcolor{mygreen}{(0.00)}  & 77.91 \tiny \textcolor{red}{(-0.02)}  && 84.10  \tiny \textcolor{mygreen}{(0.00)}& 85.15 \tiny \textcolor{mygreen}{(0.00)}&& 82.81  \tiny \textcolor{mygreen}{(+0.01)}& 82.98 \tiny \textcolor{red}{(-0.02)}  \\
      &  $\hat{L}$  & \ \ \textcolor{gray}{6.47} \tiny\textcolor{mygreen}{(-3.53)}& \ \ \textcolor{gray}{8.60} \tiny\textcolor{mygreen}{(-11.40)}&&\ \ \textcolor{gray}{7.09} \tiny\textcolor{mygreen}{(-2.91)}&\textcolor{gray}{10.14} \tiny\textcolor{mygreen}{(-9.86)}&&\ \ \textcolor{gray}{6.14}\tiny\textcolor{mygreen}{(-3.86)}&\ \ \textcolor{gray}{7.36}\tiny\textcolor{mygreen}{(-12.64)} \\
      &  $\hat{L}$-SC  & 77.39 \tiny\textcolor{mygreen}{(-0.24)}& 77.55 \tiny\textcolor{mygreen}{(-0.38)}&&83.31 \tiny\textcolor{mygreen}{(-0.79)}&84.10 \tiny\textcolor{mygreen}{(-1.05)}&&82.45\tiny\textcolor{mygreen}{(-0.35)}&82.65\tiny\textcolor{mygreen}{(-0.35)}
      \\ \midrule
      \multirow{4}{*}{Llama2-7b} &  SC  & 66.89  & 67.25 && 20.74 & 21.32&&21.21&22.49 \\
       &  ESC   & 66.89  \tiny \textcolor{mygreen}{(0.00)}  & 67.26 \tiny \textcolor{mygreen}{(+0.01)}  && 20.74  \tiny \textcolor{mygreen}{(0.00)}& 21.32 \tiny \textcolor{mygreen}{(0.00)}&& 21.21  \tiny \textcolor{mygreen}{(0.00)}& 22.48 \tiny \textcolor{red}{(-0.01)}   \\
      &  $\hat{L}$  & \ \ \textcolor{gray}{6.59} \tiny\textcolor{mygreen}{(-3.41)}& \ \ \textcolor{gray}{8.78} \tiny\textcolor{mygreen}{(-11.22)}&&\ \ \textcolor{gray}{8.99} \tiny\textcolor{mygreen}{(-1.01)}&\textcolor{gray}{17.06} \tiny\textcolor{mygreen}{(-2.94)}&&\ \ \textcolor{gray}{9.67}\tiny\textcolor{mygreen}{(-0.33)}&\textcolor{gray}{18.47}\tiny\textcolor{mygreen}{(-1.53)} \\
      &  $\hat{L}$-SC  & 66.80 \tiny\textcolor{mygreen}{(-0.09)}& 66.84 \tiny\textcolor{mygreen}{(-0.41)}&&20.59 \tiny\textcolor{mygreen}{(-0.15)}&21.21 \tiny\textcolor{mygreen}{(-0.11)}&&21.21\tiny\textcolor{mygreen}{(-0.00)}&22.32\tiny\textcolor{mygreen}{(-0.17)}
      \\ 
      \bottomrule
    \end{tabular}
    \label{tb:length_small}
\end{table}

\subsection{Discussion on the choice of introducing observation window for the design of early-stopping strategy}
\label{appendix:discussion}

\begin{figure*}[!htb]
    \centering
    \subfigure{\includegraphics[width=0.32\hsize, height=0.25\hsize]{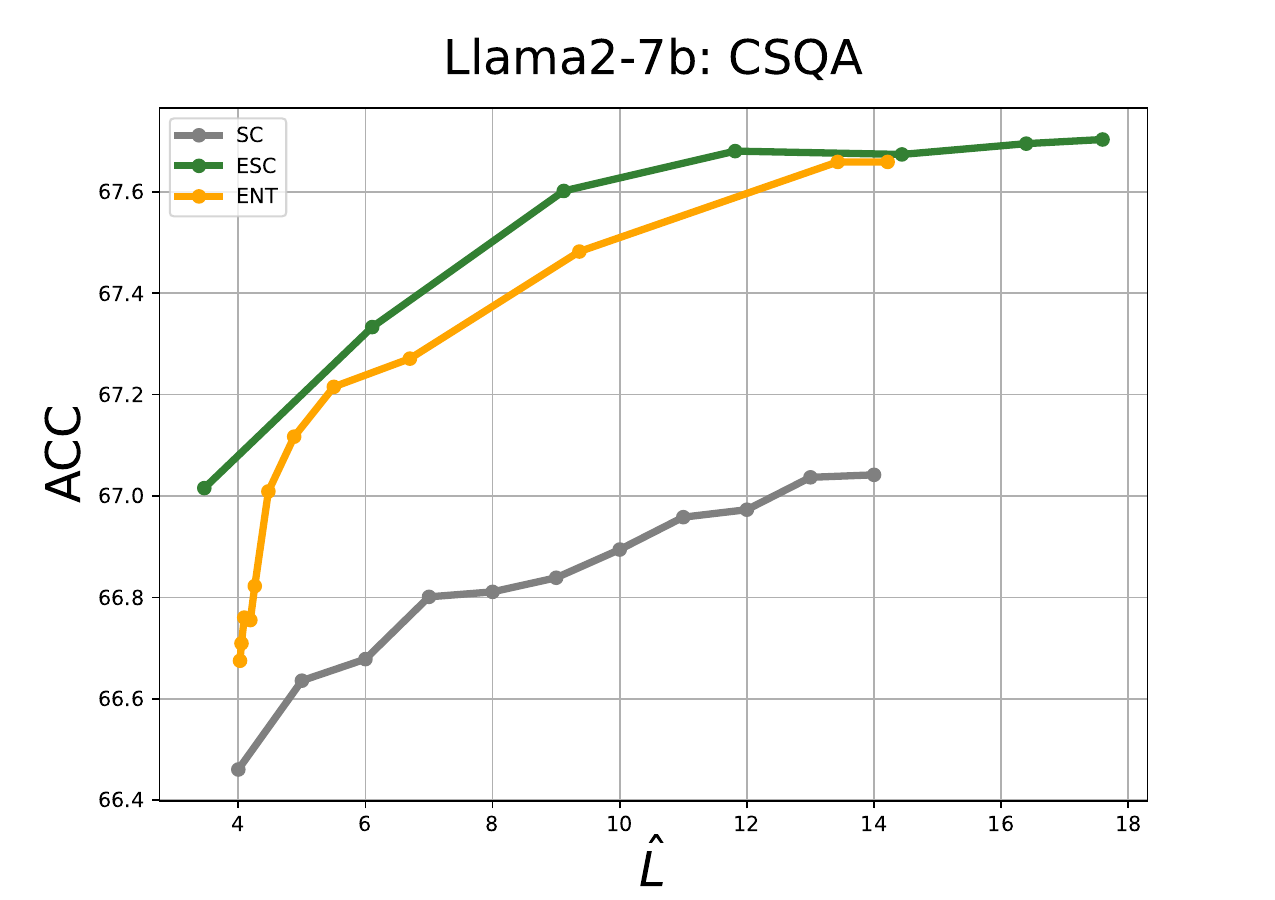}} \hspace{-0.50cm}
    \subfigure{\includegraphics[width=0.32\hsize, height=0.25\hsize]{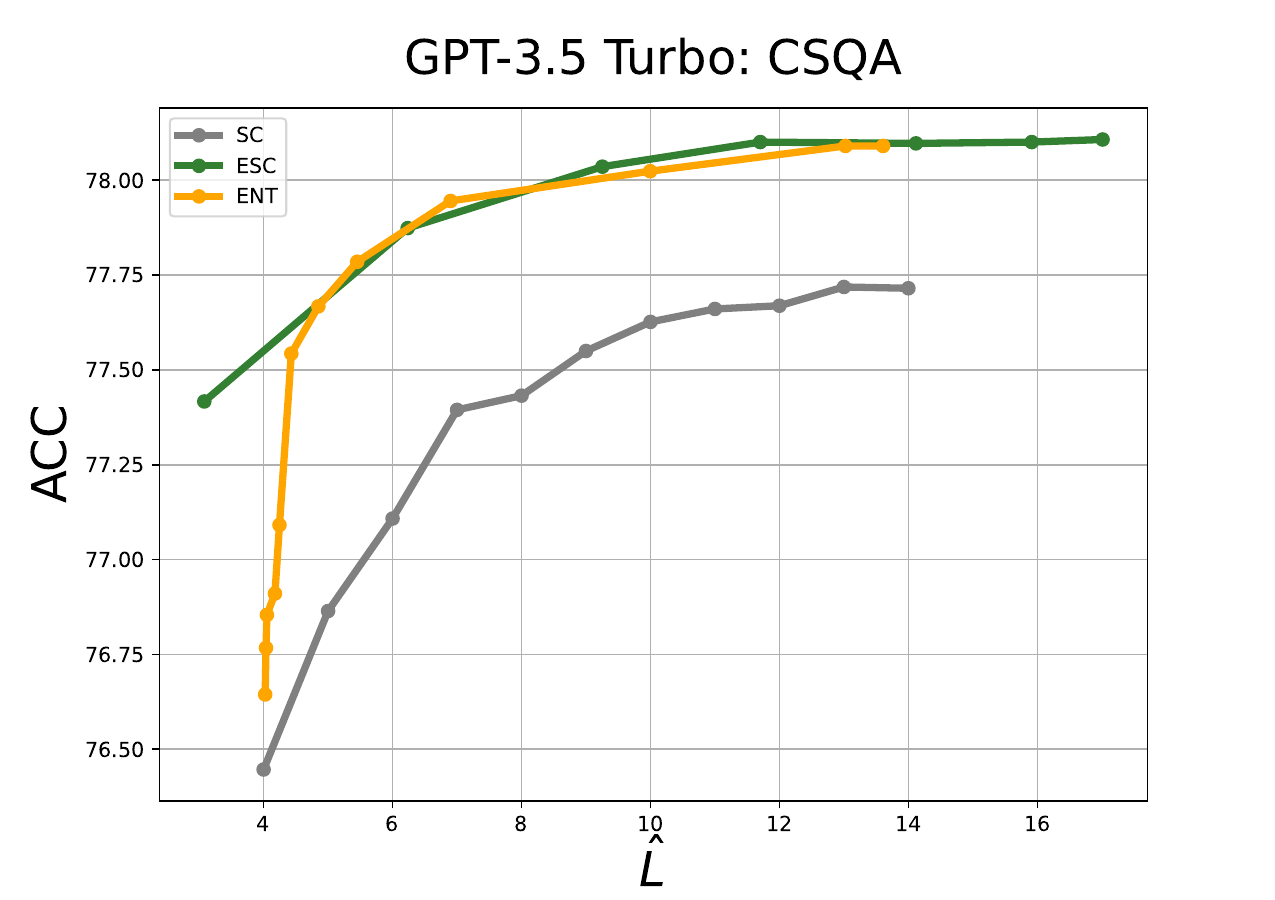} }\hspace{-0.47cm}
    \subfigure{\includegraphics[width=0.32\hsize, height=0.25\hsize]{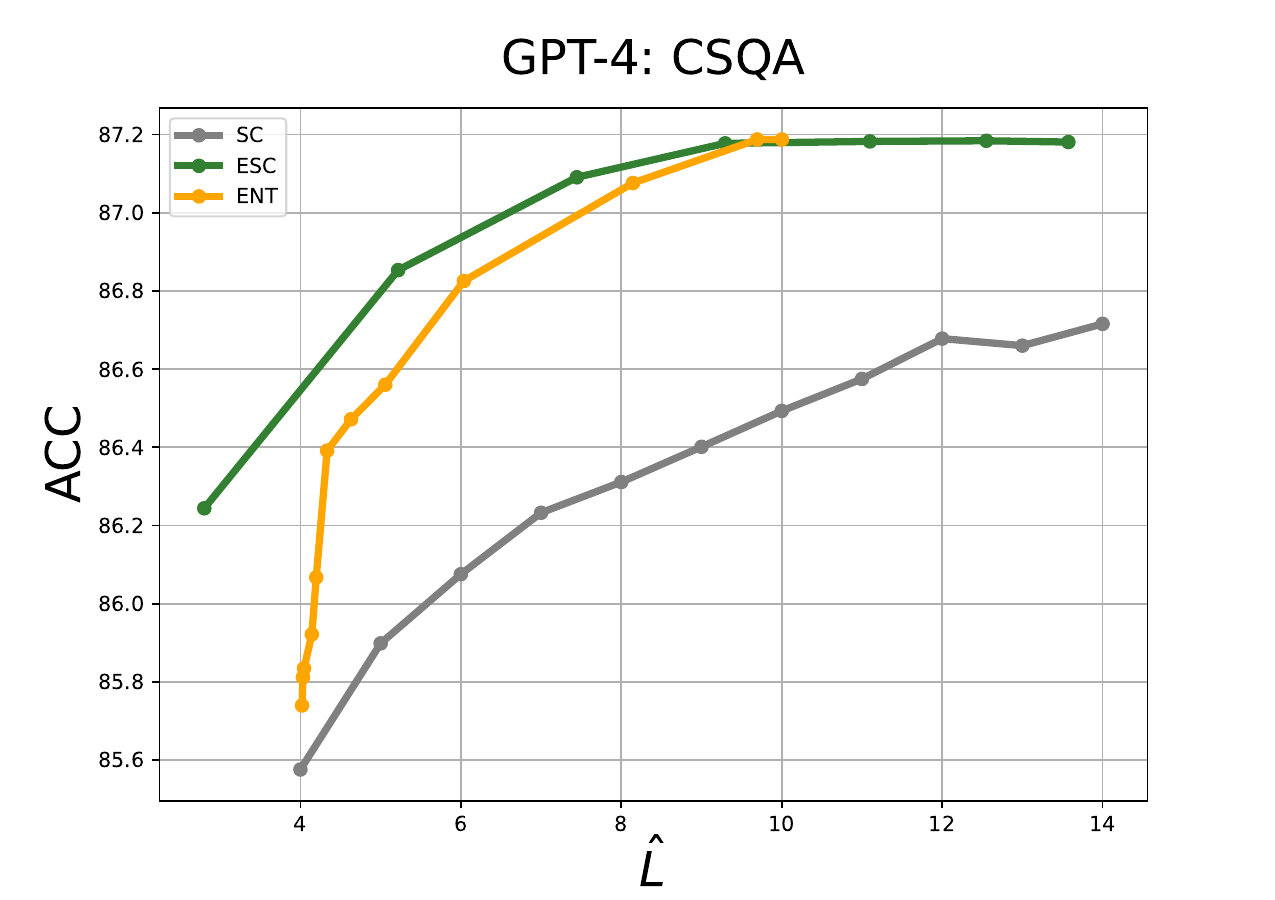}}\hspace{-0.4cm}
    \setlength{\belowcaptionskip}{0pt}
    \caption{Comparisons between SC, ESC and ENT(using entropy to determine the cut-off point) on CSQA with different models.}
    \label{fig:appendix-ent}
\end{figure*}

We design the stopping strategy with the introduction of window for the following two reasons. 
Firstly, we break the sampling process only if all of the samples in the latest window are consistent, thus avoiding any hyper-parameter. If sample one by one and stop based on the observation of the sampled samples, obviously we cannot adopt such a strict truncation condition. In this case, we need to introduce a certain statistic and its corresponding threshold (hyper-parameter), which is hard to be determined in prior. 
Secondly, we have actually considered using the normalized entropy of the sampled samples as the statistical value for cut-off. As shown in Figure~\ref{fig:appendix-ent} , we found that this method not only has the hyper-parameter problem mentioned above, but also has no advantage over ESC in terms of performance-cost trade-off. We believe this is because examining the cut-off point after each single sampling is too frequent, introducing greater randomness. This makes the model more likely to early-stop without sufficient sampling.

\subsection{The Comparisons between ESC and Other Strong Reasoning Baselines.}
ESC achieves better results compared to SC with the same overhead by saving sampling costs while almost not sacrificing performance. From the perspective of performance improvement, we are curious about the comparison results between ESC and advanced reasoning methods that focus on performance enhancement. For this purpose, we compared ESC with Progressive-Hint Prompting(PHP) \citep{PHP} in terms of both performance and overhead. PHP improves performance by approaching answers through the stepwise generation of clues. As shown in Table~\ref{tb:php}, we found that ESC outperformed CoT by 9.84\%, higher than PHP's 9.25\%. Additionally, ESC incurs less sampling overhead. PHP, due to the need for multiple inputs to the model with previously enhanced prompts, has a larger overhead in the \# prompt part. Overall, ESC achieves better performance with lower sampling costs.

\begin{table}[th]
    \centering
    \small
    \caption{Prompt token count (tokens/item, denoted as \#prompt), completion token count (tokens/item, denoted as \# completion), average sampling cost (\$/item) and accuracy (\%) comparison between CoT, PHP \citep{PHP} and ESC on GSM8K with GPT-3.5-Turbo. Max sampling size of ESC is 40. CoT and PHP apply greedy search as sampling strategy.}
    \begin{tabular}{c l l l l}
    \toprule
        Method & \# prompt & \# completion  &Sampling Cost& Accuracy  \\\toprule
        CoT &496.9 & 72.7 & 0.0006 & 75.83 \\
      PHP  &6552.0 & 360.8 & 0.0072  & 85.08 \\
       ESC  &1469.2 & 1220.8& 0.0052 & 85.67    \\
        \bottomrule
    \end{tabular}
    \label{tb:php}
\end{table}

\subsection{The Orthogonality of ESC and Other Strong Reasoning Baseline.}
To assess the orthogonality of ESC with other advanced reasoning methods, we applied ESC on PHP. As shown in Table~\ref{tb:php0}, PHP w. ESC achieved similar performance with PHP w. SC while significantly reduced sampling overhead. This indicates that existing advanced methods can significantly reduce costs by applying ESC with almost no sacrifice in performance.

\begin{table}[th]
    \centering
    \small
    \caption{Reasoning accuracy (\%) and $\hat{L}$ for GSM8K on GPT-3.5-Turbo with Progressive-Hint Prompting(PHP) \citep{PHP}. The window size is 5.}
    \begin{tabular}{c l l l l}
    \toprule
        Max sampling size & \ \ 10 & \ \ 20 & \ \ 30 & \ \ 40  \\ \toprule
      PHP w. SC  &86.32& 86.64  & 86.76 & 87.00 \\
       PHP w. ESC  &86.32 \tiny \textcolor{mygreen}{(0.00)}& 86.62  \tiny \textcolor{red}{(-0.02)}  & 86.77 \tiny \textcolor{mygreen}{(+0.01)}  & 86.98 \tiny \textcolor{red}{(-0.02)}  \\
        $\hat{L}$  &\ \ \textcolor{gray}{6.15} \tiny\textcolor{mygreen}{(-3.85)}& \ \ \textcolor{gray}{7.83} \tiny\textcolor{mygreen}{(-12.17)}& \ \ \textcolor{gray}{9.15} \tiny\textcolor{mygreen}{(-20.85)}&\textcolor{gray}{10.26} \tiny\textcolor{mygreen}{(-29.74)}\\ 
        $\hat{L}$-PHP w. SC  &86.02 \tiny\textcolor{mygreen}{(-0.30)}& 86.29 \tiny\textcolor{mygreen}{(-0.35)}& 86.32 \tiny\textcolor{mygreen}{(-0.44)}&86.35 \tiny\textcolor{mygreen}{(-0.65)}\\ 
        \bottomrule
    \end{tabular}
    \label{tb:php0}
\end{table}

\subsection{Additional Results}
\begin{figure*}[!htb]
    \centering

    \subfigure{\includegraphics[width=0.267\hsize, height=0.23\hsize]{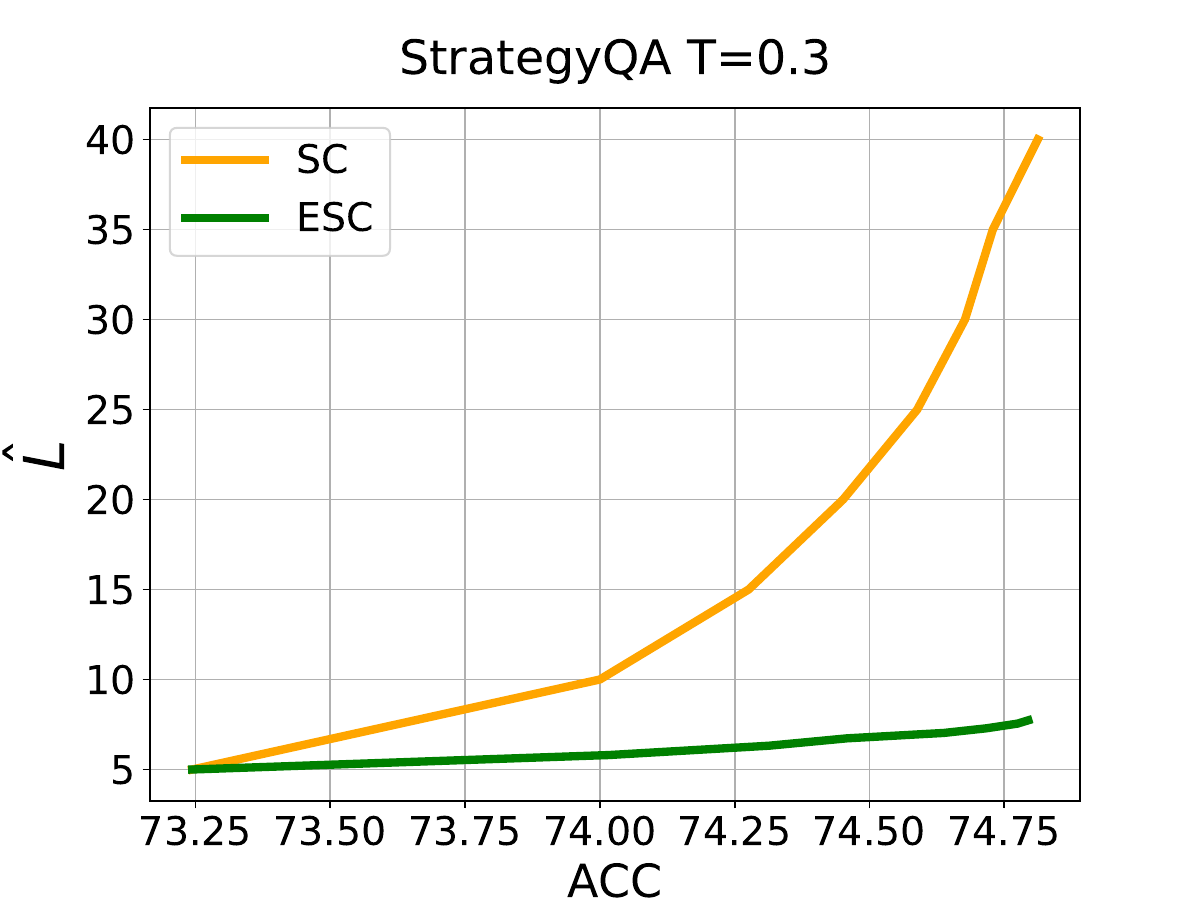}} \hspace{-0.50cm}
    \subfigure{\includegraphics[width=0.267\hsize, height=0.23\hsize]{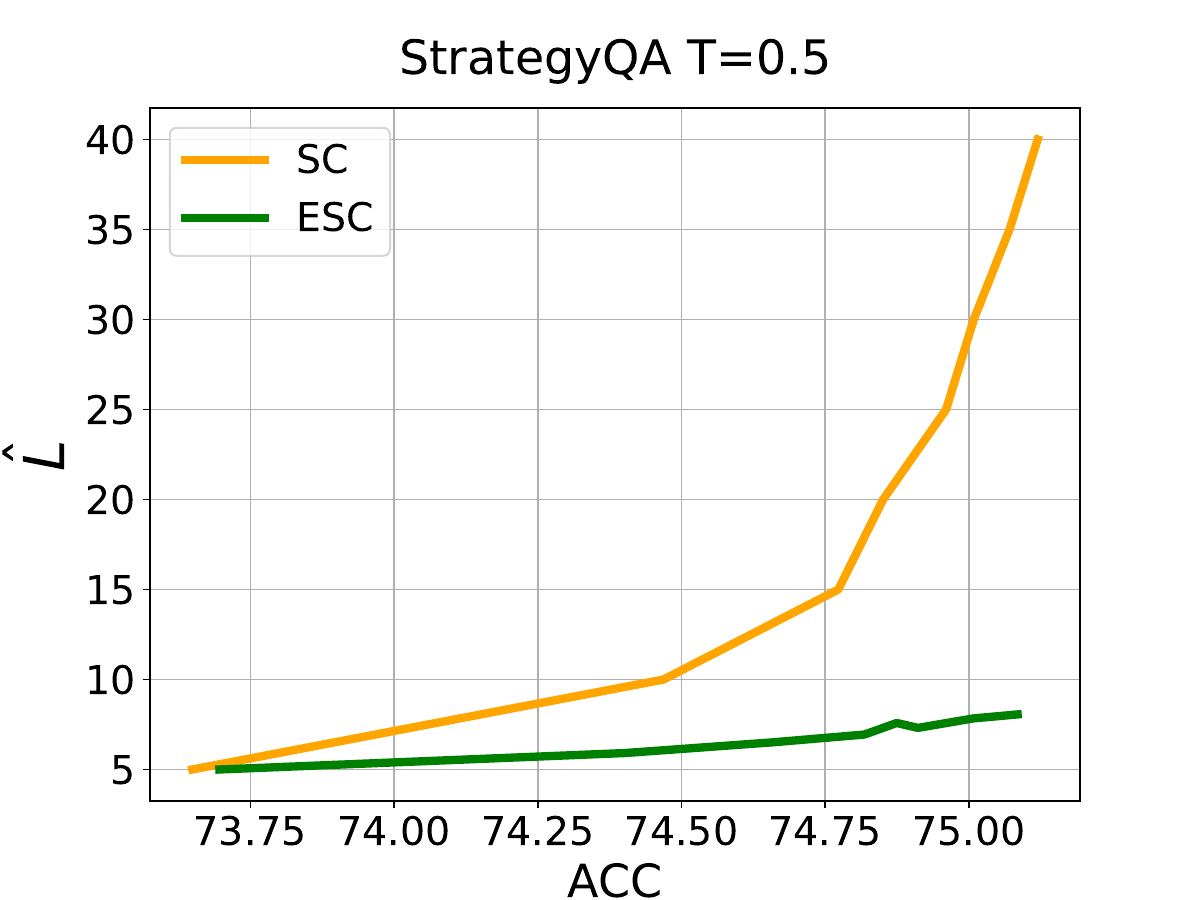}} \hspace{-0.47cm}
    \subfigure{\includegraphics[width=0.267\hsize, height=0.23\hsize]{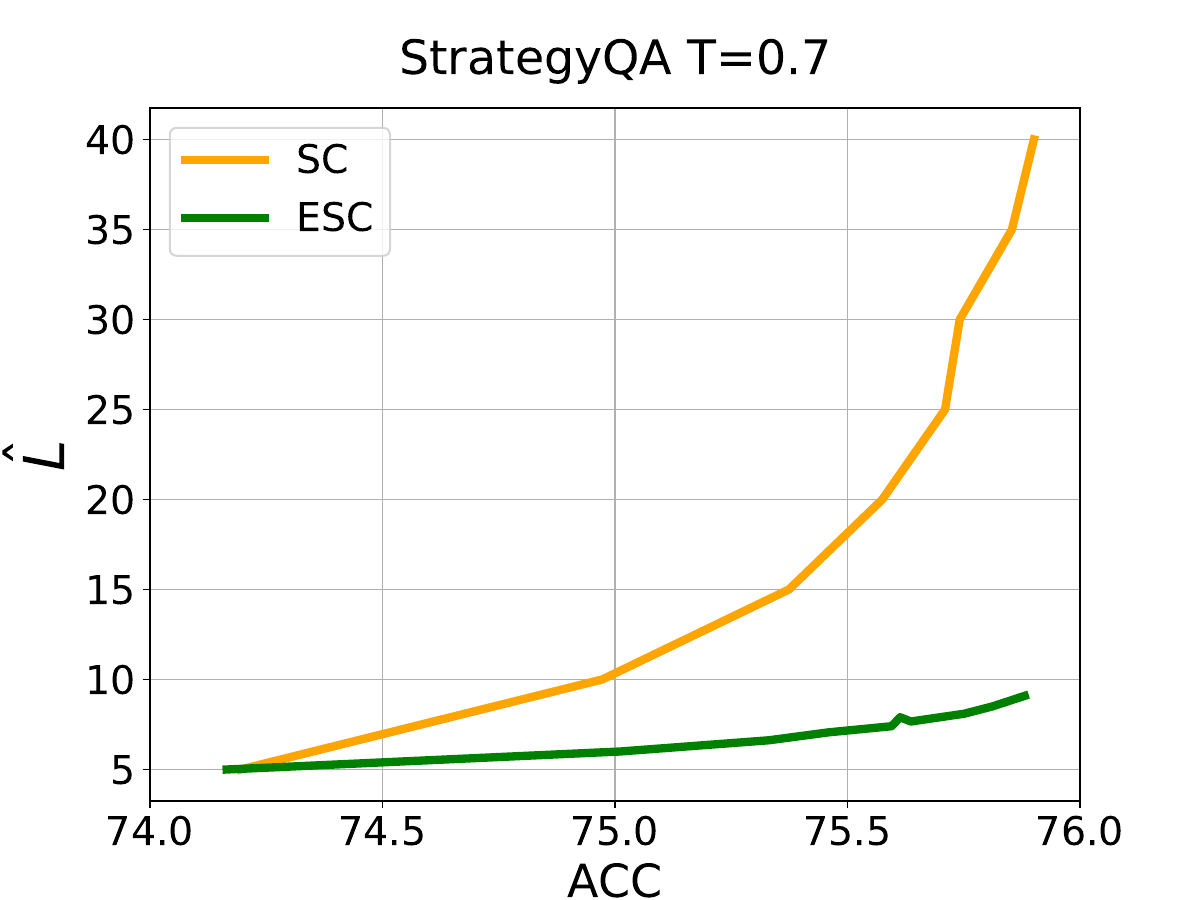}}\hspace{-0.4cm}
    \subfigure{\includegraphics[width=0.267\hsize, height=0.23\hsize]{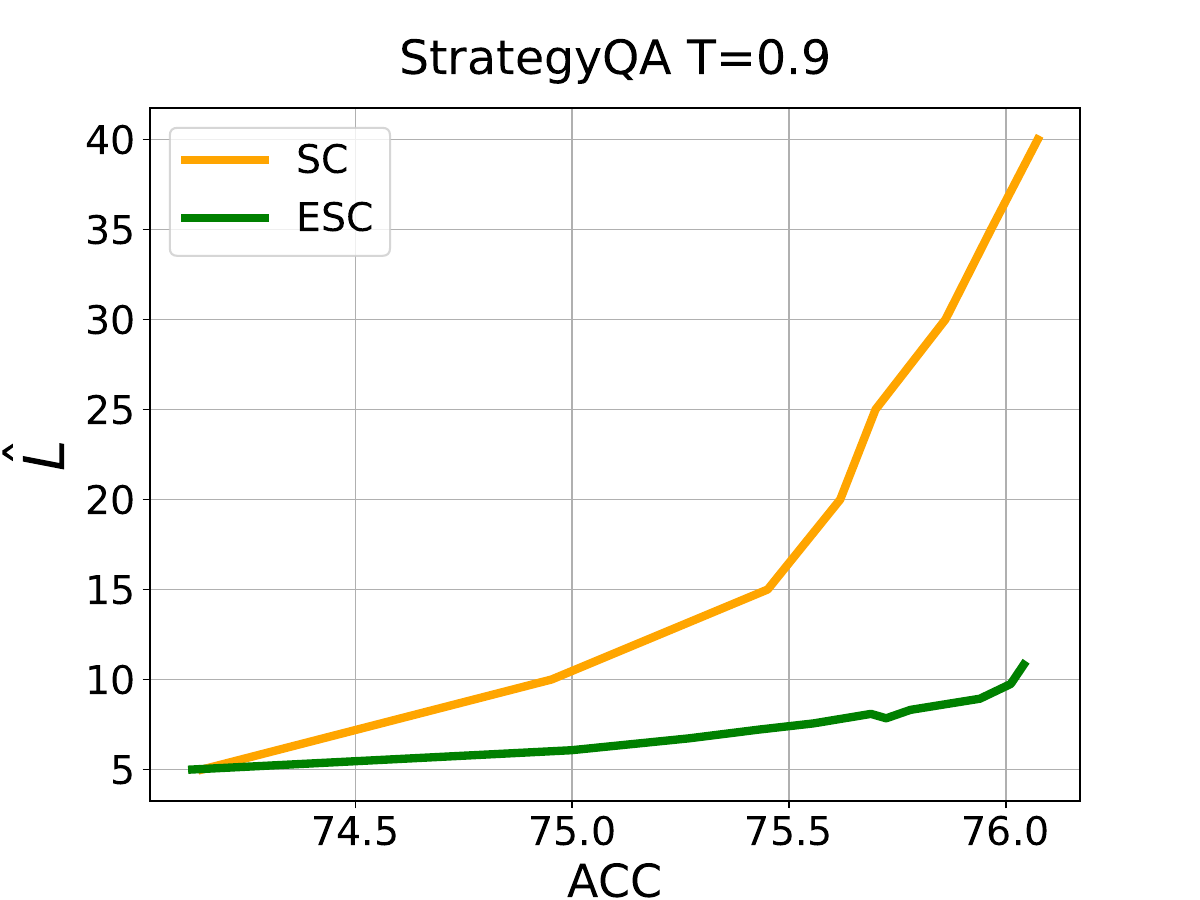}}
    \subfigure{\includegraphics[width=0.267\hsize, height=0.23\hsize]{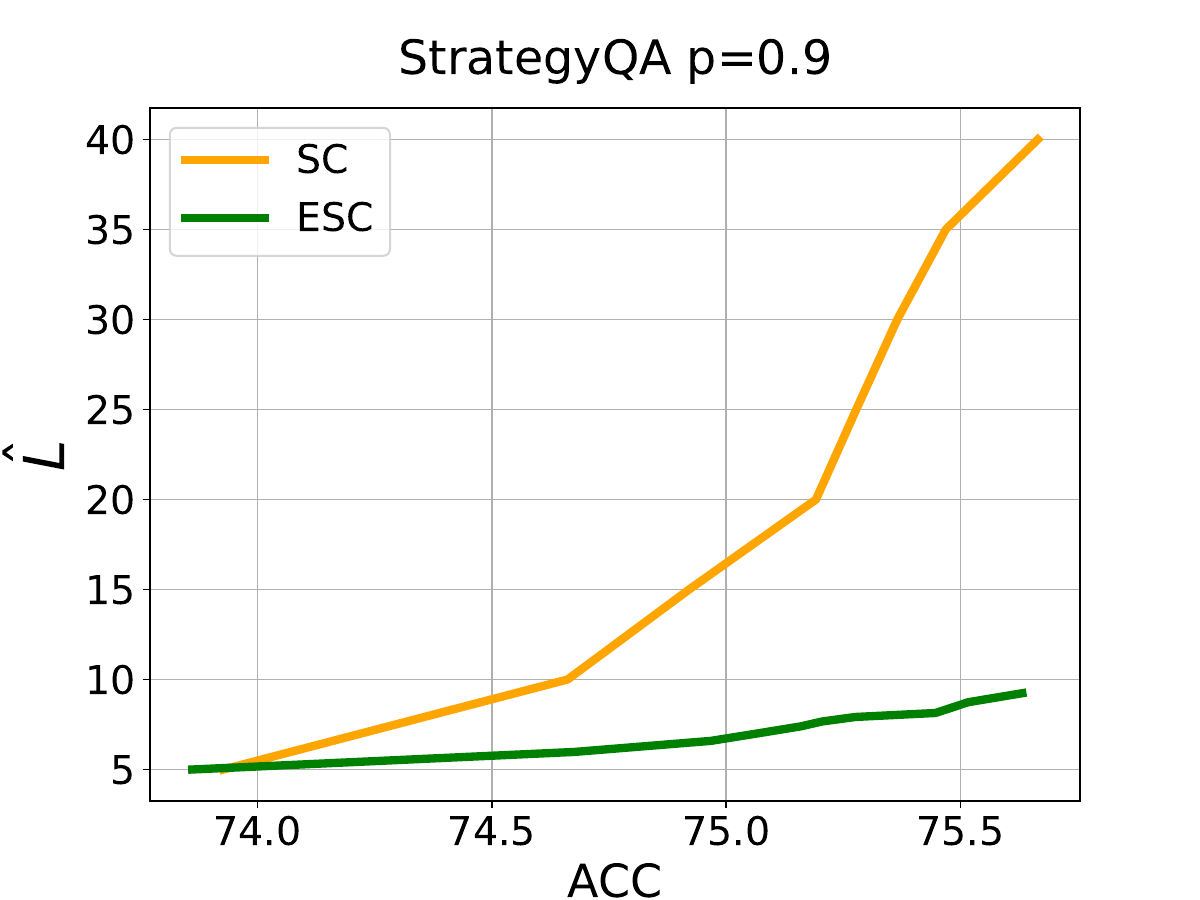}} \hspace{-0.50cm}
    \subfigure{\includegraphics[width=0.267\hsize, height=0.23\hsize]{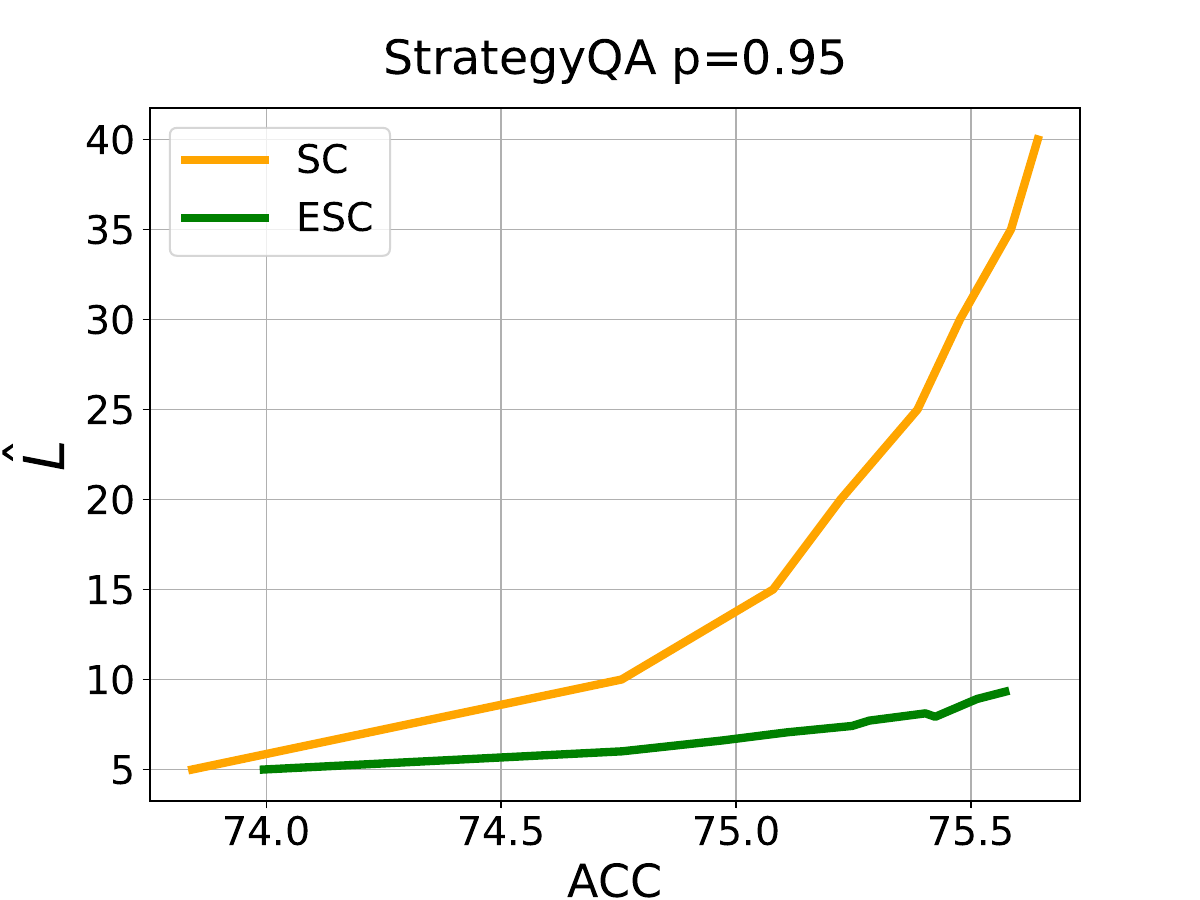}} \hspace{-0.47cm}
    \subfigure{\includegraphics[width=0.267\hsize, height=0.23\hsize]{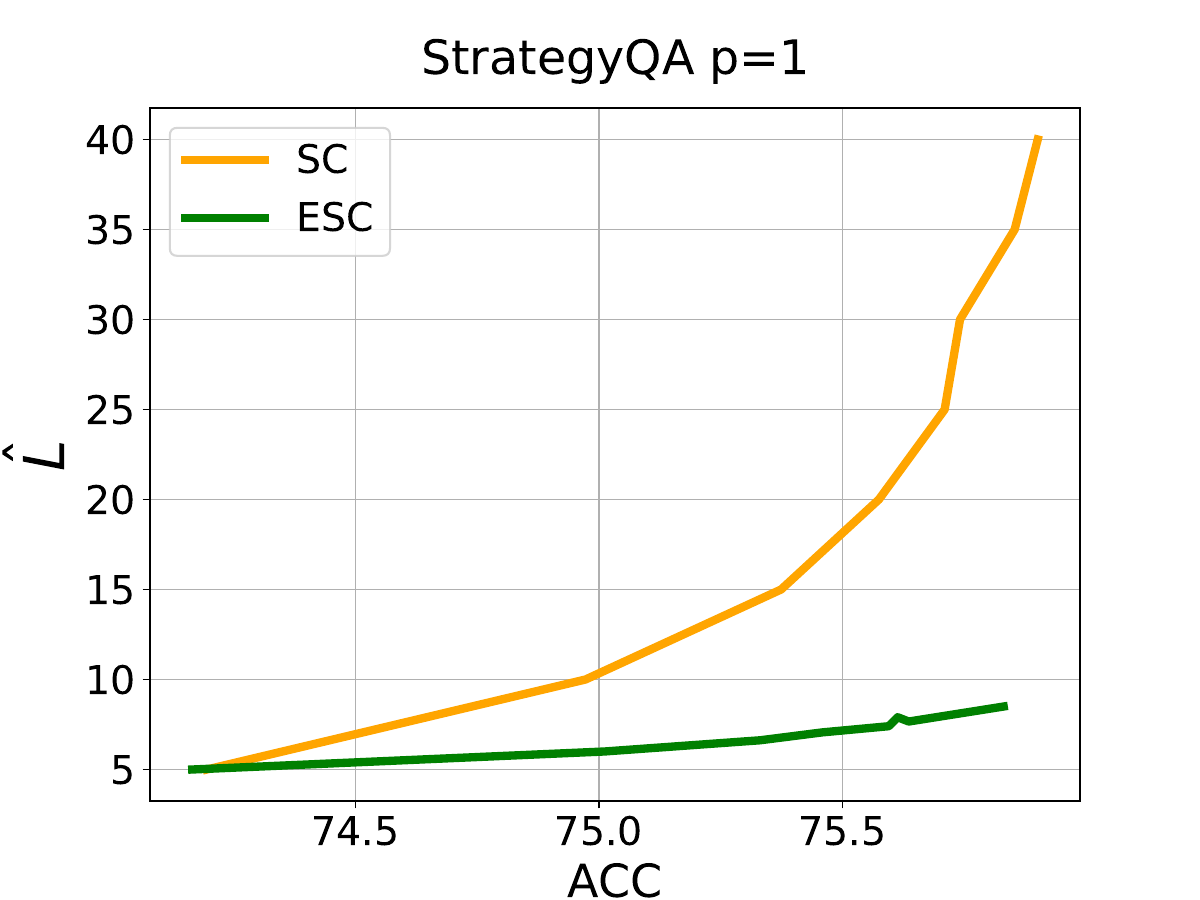}}\hspace{-0.4cm}
    \subfigure{\includegraphics[width=0.267\hsize, height=0.23\hsize]{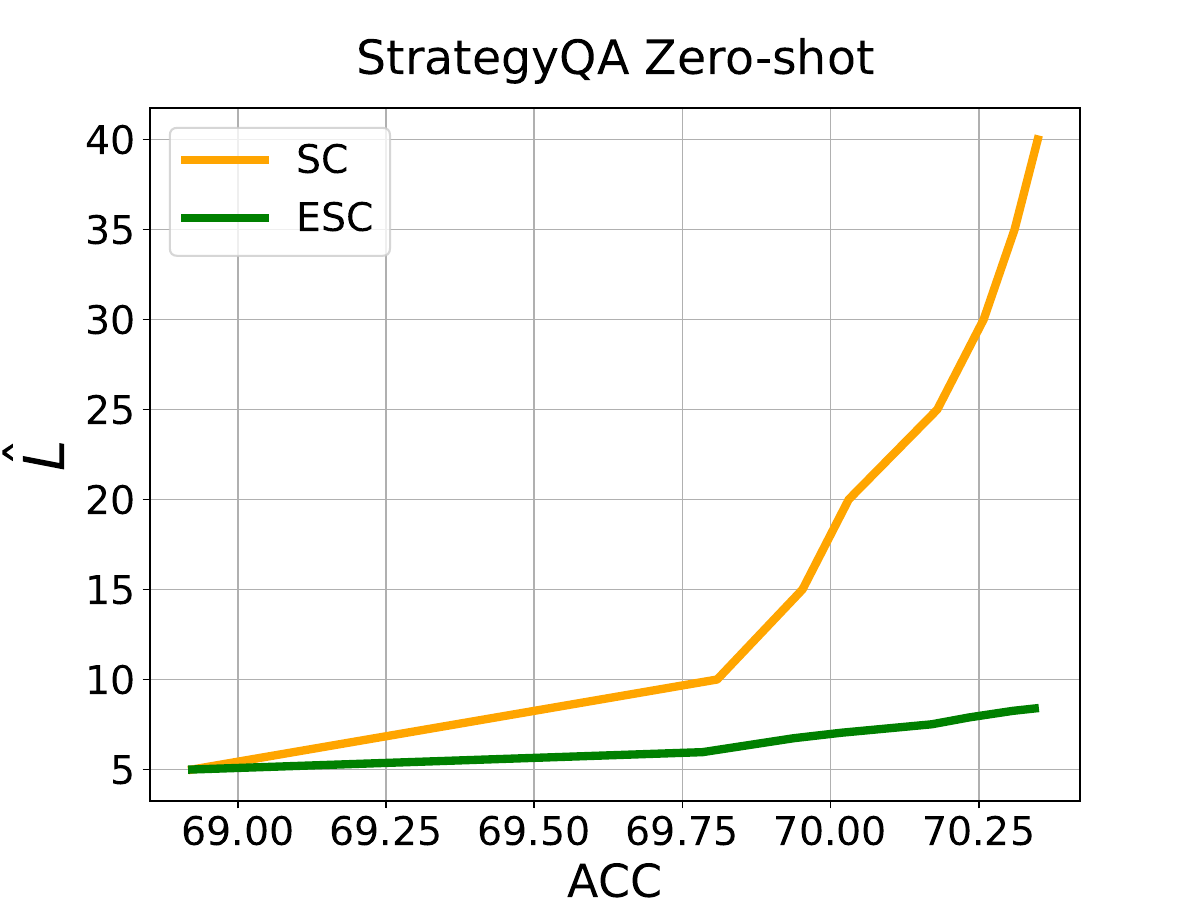}}
    \setlength{\belowcaptionskip}{0pt}
    \caption{Robustness analysis of ESC regarding the sampling temperature $T$, $p$ in nucleus sampling, and zero-shot demonstration on StrategyQA with GPT-3.5-Turbo.}
    \label{fig-appendix:robust}
\end{figure*}

\begin{figure*}[!htb]
    \centering
    \subfigure{\includegraphics[width=0.32\hsize, height=0.25\hsize]{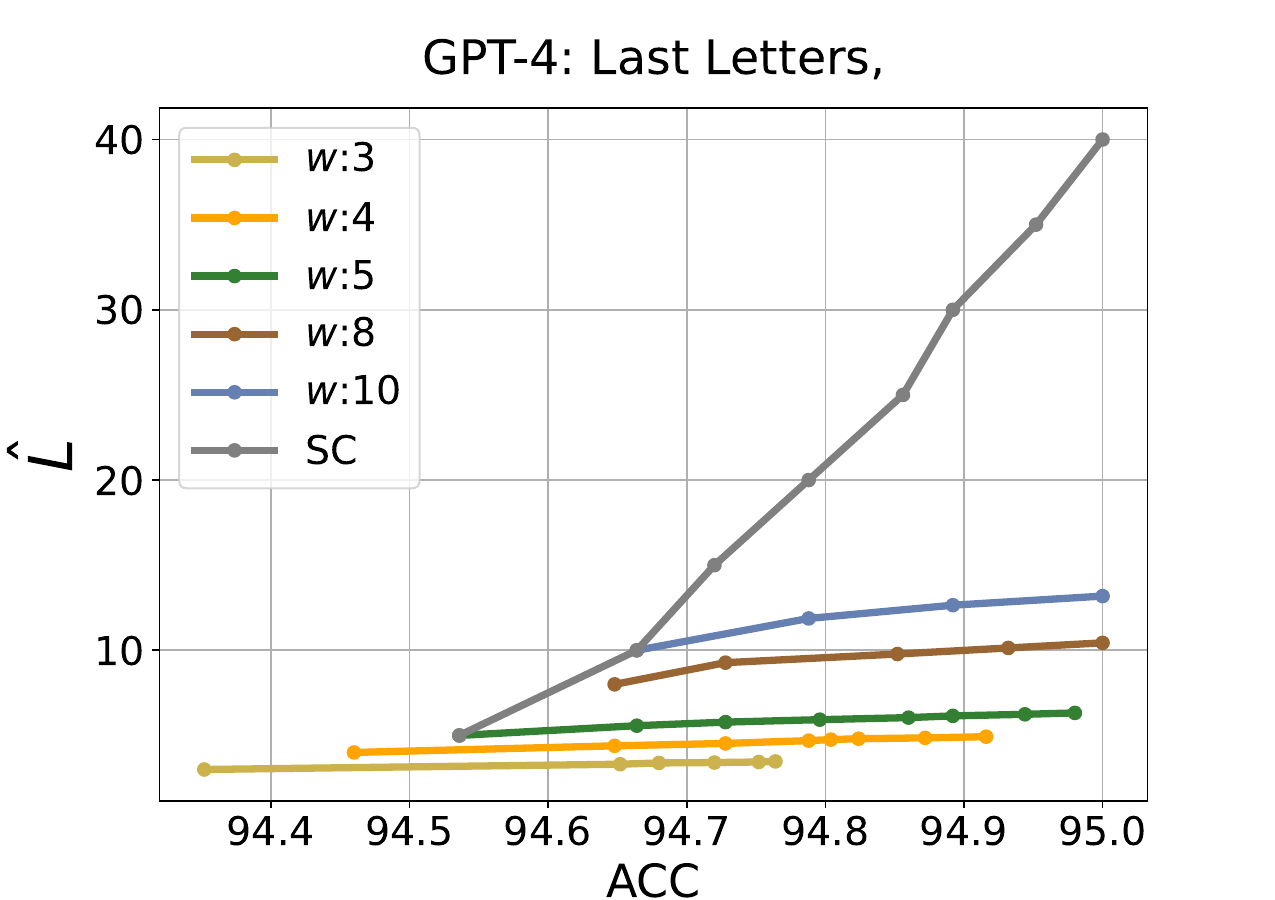}} \hspace{-0.50cm}
    \subfigure{\includegraphics[width=0.32\hsize, height=0.25\hsize]{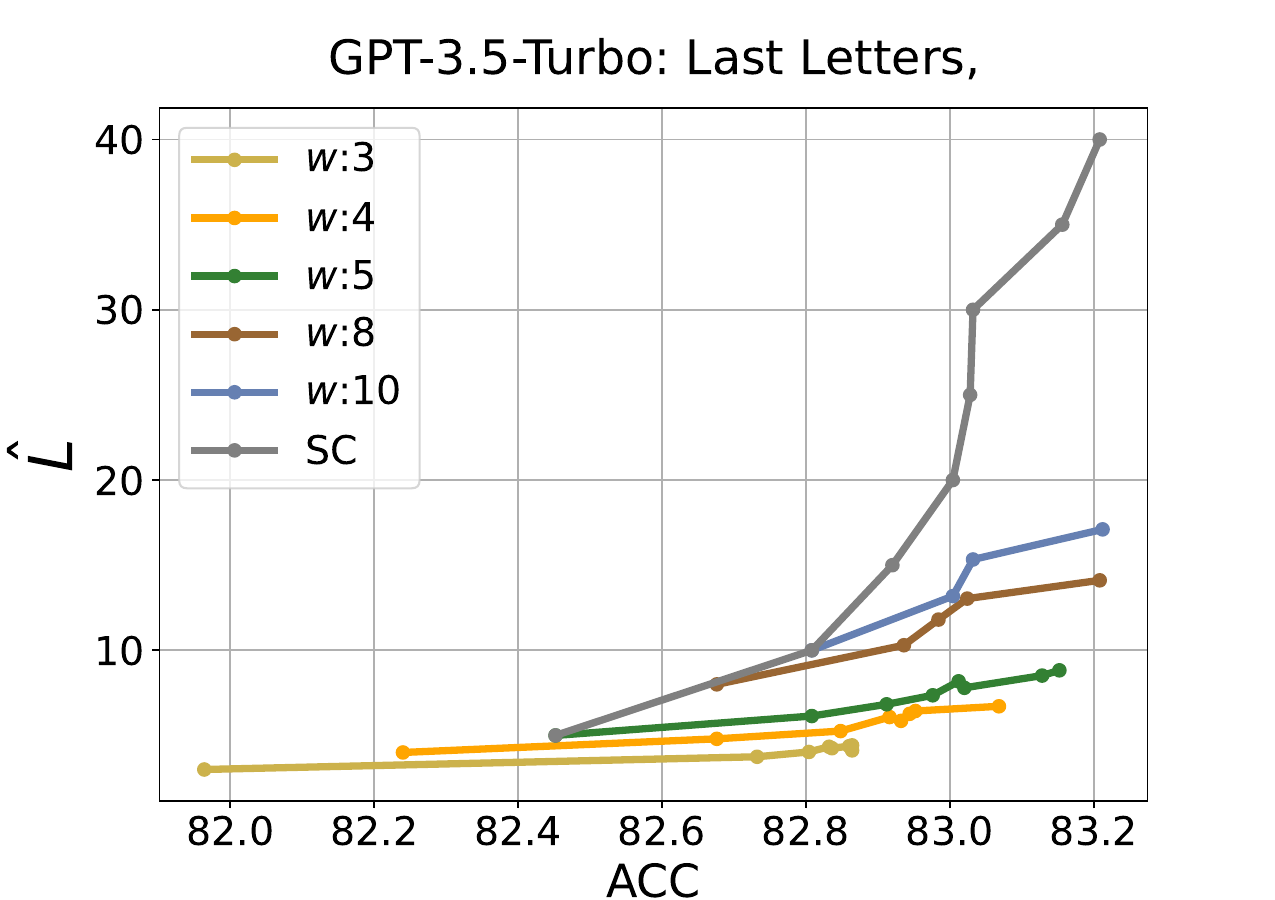}}\hspace{-0.47cm}
    \subfigure{\includegraphics[width=0.32\hsize, height=0.25\hsize]{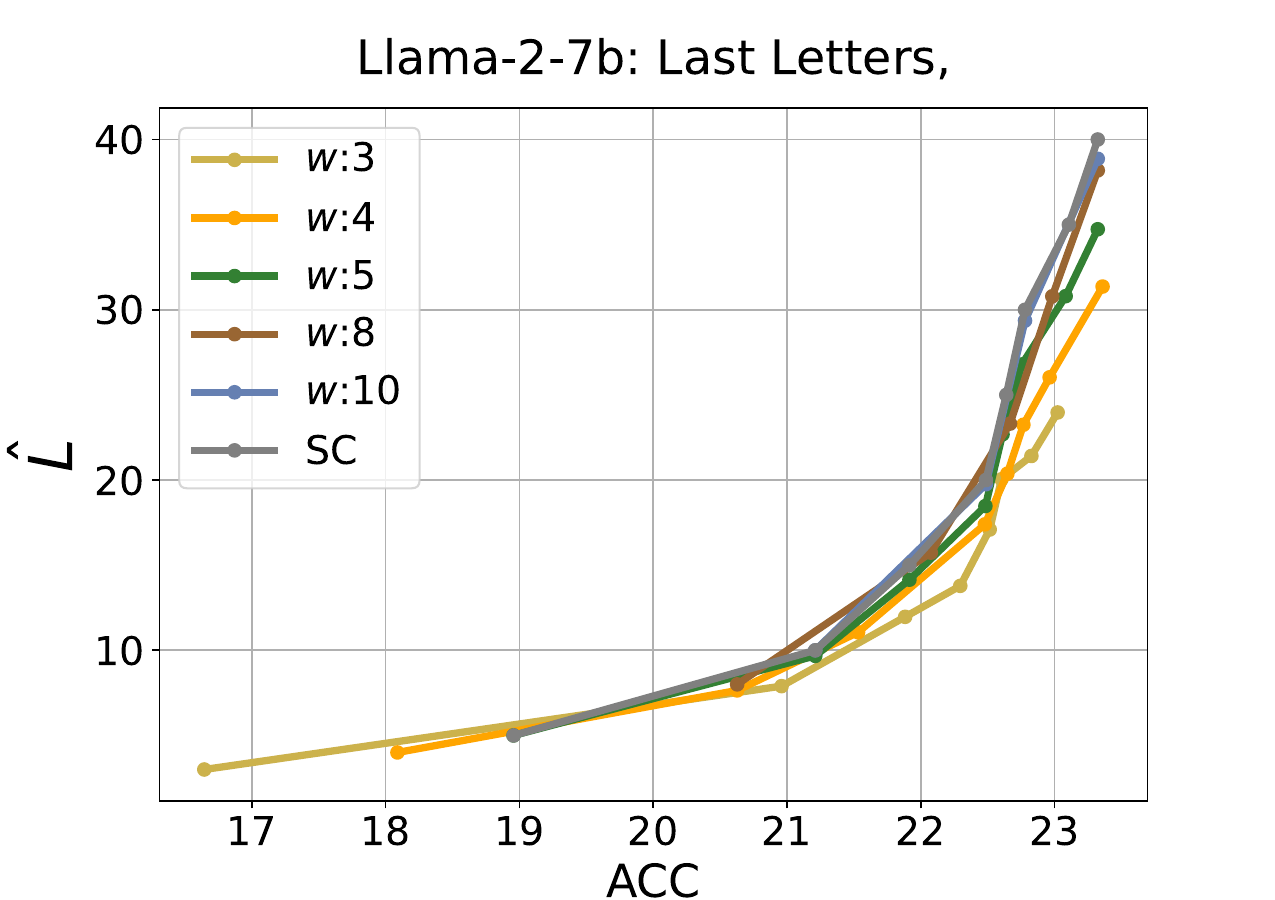}}\hspace{-0.4cm}
    \subfigure{\includegraphics[width=0.32\hsize, height=0.25\hsize]{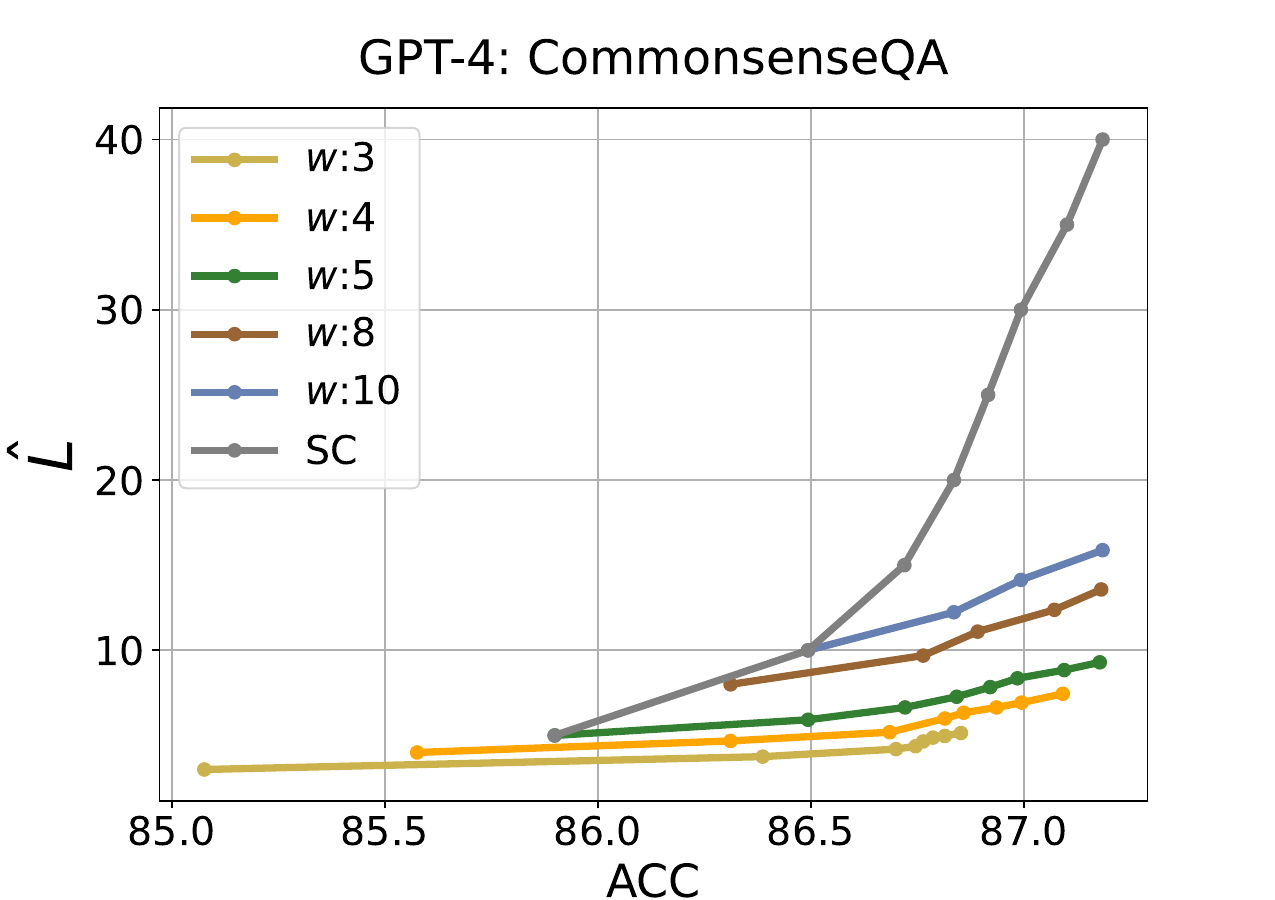}} \hspace{-0.50cm}
    \subfigure{\includegraphics[width=0.32\hsize, height=0.25\hsize]{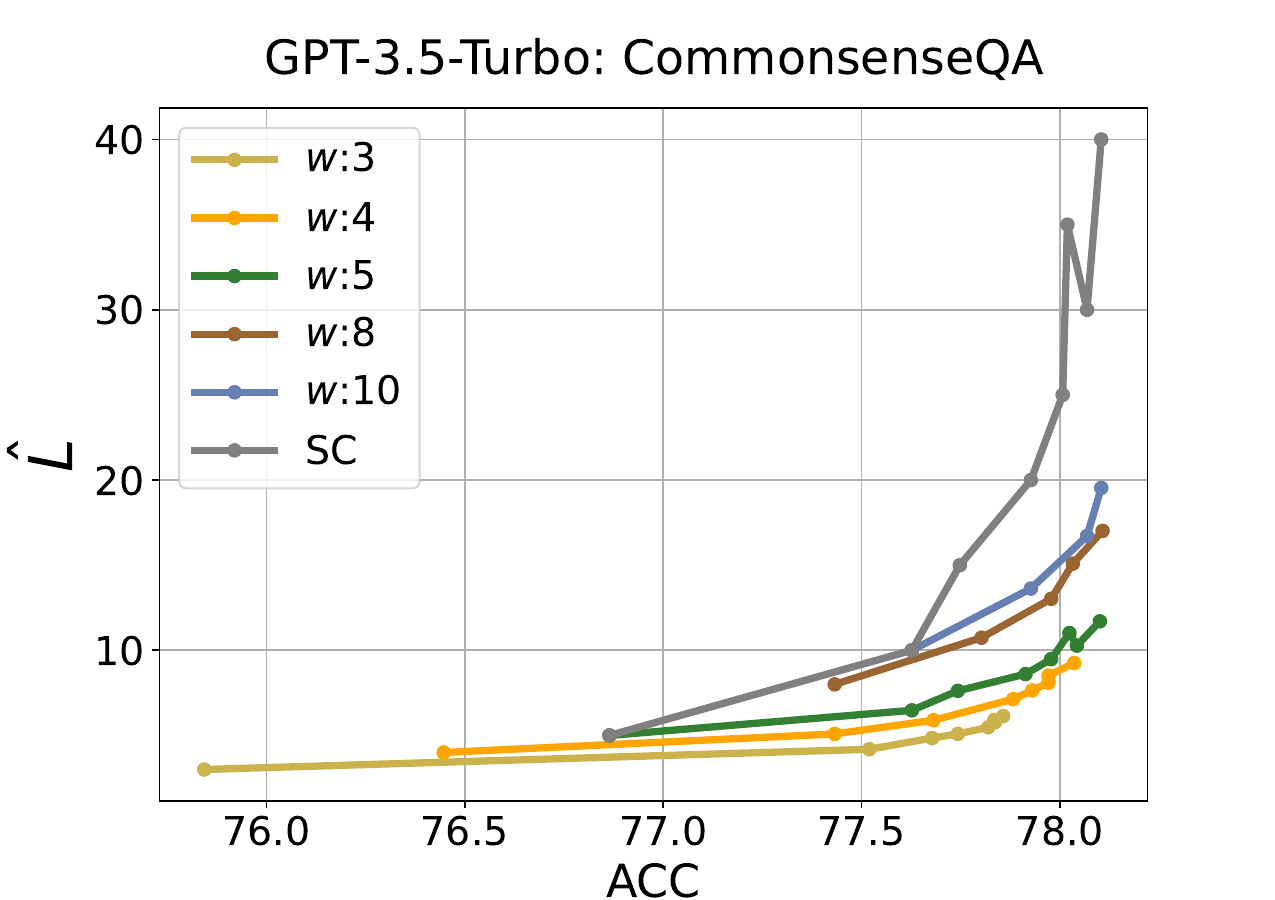}}\hspace{-0.47cm}
    \subfigure{\includegraphics[width=0.32\hsize, height=0.25\hsize]{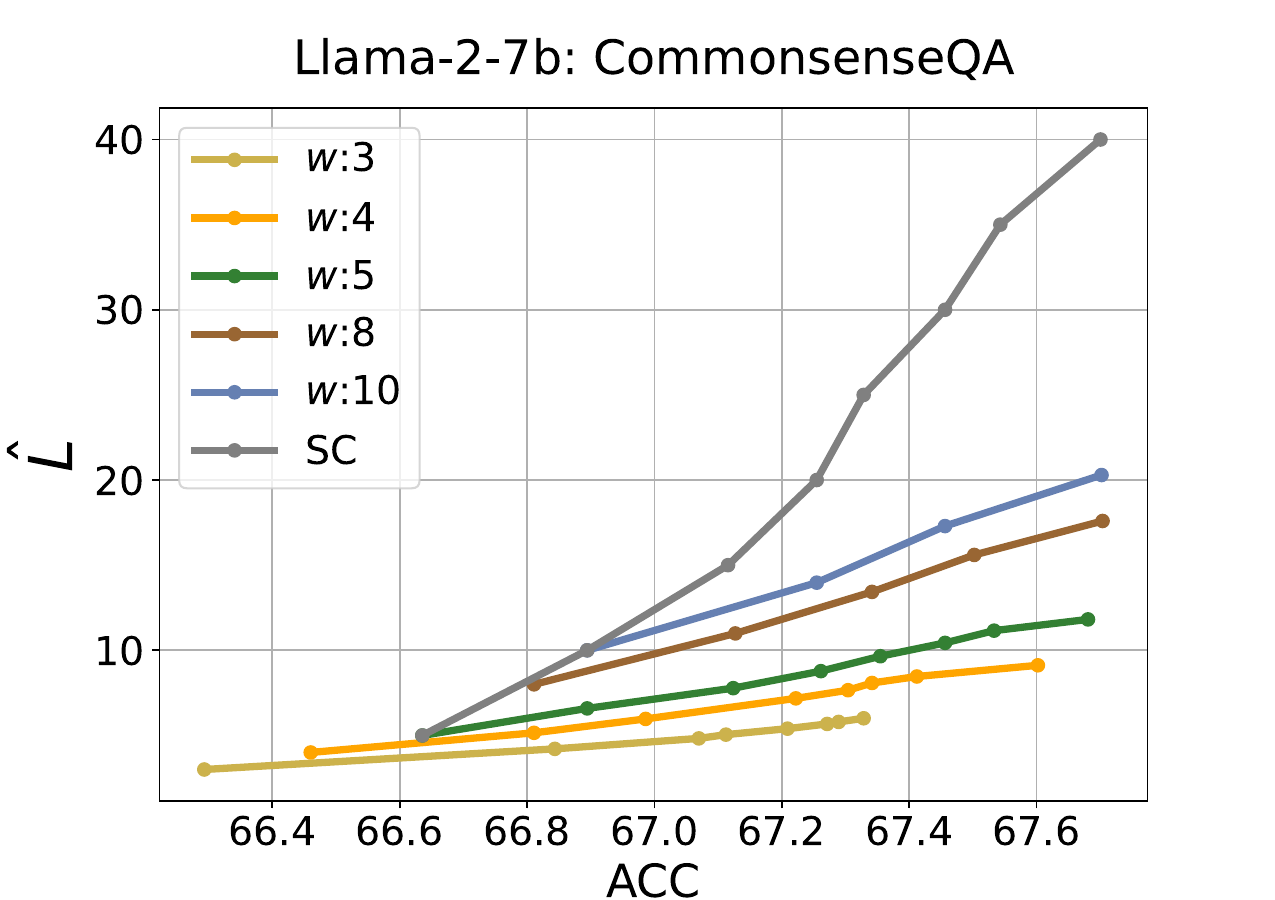}}\hspace{-0.4cm}
    \subfigure{\includegraphics[width=0.32\hsize, height=0.25\hsize]{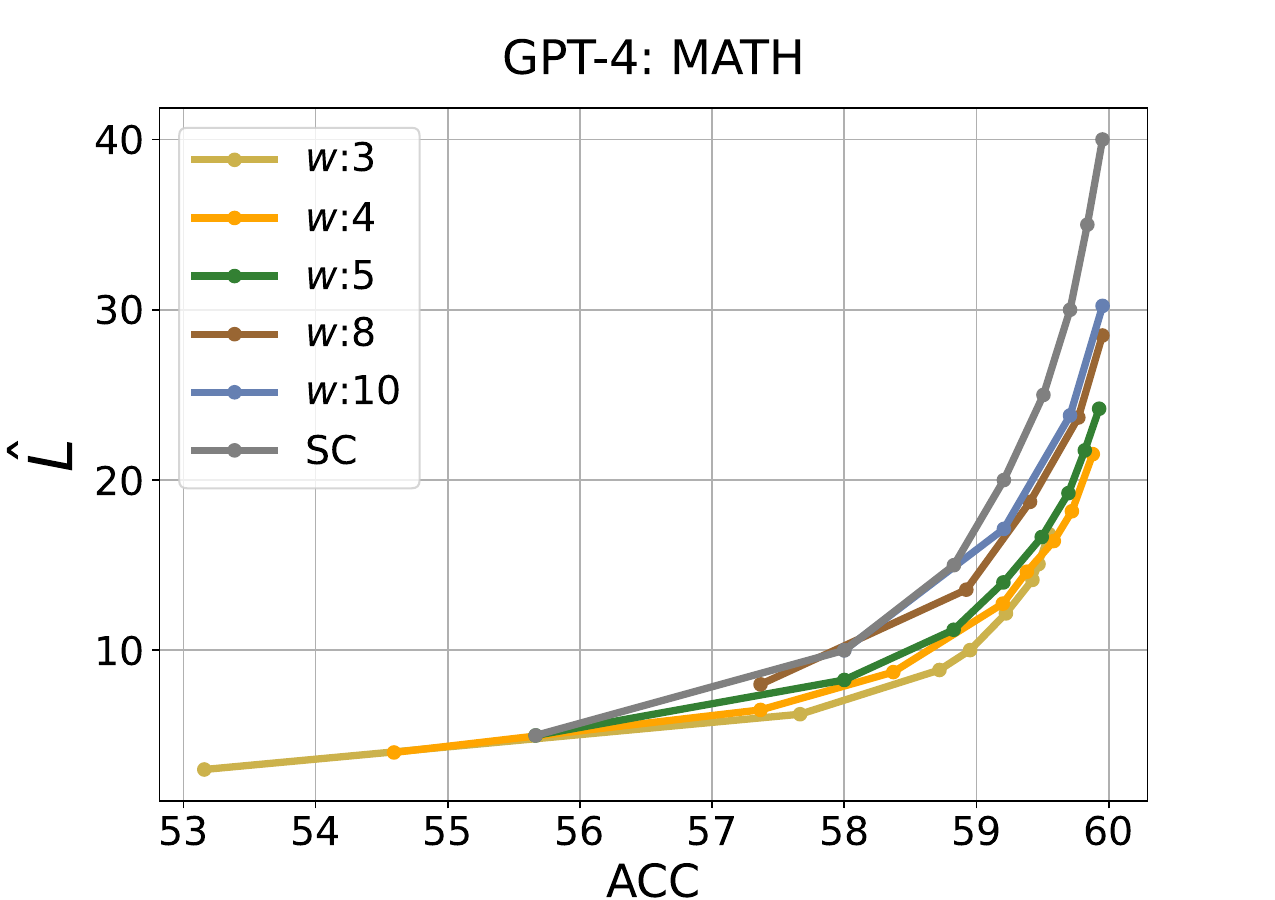}} \hspace{-0.50cm}
    \subfigure{\includegraphics[width=0.32\hsize, height=0.25\hsize]{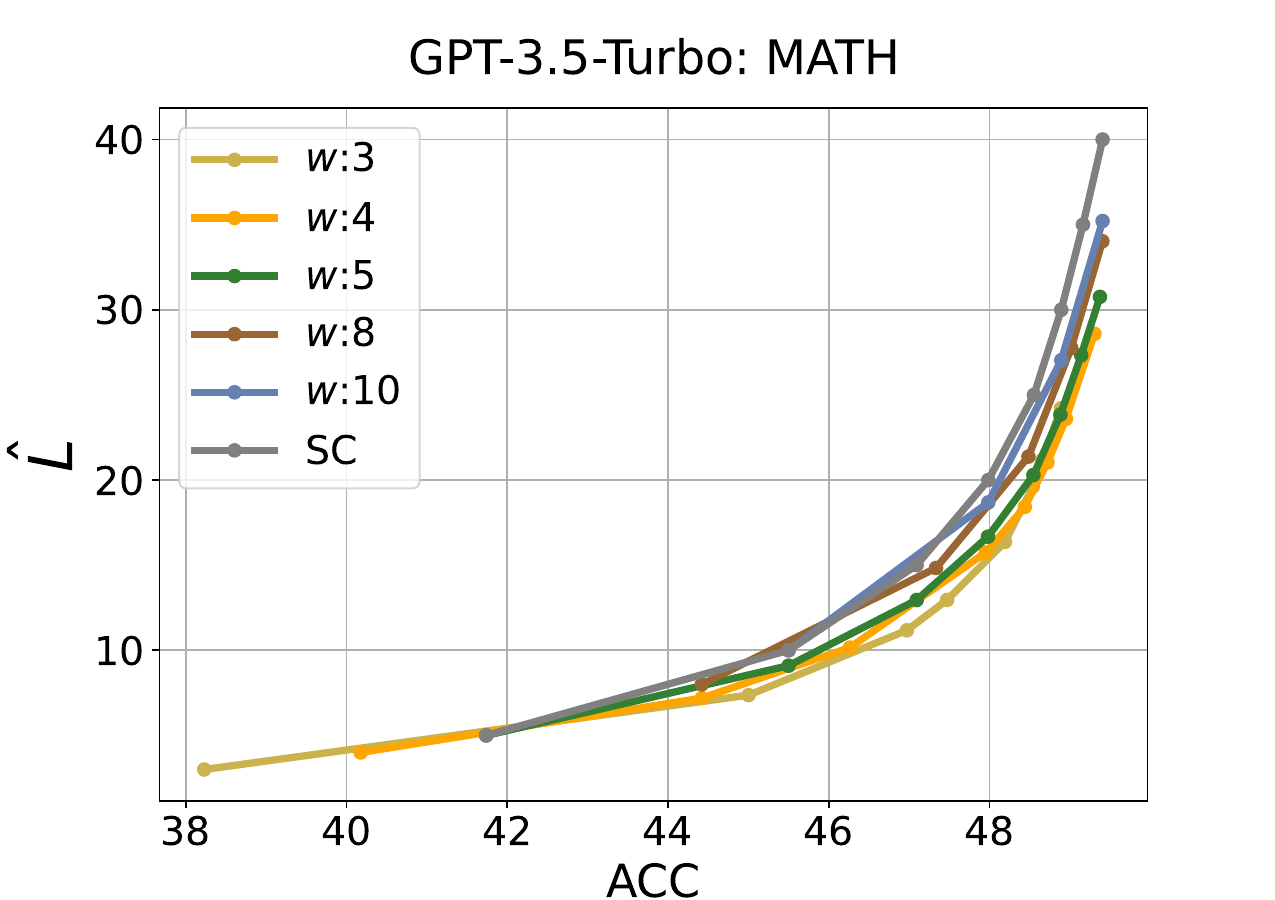}}\hspace{-0.47cm}
    \subfigure{\includegraphics[width=0.32\hsize, height=0.25\hsize]{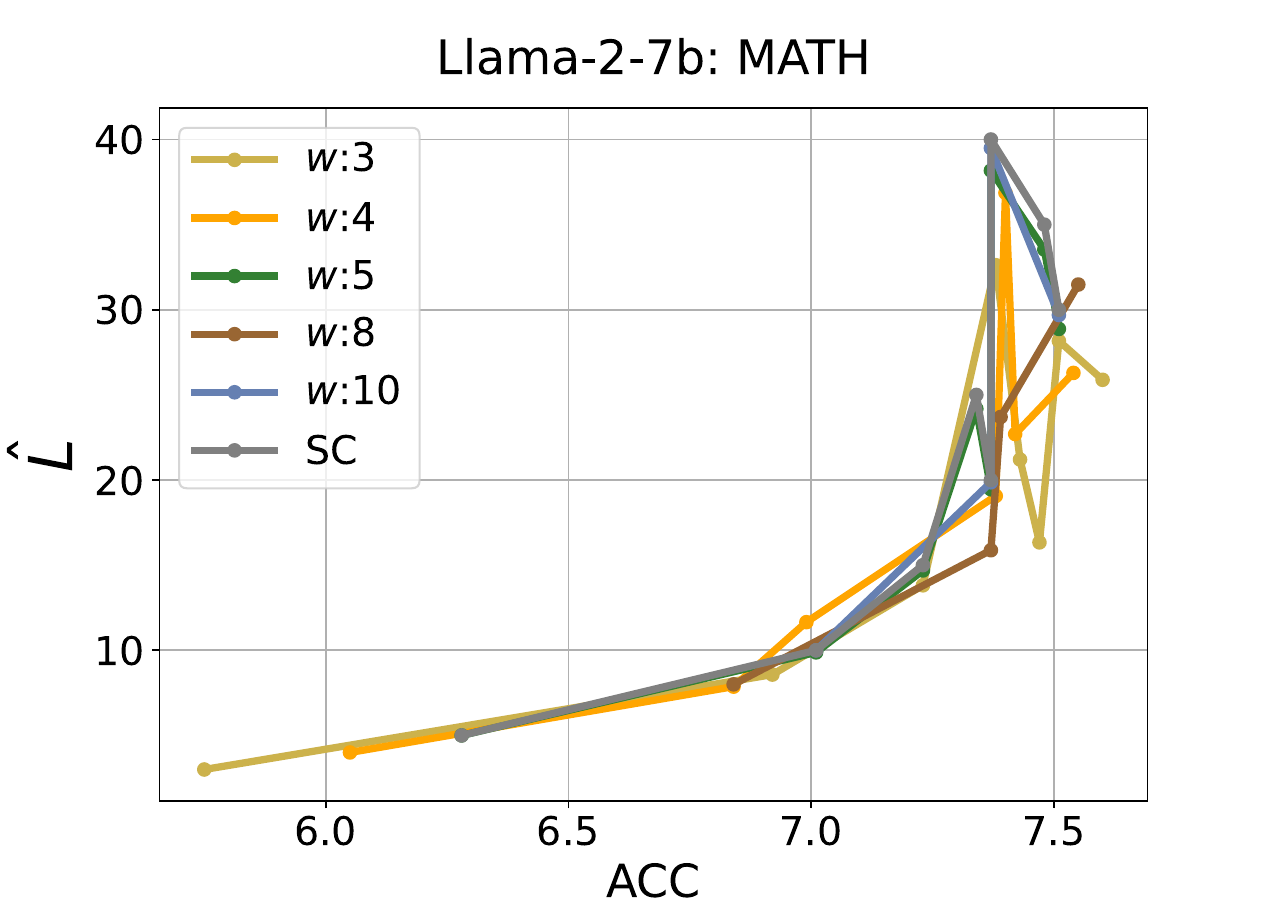}}\hspace{-0.4cm}
    \subfigure{\includegraphics[width=0.32\hsize, height=0.25\hsize]{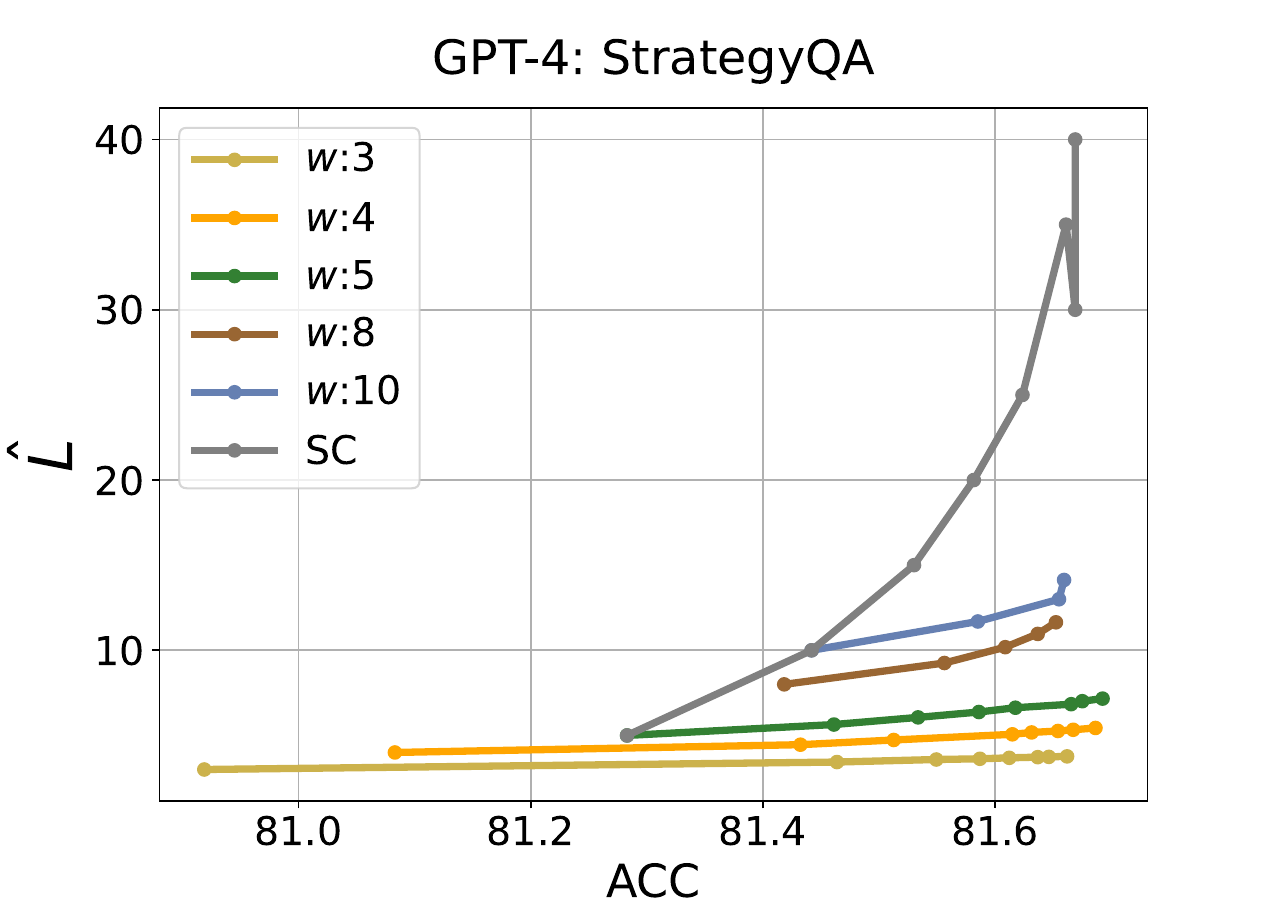}} \hspace{-0.50cm}
    \subfigure{\includegraphics[width=0.32\hsize, height=0.25\hsize]{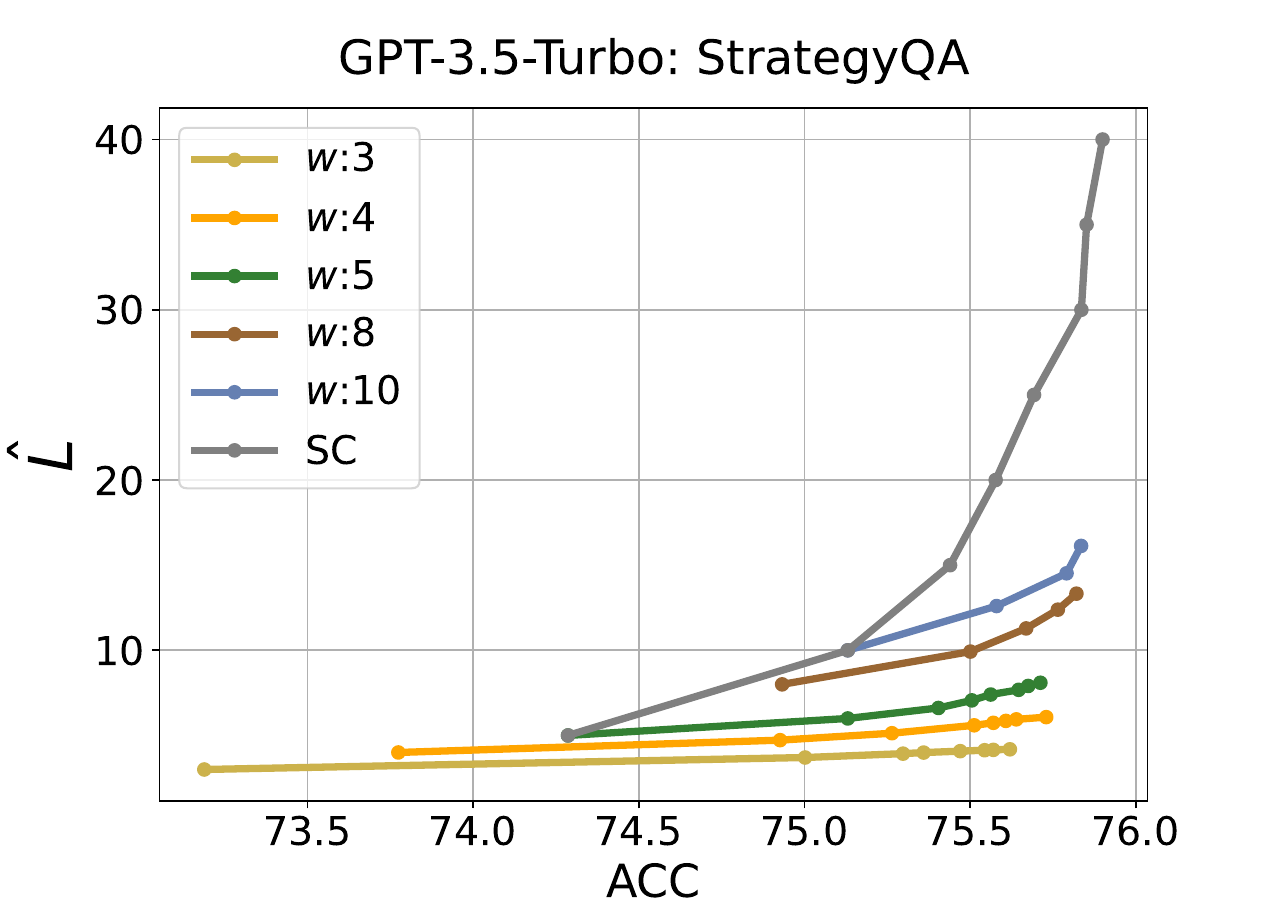} }\hspace{-0.47cm}
    \subfigure{\includegraphics[width=0.32\hsize, height=0.25\hsize]{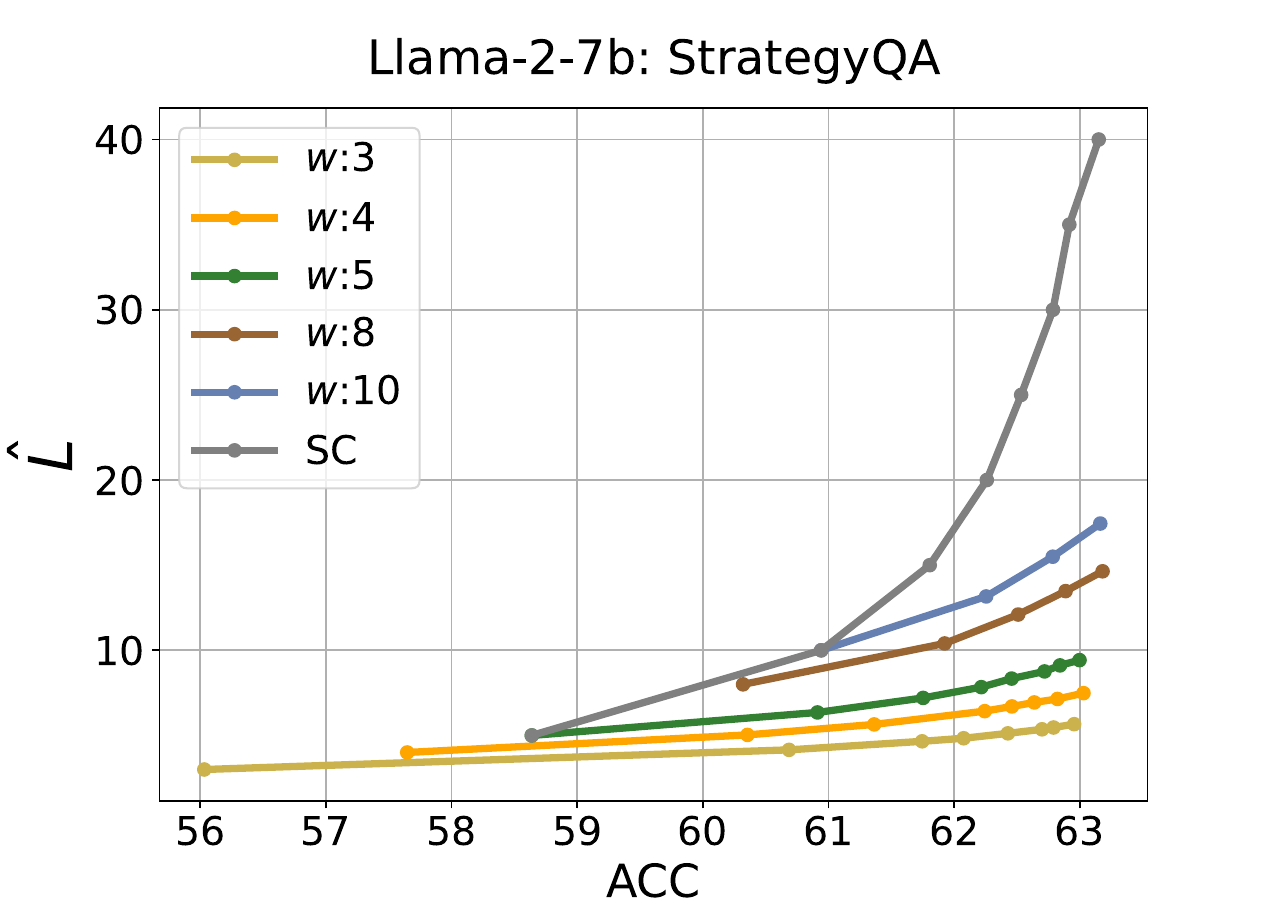}}\hspace{-0.4cm}
    \subfigure{\includegraphics[width=0.32\hsize, height=0.25\hsize]{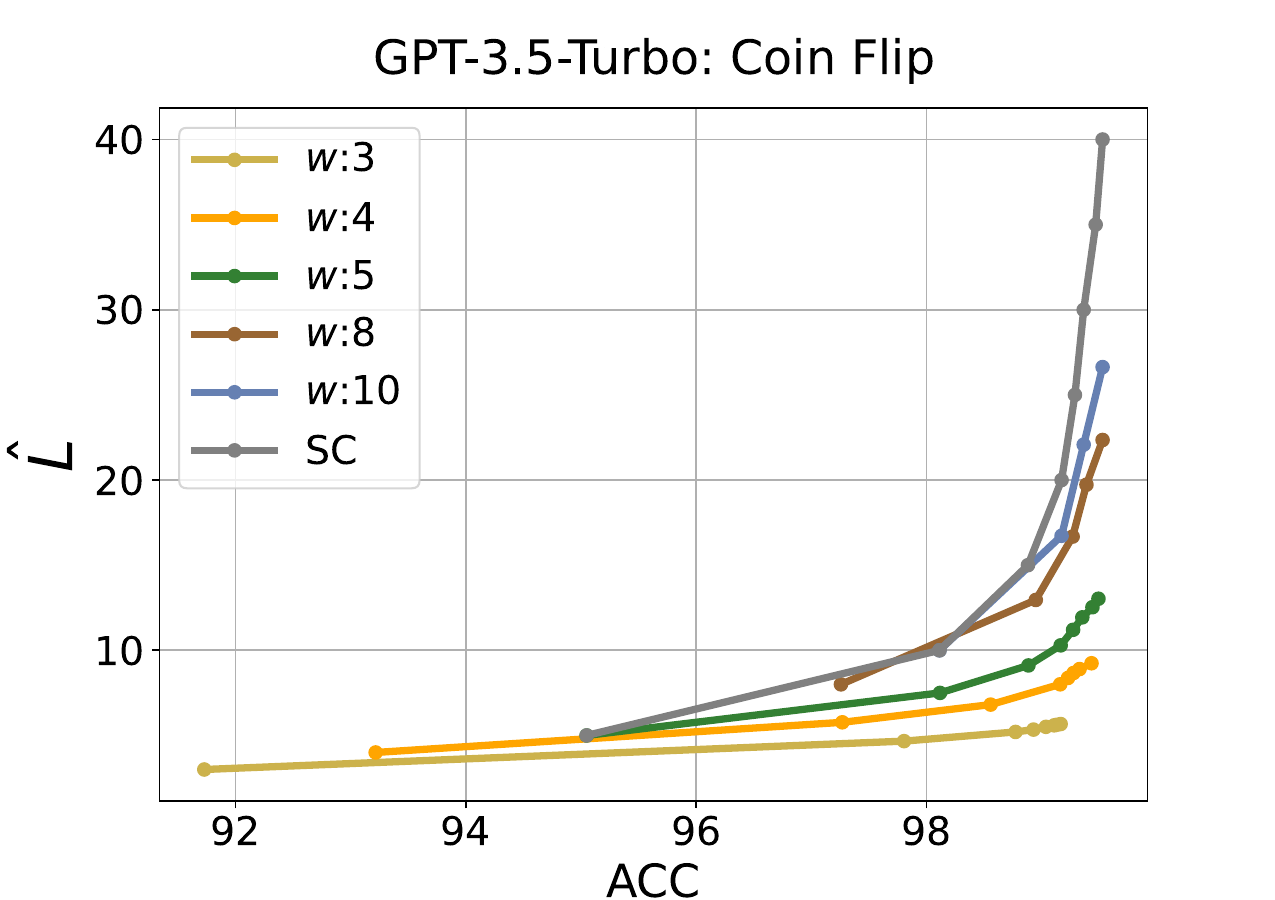}}\hspace{-0.47cm}
    \subfigure{\includegraphics[width=0.32\hsize, height=0.25\hsize]{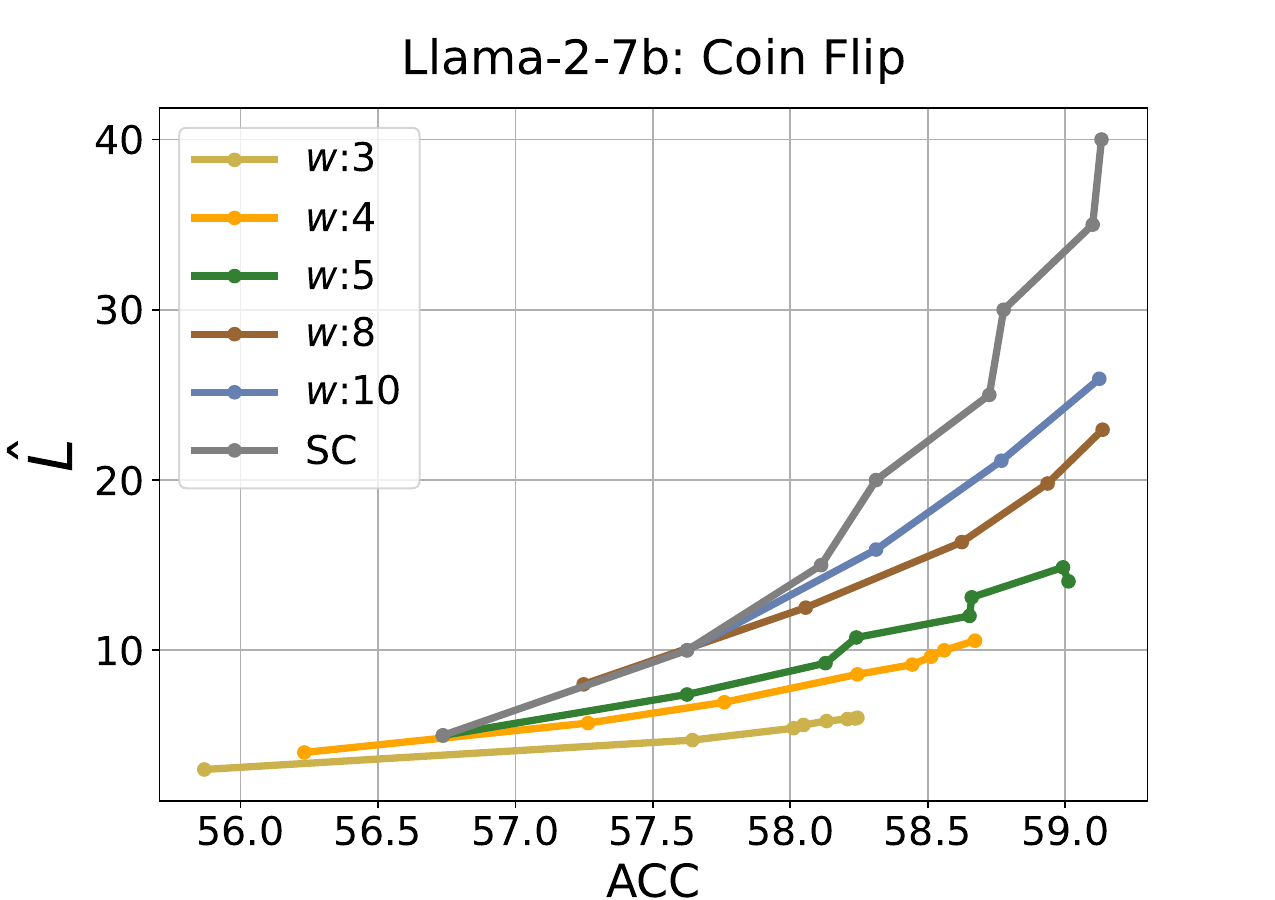}}\hspace{-0.4cm}
    \setlength{\belowcaptionskip}{0pt}
    \caption{Robustness analysis of ESC regarding the observation window size $w$ and max sampling size $L$ on multiple datasets with different models.}
    \label{fig:appendix-figs}
\end{figure*}

Figure~\ref{fig-appendix:robust} shows the robustness of ESC regarding the sampling temperature $T$, $p$ in nucleus sampling, and zero-shot demonstration on StrategyQA with GPT-3.5-Turbo. Figure~\ref{fig:appendix-figs} shows the robustness of ESC regarding the observation window size $w$ and max sampling size $L$ on multiple datasets with different models.

\subsection{Prompting Details}

We list the details of the prompts used for MATH dataset. Following \citet{SC}, we use the same prompts for other datasets as in \citet{COT} for fair comparison.

\begin{table}[t]
\centering
\caption{Few-shot prompts for MATH dataset.}
\begin{tabular}{p{13cm}}

\toprule
\textbf{system:} Think the question step by step and give the answer. \\
\textbf{user:} Problem:
Find the domain of the expression $\frac{\sqrt{x-2}}{\sqrt{5-x}}$. \\
\textbf{assistant:} Solution:
The expressions inside each square root must be non-negative. Therefore, $x-2 \ge 0$, so $x\ge2$, and $5 - x \ge 0$, so $x \le 5$. Also, the denominator cannot be equal to zero, so $5-x>0$, which gives $x<5$. Therefore, the domain of the expression is $\boxed{[2,5)}$. \\
\textbf{user:} Problem:
Compute: $55\times1212-15\times1212$ . \\
\textbf{assistant:} Solution:
We have $55 \times 1212 - 15 \times 1212 = 1212(55-15) = 1212(40) = 4848(10) = \boxed{48480}$. \\
\textbf{user:} Problem:
Compute $\dbinom{16}{15}$. \\
\textbf{assistant:} Solution:
$\dbinom{16}{15}=\dbinom{16}{1}=\boxed{16}.$ \\
\textbf{user:} Problem:
Find $x$, where $x$ is a square number which is divisible by four, and between 39 and 80. \\
\textbf{assistant:} Solution:
We know that $x$ is between 39 and 80, and since $6^2 = 36 < 39$ and $9^2 = 81 > 80$, this means that $6^2 < x < 9^2$. This leaves us with two possibilities for $x$, which are $7^2 = 49$, and $8^2 = 64$. We then see that only 64 is divisible by four, so $x =$ $\boxed{64}$. \\
\textbf{user:} Problem:
Solve the inequality
\[\frac{(x - 2)(x - 3)(x - 4)}{(x - 1)(x - 5)(x - 6)} > 0.\] \\
\textbf{assistant:} Solution:
We can build a sign chart, but since all of the factors are linear, we can track what happens to the expression as $x$ increases.  At $x = 0,$ the expression is positive.  As $x$ increases past 1, the expression becomes negative.  As $x$ increases past 2, the expression becomes positive, and so on.  Thus, the solution is
\[x \in \boxed{(-\infty,1) \cup (2,3) \cup (4,5) \cup (6,\infty)}.\] \\
\textbf{user:} Problem:
A right circular cone has a volume of $12\pi$ cubic centimeters. The height of the cone is 4 cm. How many centimeters is the circumference of the base of the cone, in terms of $\pi$? \\
\textbf{assistant:} Solution:
The volume of a cone is $\frac{1}{3}\pi r^2 h$. We are given that the volume is $12\pi$ and the height is $4$. Thus, $\frac{1}{3}\pi r^2 \cdot 4 = 12\pi$. Solving for $r$, we find $r = 3$. Therefore, the circumference of the base is $2\pi r = \boxed{6\pi}$. \\
\textbf{user:} Problem:
How many perfect squares less than 1000 have a ones digit of 2, 3 or 4? \\
\textbf{assistant:} Solution:
Checking the squares from $1^2$ to $10^2$, we see that no squares end in 2 or 3, while a square ends in 4 if its square root ends in 2 or 8.  Since $31^2 < 1000 < 32^2$, we see that the squares less than 1000 ending in 4 are $2,8,12,18,22,28$.  Thus the desired answer is $\boxed{6}$. \\
\textbf{user:} Problem:
The diagonals of a rhombus are $10$ inches and $24$ inches. What is the perimeter of the rhombus, in inches? \\
\textbf{assistant:} Solution:
The diagonals are perpendicular bisectors of each other, so therefore the side length of the rhombus can be calculated as $\sqrt{5^2+12^2} = 13$. Therefore, the perimeter of the rhombus is $4 \times 13 = \boxed{52}$ inches. \\

\bottomrule
\end{tabular}
\label{tb:demon}
\end{table}

\end{document}